\documentclass{article}

\PassOptionsToPackage{numbers, compress}{natbib}

\usepackage[final]{neurips_2021}

\usepackage{diagbox}
\usepackage[utf8]{inputenc} 
\usepackage[T1]{fontenc}    
\usepackage{hyperref}       
\usepackage{url}            
\usepackage{booktabs}       
\usepackage{amsfonts}       
\usepackage{nicefrac}       
\usepackage{caption} 
\usepackage{thmtools}
\usepackage{thm-restate}

\usepackage{microtype}      
\usepackage{xcolor}         
\usepackage{amsmath}
\usepackage{amssymb}             
\usepackage{amsfonts}            
\usepackage{mathrsfs} 

\usepackage{amsmath,amsfonts,bm}

\def\eqref#1{equation~\ref{#1}}

\def\1{\bm{1}}

\DeclareMathAlphabet{\mathsfit}{\encodingdefault}{\sfdefault}{m}{sl}
\SetMathAlphabet{\mathsfit}{bold}{\encodingdefault}{\sfdefault}{bx}{n}

\usepackage{graphicx}
\usepackage{subfigure}
\usepackage{wrapfig}
\usepackage{float}

\title{Episodic Multi-agent Reinforcement Learning with Curiosity-driven Exploration}

\author{%
	Lulu Zheng\thanks{Equal contribution.}~~$^1$, Jiarui Chen\footnotemark[1]~~$^{2~3}$, Jianhao Wang$^1$, Jiamin He$^4$\thanks{Work performed while visiting Tsinghua Univeristy.}~~, Yujing Hu$^3$, Yingfeng Chen$^3$,\\ \textbf{Changjie Fan$^3$, Yang Gao$^2$, Chongjie Zhang$^1$} \\
	$^1$Institute for Interdisciplinary Information Sciences, Tsinghua University, China \\
	$^2$Department of Computer Science and Technology, Nanjing University, China\\
	$^3$Fuxi AI Lab, NetEase, China\\
	$^4$Department of Computing Science, University of Alberta, Canada\\
	\texttt{zll19@mails.tsinghua.edu.cn}\\
	\texttt{chenjiarui@smail.nju.edu.cn}\\
	\texttt{wjh19@mails.tsinghua.edu.cn}\\
	\texttt{jiamin12@ualberta.ca}\\
	\texttt{\{huyujing, chenyingfeng01, fanchangjie\}@corp.netease.com}\\
	\texttt{gaoy@nju.edu.cn}\\
	\texttt{chongjie@tsinghua.edu.cn} \\
}

\begin{document}

\maketitle

\begin{abstract}
 Efficient exploration in deep cooperative \emph{multi-agent reinforcement learning} (MARL) still remains challenging in complex coordination problems. In this paper, we introduce a novel Episodic Multi-agent reinforcement learning with Curiosity-driven exploration, called EMC. We leverage an insight of popular factorized MARL algorithms that the ``induced" individual Q-values, i.e., the individual utility functions  used for local execution,  are the embeddings of local action-observation histories, and can capture the interaction between agents due to reward backpropagation during centralized training. Therefore, we use prediction errors of individual Q-values as intrinsic rewards for coordinated exploration and utilize episodic memory to exploit explored informative experience to boost policy training. As the dynamics of an agent's individual Q-value function captures the novelty of states and the influence from other agents, our intrinsic reward can induce coordinated exploration to new or promising states. We illustrate the advantages of our method by didactic examples, and demonstrate its significant outperformance over state-of-the-art MARL baselines on challenging tasks in the StarCraft II micromanagement benchmark.
\end{abstract}

\section{Introduction}
\label{introduction}
Cooperative \emph{multi-agent reinforcement learning} (MARL) has great promise to solve many real-world multi-agent problems, such as autonomous cars \cite{car} and robots \cite{robot}. These complex applications post two major challenges for cooperative MARL: {\em scalability}, i.e., the joint-action space exponentially grows as the number of agents increases, and {\em partial observability}, which requires agents to make decentralized decisions based on their local action-observation histories due to communication constraints. Luckily, a popular MARL paradigm, called \emph{centralized training with decentralized execution} (CTDE) \cite{CTDE2002}, is adopted to deal with these challenges. With this paradigm, agents’ policies are trained with access to global information in a centralized way and executed only based on local histories in a decentralized way. Based on the paradigm of CTDE, many deep MARL methods have been proposed, including VDN \cite{VDN}, QMIX \cite{QMIX}, QTRAN \cite{QTRAN}, and QPLEX \cite{qplex}.

A core idea of these approaches is to use value factorization, which uses neural networks to represent the joint state-action value as a function of individual utility functions, which can be referred to \textit{individial Q-values} for terminological simplicity. For example, VDN learns a centralized but factorizable joint value function $Q_{tot}$ represented as the summation of individual value function~$Q_i$. 
During  execution, the decentralized policies can be easily derived for each agent $i$ by greedily selecting actions with respect to its local value function $Q_i$. By utilizing this factorization structure, an implicit multi-agent credit assignment is realized because $Q_i$ is represented as a latent embedding and is learned by neural network backpropagation from the total temporal-difference error on the single global reward signal, rather than on a local reward signal specific to agent $i$. This value factorization technique enables value-based MARL approaches, such as QMIX and QPLEX, to achieve state-of-the-art performance in challenging tasks such as the StarCraft unit micromanagement~\citep{SMAC}. 

Despite the current success, since only using simple $\epsilon$-greedy exploration strategy, these deep MARL approaches are found ineffective to solve complex coordination tasks that require coordinated and efficient exploration~\cite{qplex}. Exploration has been extensively studied in single-agent reinforcement learning and many advanced methods have been proposed, including pseudo-counts \cite{COUNT,ostrovski2017count}, curiosity \cite{ICM,RND}, and information gain \cite{VIME}. However, these methods cannot be adopted into MARL directly, due to the exponentially growing state space and partial observability, leaving multi-agent exploration challenging. Recently, only a few works have tried to address this problem. 
 For instance, EDTI \cite{EDTI} uses influence-based methods to quantify the value of agents' interactions and coordinate exploration towards high-value interactions. This approach empirically shows promising results but, because of the need to explicitly estimate the influence among agents, it is not scalable when the number of agents increases. Another method, called MAVEN \cite{MAVEN}, introduces a hierarchical control method with a shared latent variable encouraging committed, temporally extended exploration. However, since the latent variable still needs to explore in the space of joint behaviours~\cite{MAVEN}, it is not efficient in complex tasks with large state spaces.

In this paper, we propose a novel multi-agent curiosity-driven exploration method. Curiosity is a type of intrinsic motivation for exploration, which usually uses prediction errors on different spaces (e.g., future observations \citep{RND}, actions \citep{ICM}, or learnable representation \citep{kim2018emi}) as a reward signal. Recently, curiosity-driven methods have achieved significant success in single-agent reinforcement learning~\cite{RND,NGU,Agent57}. However, curiosity-driven methods face a critical challenge in MARL: in which space should we define curiosity? The straightforward method is to measure curiosity on the global observation~\cite{RND} or joint histories in a centralized way. However, it is inefficient to find structured interaction between agents, which seems too sparse compared with the exponentially growing state space when the number of agents increases. In contrast, if curiosity is defined as the novelty of local observation histories during the decentralized execution, although scalable, it still fails to guide agents to coordinate due to partial observability. Therefore, we find a middle point of centralized curiosity and decentralized curiosity, i.e., utilizing the value factorization of the state-of-the-art multi-agent Q-learning approaches and defining the prediction errors of individual Q-value functions as intrinsic rewards.   
\begin{wrapfigure}{r}{0.5\textwidth}
\centering
\includegraphics[width=\linewidth]{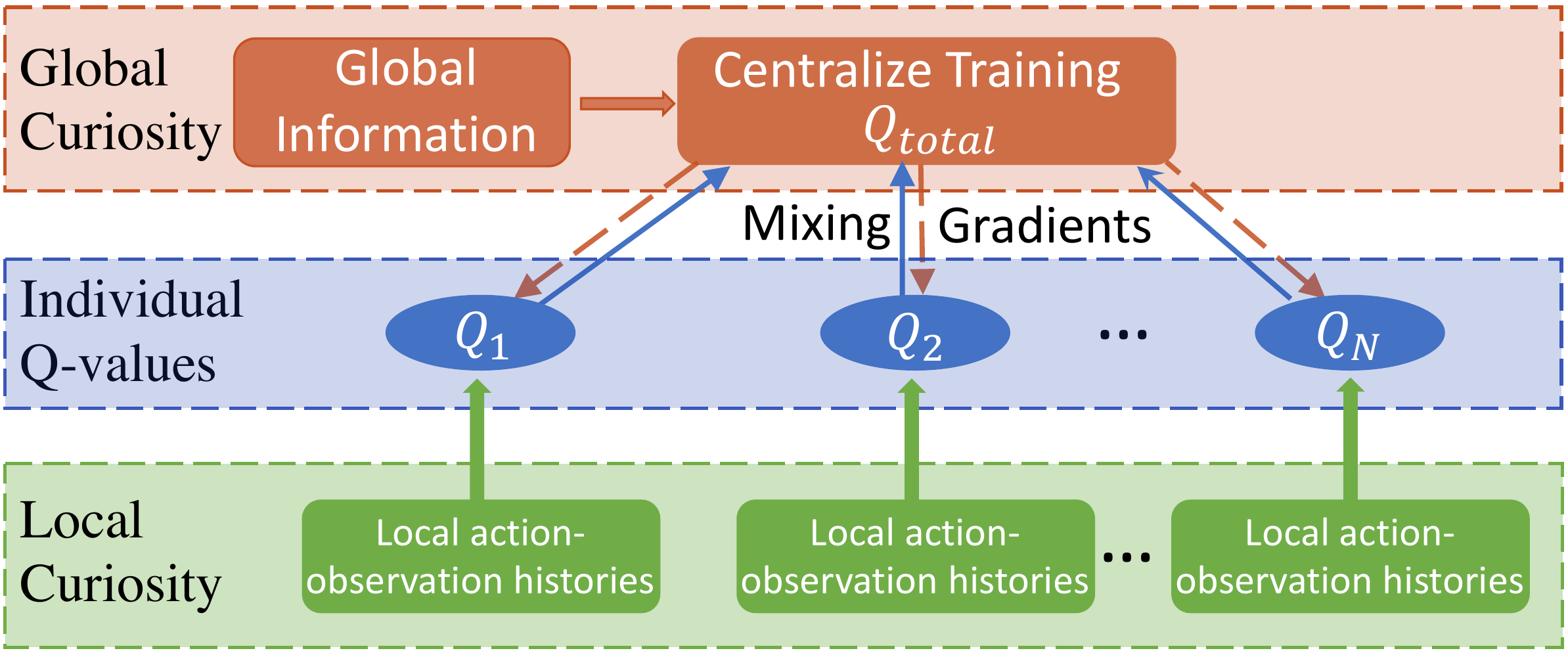}
\caption{CTDE Framework}
\label{Fig:CTDE}
\vspace{-0.1in}
\end{wrapfigure}
The significance of this intrinsic reward is two-fold: 1) it provides a novelty measure of joint observation histories with scalability because individual Q-values are latent embeddings (i.e., an effective state abstraction \citep{li2006towards}) of observation histories in factorized multi-agent Q-learning (e.g., VDN or QPLEX); and 2) as shown in Figure~\ref{Fig:CTDE}, it captures the influence from other agents due to the implicit credit assignment from global reward signal during centralized training~\cite{LVD}, and biases exploration into promising states where strong interdependence may lie between agents. Therefore, with this novel intrinsic reward, our curiosity-driven method enables efficient, diverse, and coordinated exploration for deep multi-agent Q-learning with value factorization.

Besides efficient exploration, another challenge for deep MARL approaches is how to make the best use of experiences collected by the exploration strategy. Prioritized experience replay based on TD errors shows effectiveness in single-agent deep reinforcement learning. However, it does not carry this promise in factorized multi-agent Q-learning, since the projection error induced by value factorization is also fused into the TD error and severally degrades the effectiveness of the TD error as a measure of the usefulness of experiences. To efficiently use promising exploratory experience trajectories, we augment factorized multi-agent reinforcement learning with episodic memory \cite{emdqn,zhu2019episodic}. This memory stores and regularly updates the best returns for explored states. We use the results in the episodic memory to regularize the TD loss, which allows fast latching onto past successful experience trajectories collected by curiosity-driven exploration and greatly improves learning efficiency. Therefore, we call our method Episodic Multi-agent reinforcement learning with Curiosity-driven exploration, called EMC.     

We evaluate EMC in didactic examples, and a broad set of StarCraft II micromanagement benchmark tasks \cite{SMAC}. The didactic examples along with detailed visualization illustrate that our proposed intrinsic reward can guide agents' policies to novel or promising states, thus enabling effectively coordinated exploration. Empirical results on more complicated StarCraft II tasks show that EMC significantly outperforms other multi-agent state-of-the-art baselines. 

\section{Background}
\label{background}
\subsection{Dec-POMDP}
A cooperative multi-agent task can be modelled as a Dec-POMDP \cite{Dec-POMDP}, which is defined by a tuple $G=<\mathcal{I}, \mathcal{S}, \mathcal{A}, P, R, \Omega, O, n, \gamma >$, where $\mathcal{I}$ is the sets of $n$ agents, $\mathcal{S}$ is the global state space, $\mathcal{A}$ is the finite action set, $\gamma \in[0,1)$ is the discount factor. We consider a partially observable setting in a Dec-POMDP, i.e., at each timestep, agent $i \in \mathcal{I}$ only has access to the observation $o_i \in \varOmega$ drawn from the observation function $O(s,i)$. Besides, each agent has an action-observation history $\tau_i\in {\mathcal{T} }\equiv \left(\varOmega\times\mathcal{A}\right)^*\times\varOmega$ and constructs its individual policy to jointly maximize team performance. With each agent $i$ selecting an action $a_i\in \mathcal{A}$, the joint action $\boldsymbol{a}\equiv[a_i]_{i=1}^{n}\in\boldsymbol{\mathcal{A}}\equiv{{\mathcal{A}}^N}$ leads to a shared reward $r=R(s,\boldsymbol{a})$ and the next state $s'$ according to the transition distribution $P(s'|s,\boldsymbol{a})$. The formal objective function is to find a joint policy $\boldsymbol{\pi}$ that maximizes a joint value function $V^{\boldsymbol{\pi}}(s)=\mathbb{E}[\sum_{t=0}^{\infty}\gamma^t r_t|s=s_0,\boldsymbol{\pi}]$, or a joint action-value function $Q^{\pi}(s,\boldsymbol{a})=r(s,\boldsymbol{a})+\gamma\mathbb{E}_{s'}[V^{\boldsymbol{\pi}}(s')]$.
\subsection{Centralized Training With Decentralized Execution (CTDE)}
CTDE is a promising paradigm in deep cooperative multi-agent reinforcement learning~\cite{CTDE2002,Dec-POMDP,CTDE1}, where the local agents execute actions only based on local observation histories, while the policies can be trained in centralized manager which has access to global information. During the training process, the whole team cooperate to find the optimal joint action-value function $Q_{tot}^{*}(s,\boldsymbol{a})=r(s,\boldsymbol{a})+\gamma\mathbb{E}_{s'}[\max_{\boldsymbol{a}'}Q_{tot}^{*}(s',\boldsymbol{a}')]$. 
Due to partial observability, we use $Q_{tot}(\boldsymbol{\tau},\boldsymbol{a};\boldsymbol{\theta})$ instead of $Q_{tot}(s,\boldsymbol{a};\boldsymbol{\theta})$, where $\boldsymbol{\tau}\in \boldsymbol{\mathcal{T} }\equiv \mathcal{T}^N$. Then the Q-value neural network will be trained to minimize the following expected TD-error:
\vspace{-0.05in}
\begin{equation}
    \mathcal{L(\boldsymbol{\theta})}=\mathbb{E}_{\boldsymbol{\tau},\boldsymbol{a},\boldsymbol{r},\boldsymbol{\tau}'\in D}\left[r+\gamma V(\boldsymbol{\tau}';\boldsymbol{\theta}^{-})-Q_{tot}(\boldsymbol{\tau},\boldsymbol{a};\boldsymbol{\theta})\right]^2,
\end{equation}
where $D$ is the replay buffer and $\boldsymbol{\theta}^{-}$ denotes the parameters of the target network, which is periodically updated by $\boldsymbol{\theta}$. And $V(\boldsymbol{\tau}';\boldsymbol{\theta}^{-})$ is the one-step expected future return of the TD target.
Local agents can only obtain local action-observation history and need inference based on individual Q-value functions $Q_i(\tau_i,a_i)$. Therefore, many works have made efforts in finding the factorization structures between joint Q-value functions $Q_{tot}$ and individual Q-functions $Q_i(\tau_i,a_i)$ \cite{VDN,QMIX,qplex}.

\section{Related Work}
\label{relatedwork}
\textbf{Curiosity-driven Exploration}
Curiosity-driven exploration has been well studied in single-agent reinforcement learning. Previous literature  \cite{curiositysummry2,curiositysummry} has provided a good summary in this topic. Recently, curiosity-driven methods have made great progress in deep reinforcement learning. For example, some works use pseudo-state counts to get intrinsic rewards \cite{COUNT,ostrovski2017count,tang2017exploration} instead of count-based methods to get better scalability.  Stadie et al. \cite{stadie2015incentivizing} use prediction errors in the feature space of an auto-encoder to measure the novelty of states and encourage exploration. On the other hand, Mohamed and Rezende \cite{mohamed2015variational} propose to use empowerment, measured by the information gain based on the entropy of actions, as intrinsic rewards for exploring novel states efficiently. Another information-based method \cite{VIME} tries to maximize information gain about the agent’s belief of the environment’s dynamics as an exploration strategy. ICM \cite{ICM} learns an inverse model which predicts the agent’s action given its current and next states and tries to predict the next state in the learned hidden space by current state and action. RND \cite{RND} uses curiosity as intrinsic rewards in a simpler but effective way, which uses a fixed randomly initialized neural network as a representation network and directly predicts the embedding of the next state. Different from these methods, we are the first to propose an advanced curiosity-driven exploration method in MARL setting for diverse and coordinated exploration.

\textbf{Multi-agent Exploration}
Although single-agent exploration is extensively studied and has achieved considerable success, few exploration methods were designed for cooperative MARL. Bargiacchiet al. \cite{bargiacchi2018learning} proposes an exploration method that can only be used in repeated single-stage problems. Jaques et al. \cite{SocialInfluence} defines intrinsic reward by ``social influence'' to encourage agents to choose actions that can influence other agents' actions. Iqbal and Sha \cite{iqbal2019coordinated} uses various simple exploration methods to learn simultaneously and then put the samples of every method in a shared buffer to achieve the coordinated exploration. Wang et al. \cite{EDTI} use mutual information (MI) to capture the interdependence of the rewards and transitions between agents. MAVEN \cite{MAVEN} is the state-of-the-art exploration method in MARL that uses a hierarchical policy to produce a shared latent variable and learns several state-action value functions for each agent. These works, although important, still face the challenge of achieving scalable and effective multi-agent exploration.

\textbf{Episodic Control}
Our work is also related to episodic control reinforcement learning, which is usually adopted in single-agent settings for better sample efficiency. Previous works propose to use episodic memory in near-deterministic environment \cite{lengyel2008hippocampal,MFEC,pritzel2017neural,hansen2018fast}. Model-free episodic control~\cite{MFEC} uses a completely non-parametric table to keep the best Q-values of state-action pair in a tabular-based memory and uses a k-nearest-neighbors fashion to find the sequence of actions that so far yielded the highest return from a given start state in the memory. 
Recently, several extensions have been proposed to integrate episodic control with parametric DQN. Gershman and Daw \cite{gershman2017reinforcement} uses episodic memory to retrieve samples and then average future returns to approximate the action values. EMDQN~\cite{emdqn} uses a fixed random matrix as a representation function and uses the projection of states as keys to store the information of episodic memory into a non-parametric model. Using the episodic-memory based target as a regularization term to guide the training process, the performance of EMDQN is significantly improved compared with the original DQN. Despite the fruitful progress made in single-agent episodic reinforcement learning, few works study episodic control in a multi-agent setting. To the best of our knowledge, we are the first to utilize the mechanism of episodic control in deep multi-agent reinforcement learning.

\section{Episodic Multi-agent Reinforcement Learning with Curiosity-Driven Exploration}
\label{method}
In this section, we introduce EMC, a novel episodic multi-agent exploration framework. EMC takes prediction errors of individual Q-value functions as intrinsic rewards for guiding the diverse and coordinated exploration. After collecting informative experience, we leverage an episodic memory to memorize the highly rewarding sequences and use it as the reference of a one-step TD target to boost multi-agent Q-learning. 
First, we analyze the motivations for predicting individual Q-values, then we introduce the curiosity module for exploration. Finally, we describe how to utilize episodic memory to boost training. 

\subsection{Curiosity-Driven Exploration by Predicting Individual Q-values}
\label{Curiosity-Driven Exploration}
\begin{figure}[ht]
\centering
		\centerline{\includegraphics[width=0.98\linewidth]{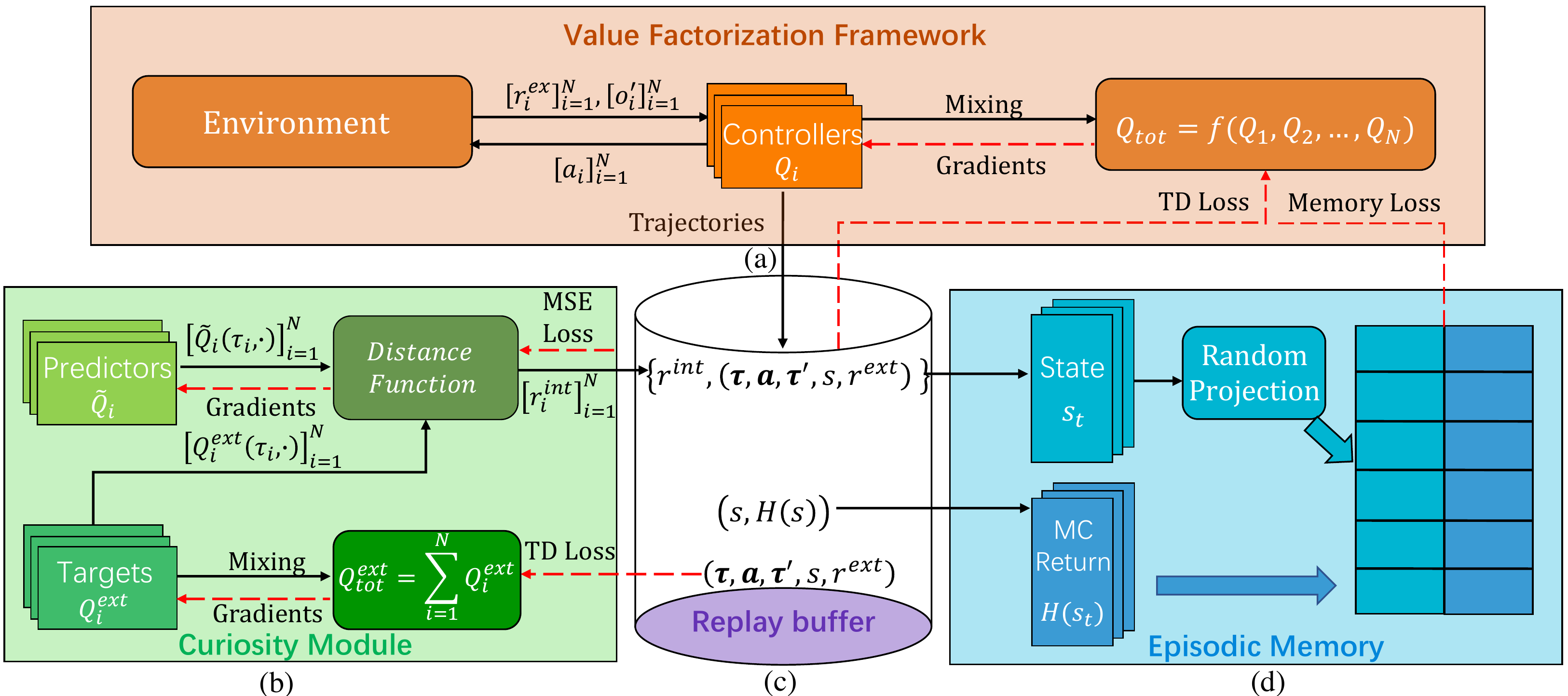}}
		\caption{An overview of EMC's framework}
		\label{Frameworkpic}
\end{figure}

As shown in Figure~\ref{Frameworkpic}, in the paradigm of CDTE, local agents make decisions based on individual Q-value functions, which take local observation histories as inputs, and are updated by the centralized module which has access to global information for training. The key insight is that, different from single-agent cases, individual Q-value functions in MARL are used for both decision-making and embedding historical observations. Furthermore, due to implicit credit assignment by global reward signal during centralized training, individual Q-value functions $Q_i(\tau_i,\cdot)$ are influenced by environment as well as other agents' behaviors. More concretely, it has been proved by~Wang et al.~\cite{LVD} that, when the joint Q-function $Q_{tot}$ is factorized into linear combination of individual Q-functions $Q_i$, i.e., $Q_{tot}^{(t+1)}(\bm{\tau}, \bm{a})=\sum_{i=1}^N Q_i^{(t+1)}(\tau_{i}, a_{i})$, then $Q_i^{(t+1)}(\tau_{i}, a_{i})$ has the following closed-form solution:
\vspace{-0.1in}
\begin{equation}
	\begin{aligned}\label{eq:credit_assignment}
Q_i^{(t+1)}(\tau_{i}, a_{i})=&	\underbrace{\mathop{\mathbb{E}}_{(\tau'_{-i},a'_{-i})\sim p_D(\cdot|{\tau_i})}\left[y^{(t)}\left(\tau_i\oplus\tau'_{-i}, a_i \oplus a'_{-i}\right)\right]}_{\text{evaluation of the individual action $a_i$}} \\
	&- \frac{n-1}{n}\underbrace{\mathop{\mathbb{E}}_{{\bm\tau}',\bm{a}'\sim p_D(\cdot|\Lambda^{-1}({\tau_i}))}\left[ y^{(t)}\left({\bm\tau}',\bm{a}'\right)\right]}_{\text{counterfactual baseline}} + w_i(\tau_i),
	\end{aligned}
\end{equation}
where $y^{(t)}({\bm\tau},\bm{a})= r+\gamma\mathbb{E}_{{\bm\tau}'}\left[\max_{\bm{a}'} Q^{(t)}_{tot}\left({\bm\tau}',\bm{a}'\right)\right]$ denotes the expected one-step TD target, and $p_D(\cdot|\tau_i)$ denotes the conditional empirical probability of $\tau_i$ in the given dataset $D$. The notation $\tau_i \oplus \tau'_{-i} $ denotes $\langle \tau_1', \dots, \tau_{i-1}', \tau_i, \tau_{i+1}', \dots, \tau_n' \rangle$, and $\tau'_{-i}$ denotes the elements of all agents except for agent $i$. $\Lambda^{-1}(\tau_i)$ denotes the set of trajectory histories that may share the same latent-state trajectory as $\tau_i$. The residue term $\bm{w}\equiv\left[w_i\right]_{i=1}^n$ is an arbitrary function satisfying $\forall \bm\tau\in\bm{\Gamma}$, $\sum_{i=1}^n w_i(\tau_i) = 0$.

Eq.~(\ref{eq:credit_assignment}) shows that by \textit{linear value factorization}, the individual Q-value $Q_i(\tau_i,a_i)$ is not only decided by local observation histories but also influenced by other agents' action-observation histories. Thus predicting $Q_i$ can capture both the novelty of states and the interaction between agents and lead agents to explore promising states. Motivated by this insight, in this paper, we use a linear value factorization module separate from the inference module to learn the individual value function $Q_i$, and use the prediction errors of $Q_i$ as intrinsic rewards to guide exploration. In this paper, we define the prediction errors of individual Q-values as \textit{curiosity} and propose our curiosity-driven exploration module as follows.

Figure~\ref{Frameworkpic}b demonstrates the \textit{Curiosity Module}, separated from the inference module~(Figure~\ref{Frameworkpic}a). The curiosity module consists of four components:  (i) The centralized training part with linear value factorization, which shares the same implementation as VDN \cite{VDN}, but only trained with extrinsic rewards $r^{ext}$ from the environment; (ii) the \textit{Target} for prediction, i.e., the corresponding individual Q-values $Q_i^{ext}$, represented by a recurrent Q-network; (iii) \textit{Predictor}~$\widetilde{Q}_{i}(\tau_i)$, which is used for predicting $Q_i^{ext}$ and shares the same network architecture as \textit{Target}~$Q_i^{ext}$; and (iv) \textit{Distance Function}, which measures the distance between $Q_i^{ext}$ and $\widetilde{Q}_{i}$, e.g., $L_2$ distance. The predictors are  trained by minimizing the Mean Squared Error (MSE) of the distance in an end-to-end manner. For stable training, we use the soft-update target~\cite{softupdate} of $Q_i^{ext}$ to smooth the outputs of the targets. In general, (ii) is trained with (i) and outputs individual Q-values , while (iii) is trained with (ii) and (iv), and aims to predict the soft-update target of individual Q-values. Motivated by the implicit credit assignment of linear value factorization (Eq.~(\ref{eq:credit_assignment})), the curiosity module predicts the individual Q-values $\left[Q^{ext}_i\right]_{i=1}^n$ in linear factorization, i.e., $Q^{ext}_{tot}=\sum_{i=1}^{N}Q_i^{ext}$. 
Then the curiosity-driven intrinsic reward is generated by the following equation: 
\vspace{-0.1in}
\begin{equation}
r^{int}=\frac{1}{N}\sum_{i=1}^{N}{\left \|\widetilde{Q}_{i}(\tau_i,\cdot) - {Q}^{ext}_i(\tau_i,\cdot)\right \|}_2\\,
\label{intrinsic_reward}
\end{equation}
This intrinsic reward is used for the centralized training of the inference module, as shown in Figure~\ref{Frameworkpic}a:
\begin{equation}
    \mathcal{L_{\text{inference}}(\boldsymbol{\theta})}=\mathbb{E}_{\boldsymbol{\tau},\boldsymbol{a},\boldsymbol{r},\boldsymbol{\tau}'\in D}\left[
    \left(y(\boldsymbol{\tau},\boldsymbol{a})-Q_{tot}(\boldsymbol{\tau},\boldsymbol{a};\boldsymbol{\theta})\right)^2\right],
\end{equation}
where $y(\boldsymbol{\tau},\boldsymbol{a})=r^{ext}+\beta r^{int}+\gamma \max_{\bm{a}'} Q_{tot}\left({\bm\tau}',\bm{a}';\boldsymbol{\theta}^{-})\right)$, denoting one step TD target of the inference module, and $\beta$ is the weight term of the intrinsic reward. 
We use a separate training model for inference~(Figure~\ref{Frameworkpic}a) to avoid the accumulation of projection errors of $Q_i$ during training. 

The independence of inference module leads to another advantage, that EMC's architecture can be adopted into many value-factorization-based multi-agent algorithms which utilize the CDTE paradigm, i.e., the general function $f$ in Figure~\ref{Frameworkpic}a can indicate specific (linear, monotonic and IGM) value factorization structures in VDN \cite{VDN}, QMIX \cite{QMIX}, and QPLEX \cite{qplex}, respectively. In this paper, we utilize these state-of-the-art algorithms for the inference module. With this curiosity-driven bias plugged into ordinary MARL algorithms, EMC will achieve efficient, diverse and coordinated exploration.

\subsection{Episodic Memory}
Equipped with efficient exploration ability, another challenge is how to make the best use of good trajectories collected by exploration effectively.
Recently, episodic control has been widely studied in single-agent reinforcement learning \cite{emdqn,zhu2019episodic}, which can replay the highly rewarding sequences, thus boosting training. 
Inspired by this framework, we generalize single-agent episodic control to propose a multi-agent episodic memory, which records the best memorized Monte-Carlo return in the episode, and provide a memory target $H$ as a reference to regularize the ordinary one-step inference TD target estimation in the inference module (Figure~\ref{Frameworkpic}a):
\begin{equation}
\mathcal{L_{\text{memory}}(\bm{\theta})}=
\mathbb{E}_{\boldsymbol{\tau},\boldsymbol{a},\boldsymbol{r},\boldsymbol{\tau}'\in D}\left[
    \left(H-Q_{tot}(\bm{\tau},\boldsymbol{a};\boldsymbol{\theta})\right)^2
    \right].
\end{equation}
 However, different from the single-agent episodic control, the action space of MARL exponentially grows as the number of agents increases, and partial observability also limits the information of individual value functions. Thus, we maintain our episodic memory by storing the state-value function on the global state space and utilizing the global information during the centralized training process under the CTDE paradigm. Figure~\ref{Frameworkpic}d shows the architecture of the \textit{Episodic Memory}. We keep a memory table $M$ to record the maximum remembered return of the current state, and use a fixed random matrix drawn from Gaussian distribution as a representation function to project states into low-dimensional vectors $\phi(s):S\rightarrow{\mathbb{R}^k}$, which are used as keys to look up corresponding global state value function $H(\phi(s_t))$. When our exploration method collects a new trajectory, we update our memory table $M$ as follows:
\begin{equation}
H(\phi(s_t))=
\begin{cases}
\max\{H(\phi(\hat{s}_t)),R_t(s_t, \bm{a}_t)\} & {\textit{if } \|\phi(\hat{s}_t) -\phi(s_t)\|_2<\delta}\\
R_t(s_t, \bm{a}_t)& otherwise
\end{cases},
\end{equation}
where $\phi(\hat{s}_t)$ is $\phi(s_t)$'s nearest neighbor in the memory $M$, $\delta$ is a threshold, and $R(s_t, \bm{a}_t)$ represents the future return when agents taking joint action $\bm{a}_t$ under global state $s_t$ at the $t$-th timestep in a new episode. In our implementation, $\phi(s_t)\in M$ is indeed evaluated approximately based on the embedding distance. Specifically, when the key of the state $\phi(s_t)$ is close enough to one key in the memory, we assume that $\phi(s_t)\in M$ and find the best memorized Monte-Carlo return correspondingly. Otherwise, we think $\phi(s_t)\notin M$ and record the state's return into the memory. Leveraging the episodic memory, we can directly obtain the maximum remembered return of the current state, and use the one-step TD memory target $H$ as a reference to regularize learning:
\begin{equation}
{H}(\phi(s_t),\boldsymbol{a}_t)=r_t(s_t,\boldsymbol{a}_t)+\gamma H(\phi(s_{t+1})).
\label{onestepv}
\end{equation}
Thus, the new objective function for the inference module is:
\begin{equation}
\begin{aligned}
\mathcal{L_{\text{total}}(\bm{\theta})}&=\mathcal{L_{\text{inference}}(\bm{\theta})}+\lambda \mathcal{L_{\text{memory}}(\bm{\theta})}\\
&=\mathbb{E}_{\boldsymbol{\tau},\boldsymbol{a},\boldsymbol{r},\boldsymbol{\tau}'\in D}\left[
    \left(y(\bm{\tau},\boldsymbol{a})-Q_{tot}(\bm{\tau},\boldsymbol{a};\boldsymbol{\theta})\right)^2+
    \lambda\left(H\left(\phi(s_t),\boldsymbol{a}_t\right)-Q_{tot}(\bm{\tau},\boldsymbol{a};\boldsymbol{\theta})\right)^2
    \right],
    \end{aligned}
\end{equation}
where $\lambda$ is the weighting term to balance the effect of episodic memory's reference.
Using the maximum return from the episodic memory to propagate rewards, we can compensate for the disadvantage of slow learning induced by the original one-step reward update and improve sample efficiency.

\section{Experiments}

In this section, we will conduct a large set of empirical experiments for answering the following questions: (1) Is exploration by predicting individual Q-value functions better than exploration by decentralized curiosity or global curiosity~(Section~\ref{toygamesection})? (2) Can our method perform efficient coordinated exploration in challenging multi-agent tasks~(Section~\ref{Predator Prey}-\ref{smac})? (3) If so, what role does each key component play for the superior performance~(Section~\ref{ablation})? (4) Why do we choose to predict target $Q_i^{ext}$ for generating intrinsic rewards rather than other choices~(Section~\ref{ablation})?
We will propose several didactic examples and demonstrate the advantage of our method in coordinated exploration, and evaluate our method on the StarCraft II micromanagement (SMAC) benchmark~\cite{SMAC} compared with existing state-of-the-art multi-agent reinforcement learning (MARL) algorithms: QPLEX~\cite{qplex}, Weighted-QMIX~\cite{WQMIX}, QTRAN~\cite{QTRAN}, QMIX~\cite{QMIX}, VDN~\cite{VDN}, RODE~\cite{RODE}, and MAVEN~\cite{MAVEN}.
\subsection{Didactic Example}
\label{toygamesection}

\begin{wrapfigure}{r}{0.25\textwidth}
\centering

\includegraphics[width=\linewidth]{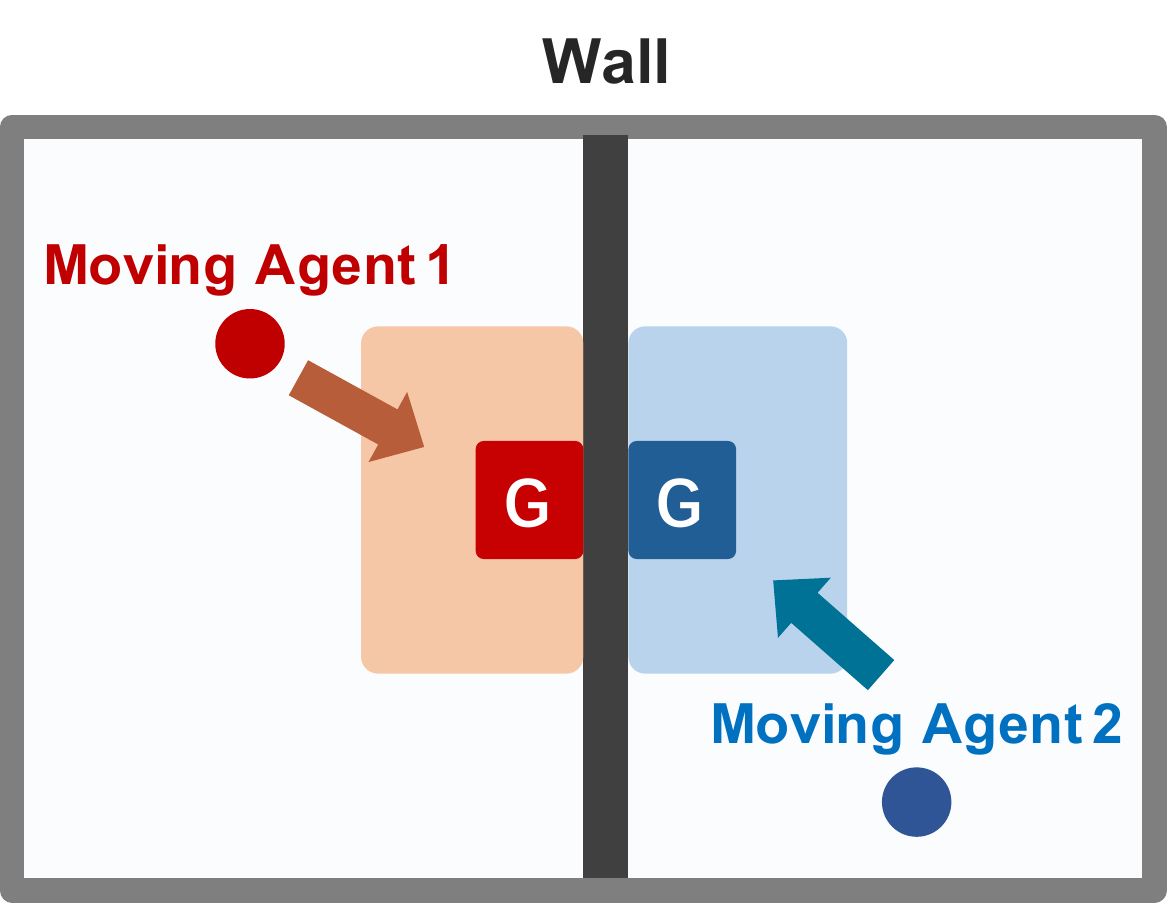}
\caption{Coordinated Toygame}
\label{Fig:gridworld}
\vspace{-0.2in}
\end{wrapfigure}

Figure~\ref{Fig:gridworld} shows an $11\times 12$ grid world game that requires coordinated exploration. The blue agent and the red agent can choose one of the five actions: \textit{[up, down, left, right, stay]} at each time step. The wall shown in the picture isolates the two agents, and one agent cannot be observed by the other until it gets into the shaded area.
The two agents will receive a positive global reward $r=10$ if and only if they arrive at the corresponding goal grid~(referred to the character $G$ in Figure~\ref{Fig:gridworld}) at the same time. If only one arrives, the incoordination will be punished by a negative reward $-p$. 

\begin{figure}[ht]
\vspace{-0.2in}
\centering
		\centerline{\includegraphics[width=0.99\linewidth]{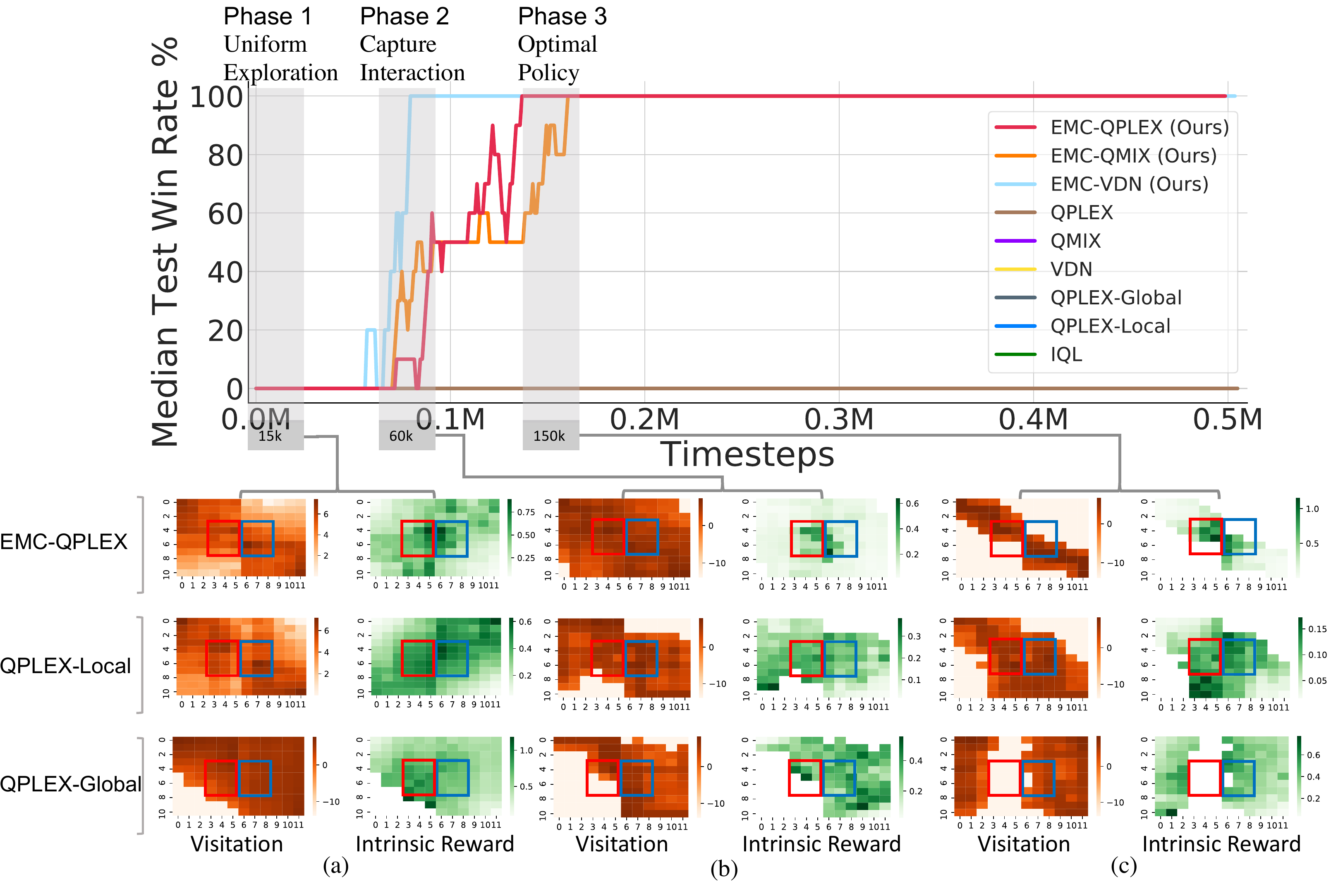}}
		\caption{The heat map of gridworld game.}
		\label{gridworld_visualization}
\vspace{-0.1in}
\end{figure}

To evaluate the effectiveness of our curiosity-driven exploration, we implement our method into QPLEX, QMIX, and VDN (denoted as \textit{EMC-QPLEX}, \textit{EMC-QMIX}, and \textit{EMC-VDN}, respectively) and test them in this toy game compared with the state-of-the-art MARL algorithms: VDN, IQL, QMIX, and QPLEX. Moreover, to demonstrate the motivation of predicting individual Q-functions, we add two more baselines: QPLEX with the prediction error of global state as intrinsic rewards~(denoted as \textit{QPLEX-Global}), and QPLEX with the prediction error of local joint histories as intrinsic rewards (denoted as \textit{QPLEX-Local}). Both of them use a fixed network to project the inputs into latent embedding, then predict the latent embedding to generate intrinsic reward, just like the Random Network Distillation~(RND)~\cite{RND}. 
 We test different punishment degrees, i.e., different $p$s (which are deferred to Appendix C), and the results show QPLEX-Global and QPLEX-Local are effective enough for exploration when $p$ is relatively small. However, as $p$ increases, the task becomes more challenging since it requires sufficient and coordinated exploration.
In Figure~\ref{gridworld_visualization}, we show the median test win rate of all methods over $6$ random seeds when $p=2$, and only our methods can learn the optimal policy and win the game, while other methods failed.

To understand this result better, we have made several visualisations to demonstrate our advantage in coordinated exploration.
Figure~\ref{gridworld_visualization} shows the heatmaps of visitation and intrinsic reward by EMC-QPLEX,  QPLEX-Global, and QPLEX-Local. During the early stage of training, all methods uniformly explore the whole area~(Figure~\ref{gridworld_visualization}a). 
As the exploration progresses, the global curiosity ~(QPLEX-Global) encourages agents to visit all configurations without bias, which is inefficient and fail to leverage the potential locality influence between agents~(Figure~\ref{gridworld_visualization}b), resulting in extrinsic rewards beginning to dominate the behaviors~(Figure~\ref{gridworld_visualization}c). On the other hand, the visitation heatmap of QPLEX-Local shows the decentralized curiosity encourages agents to explore around the goal grid, but it cannot promise to encourage agents to coordinate and gain the reward due to the partial observability in decentralized execution. In contrast, the heatmap of intrinsic reward for EMC-QPLEX shows that predicting individual Q-values will bias exploration into areas where individual Q-values are more dynamic due to the potential correlation between agents. Therefore, QPLEX-Local and QPLEX-Global both fail in this task (Figure~\ref{gridworld_visualization}c), while our methods are able to find the optimal policy. This didactic example shows the global curiosity or local curiosity may fail to handle complex tasks where coordinated exploration needs to be addressed. While since individual Q-values $Q_i$ are the embeddings of historical observations, and are dynamically updated by the backpropagation of the global reward signal gained through cooperation during centralized training. Thus $Q_i$ can implicitly reflect the influence from the environment and other agents, and predicting $Q_i$ can capture valuable and spare interactions among agents and bias exploration into new or promising states. 

\vspace{-0.1in}
\subsection{Predator Prey}
\label{Predator Prey}
\begin{wrapfigure}{r}{0.4\textwidth}
\centering
\vspace{-0.1in}
\includegraphics[width=\linewidth]{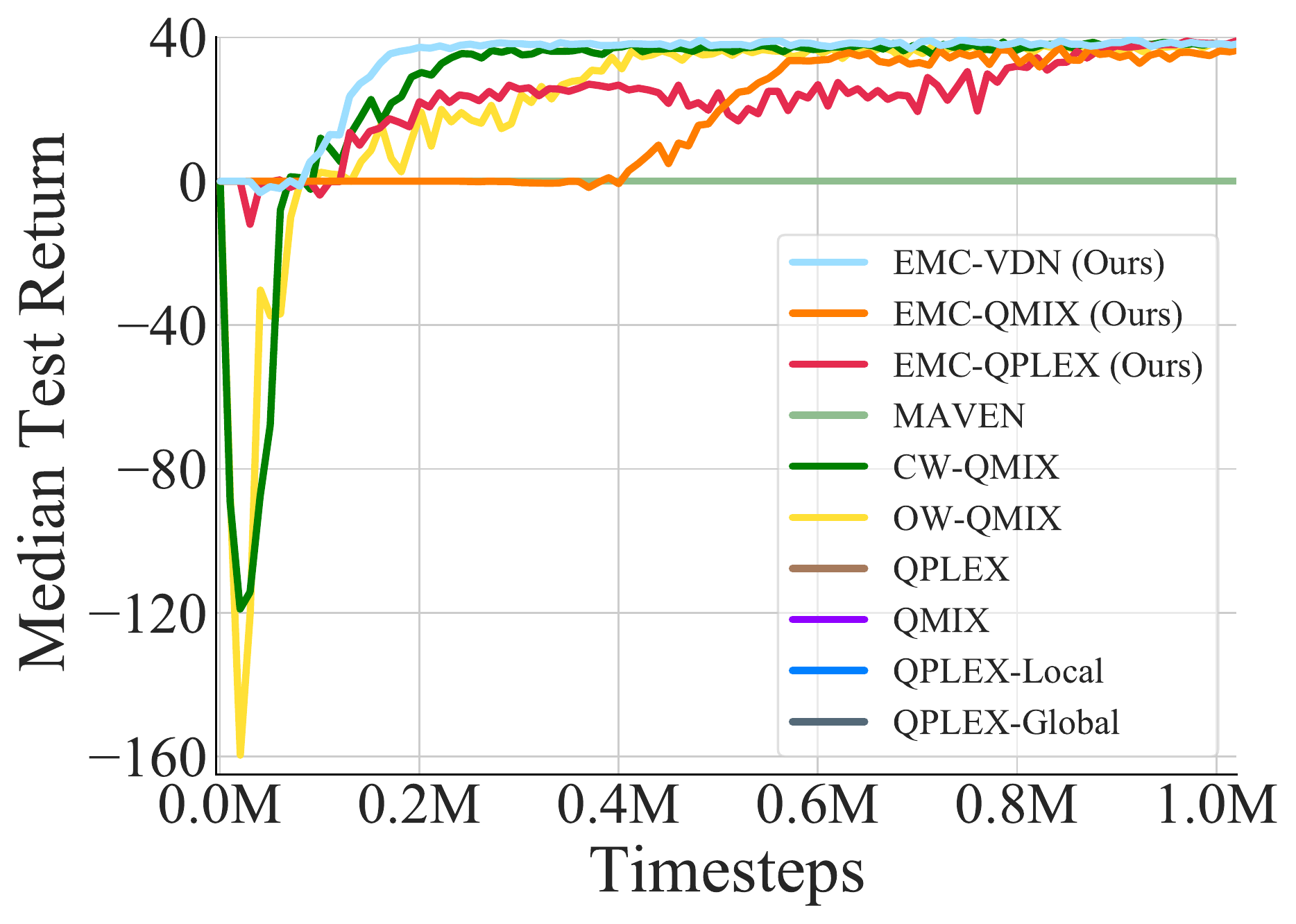}
\caption{The performance of Predator Prey.}
\label{Fig:stag}

\end{wrapfigure}
Predator-Prey is a partially-observable multi-agent coordinated game with miscoordination penalties used by WQMIX \cite{WQMIX}.
As shown in Figure~\ref{Fig:stag}, since extensive exploration is needed to jump out of the local optima, WQMIX is the only baseline algorithm to find the optimal policy, due to its shaped data distribution which can be seen as a type of exploration. Other state-of-the-art multi-agent Q-learning algorithms, such as QPLEX and QMIX, fail to solve this task. For  MAVEN, QPLEC-Global and QPLEX-Local, although equipped with improved exploration ability, they still failed to address coordination due to uniform exploration nature or partial observability. However, plugged with EMC, EMC-VDN, EMC-QMIX, and EMC-QPLEX can guarantee coordinated exploration effectively and achieve good performance.

\vspace{-0.1in}
\subsection{StarCraftII Micromanagement (SMAC) Benchmark}
\label{smac}

\begin{wrapfigure}{r}{0.4\textwidth}
\centering
\vspace{-0.3in}
\includegraphics[width=\linewidth]{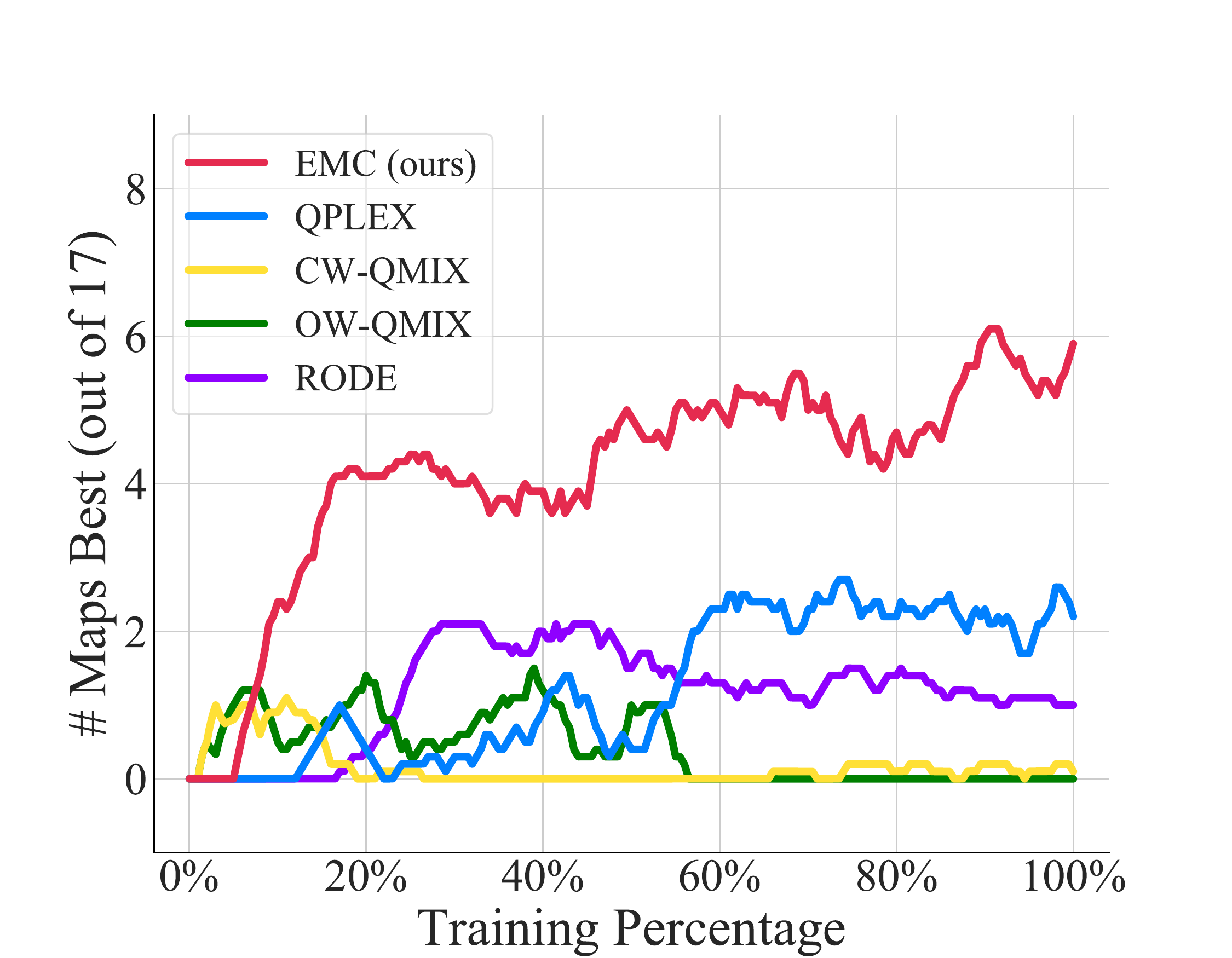}
\caption{The number of scenarios in which the algorithm's median test win rate is the highest by as least 1/32.}
\label{Fig:SmacExpOverall}
\vspace{-0.25in}
\end{wrapfigure}
StarCraft II Micromanagement (SMAC) is a popular benchmark in MARL \cite{VDN,QMIX,qplex,RODE,WQMIX}.  We conduct experiments in 17 benchmark tasks of StarCraft II, which contains 14 popular tasks proposed by SMAC \cite{SMAC} and three more super hard cooperative tasks proposed by QPLEX~\cite{qplex}. In the micromanagement scenarios, each unit is controlled by an independent agent that must act based on its own local observation, and the enemy units are controlled by a built-in AI. 

For evaluation, we compare EMC with the state-of-the-art algorithms: RODE \cite{RODE}, QPLEX \cite{qplex}, MAVEN \cite{MAVEN}, and the two variants of QMIX \cite{QMIX}: CW-QMIX and OW-QMIX \cite{WQMIX}. All experimental results are illustrated with the median performance and 25-75\% percentiles.
Figure~\ref{Fig:SmacExpOverall} shows the overall performance of the tested algorithms in all these $17$ maps. Due to the effective exploration with episodic memory which can efficiently use promising exploratory experience trajectories, EMC is the best performer on up to 6 tasks, underperforms on just three tasks, and ties for the best performer on the rest tasks. 

\begin{figure*}
\vspace{-0.2in}
\begin{minipage}[t]{\linewidth}
\centering
\includegraphics[width=0.95\textwidth]{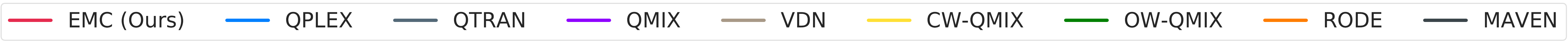}
\end{minipage}%

\centering
\subfigure[corridor]{
\begin{minipage}[t]{0.32\linewidth}
\centering
\includegraphics[width=1.8in]{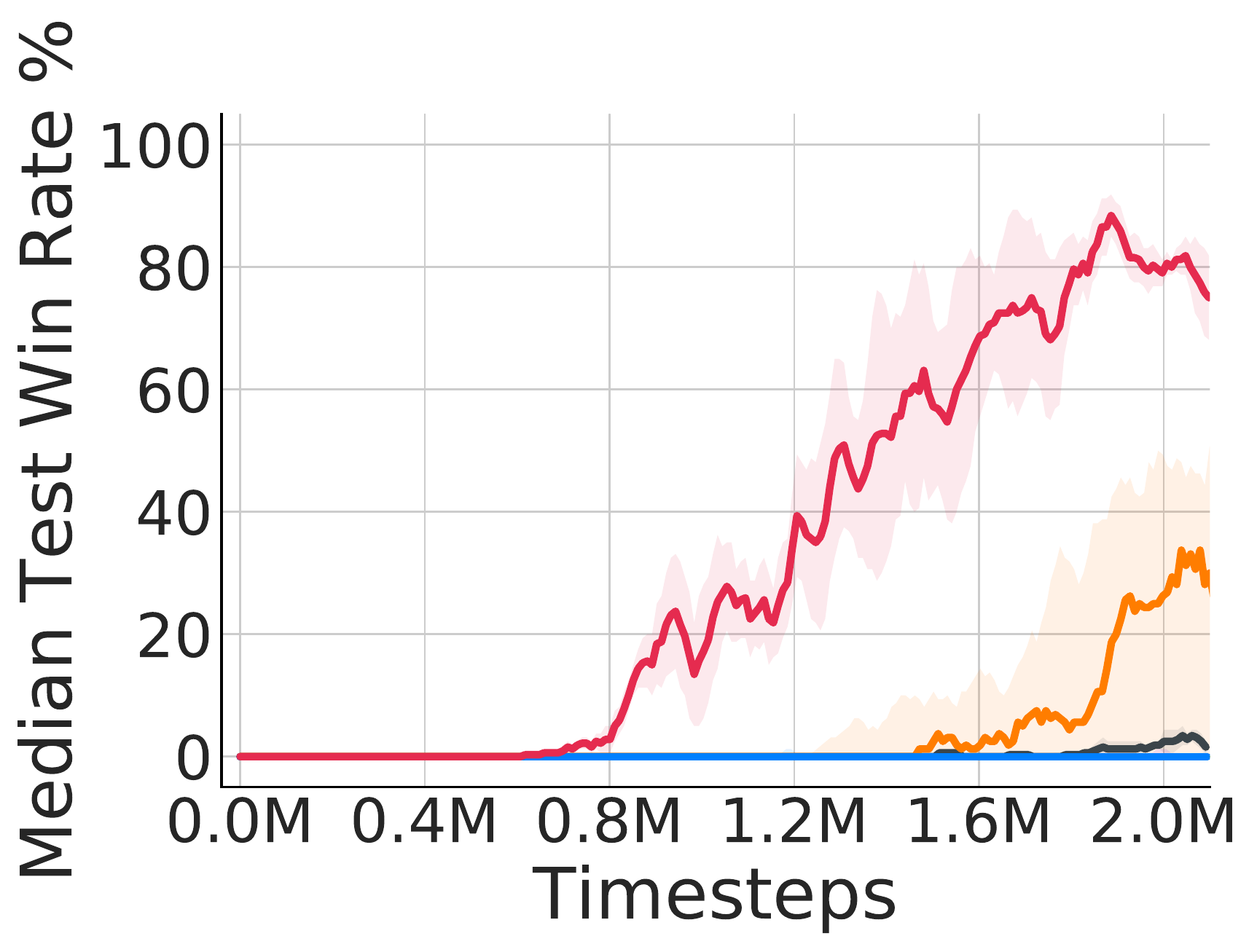}
\end{minipage}%
}%
\subfigure[3s5z\_vs\_3s6z]{
\begin{minipage}[t]{0.32\linewidth}
\centering
\includegraphics[width=1.8in]{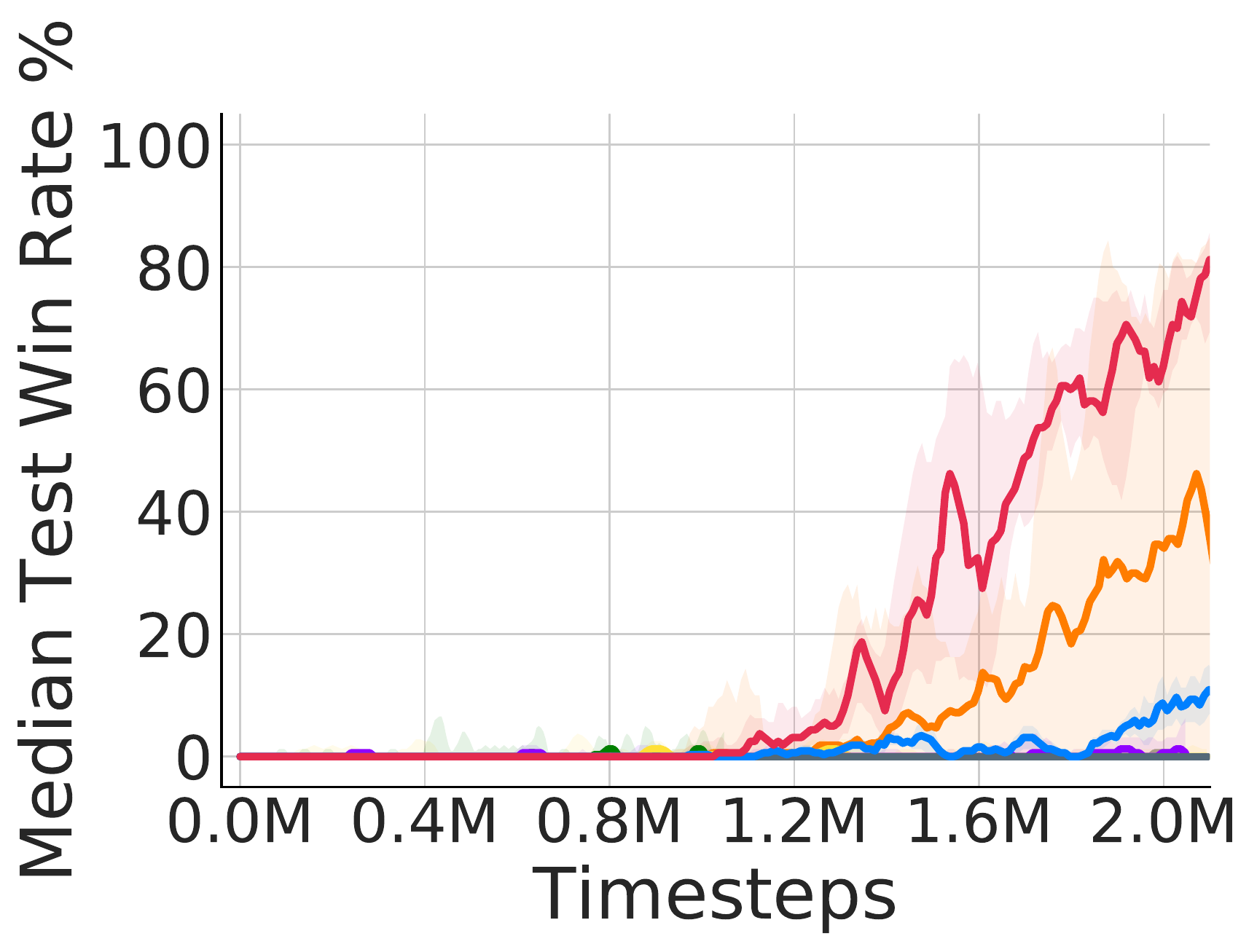}
\end{minipage}%
}%
\subfigure[6h\_vs\_8z]{
\begin{minipage}[t]{0.32\linewidth}
\centering
\includegraphics[width=1.8in]{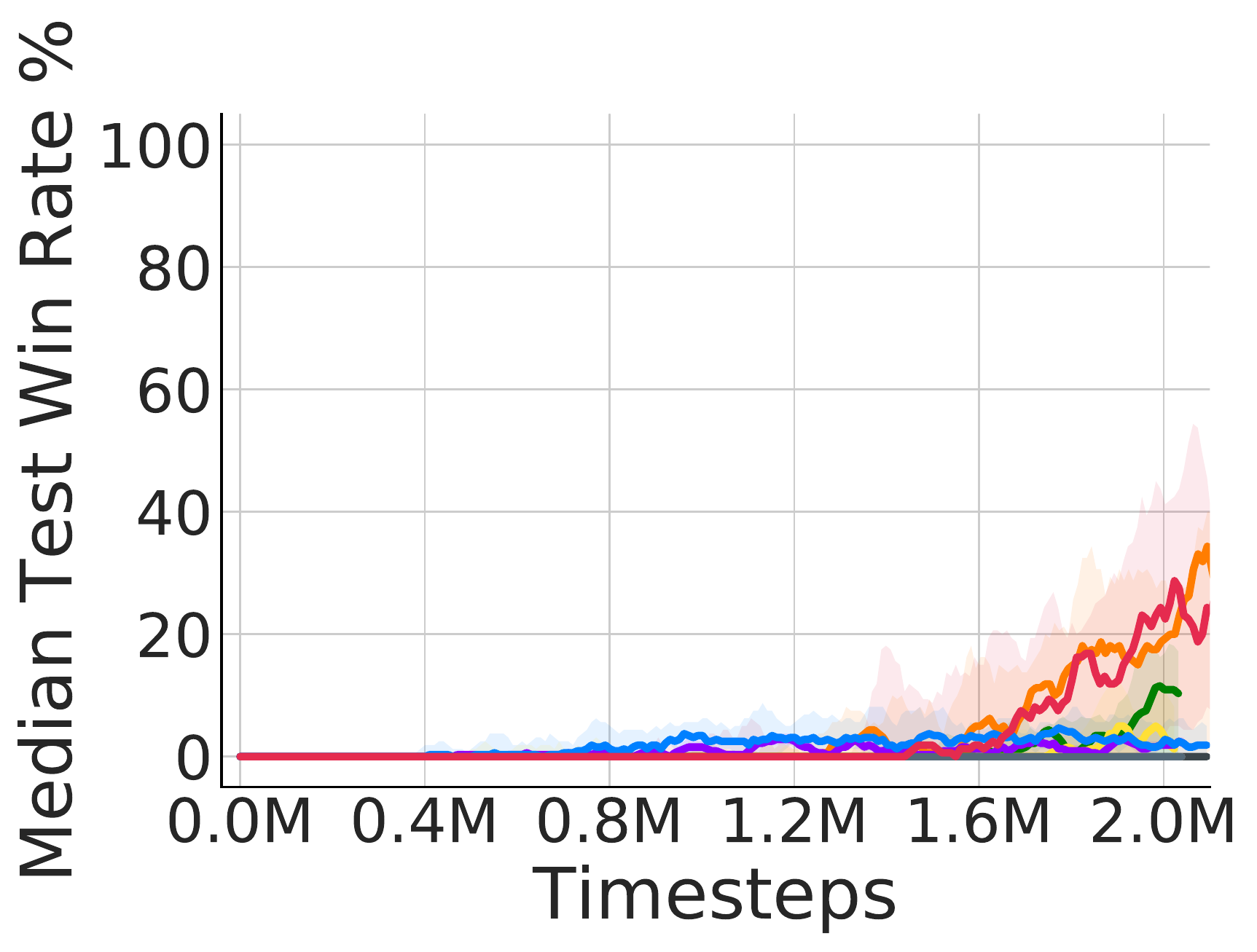}
\end{minipage}%
}%

\subfigure[1c3s8z\_vs\_1c3s9z]{
\begin{minipage}[t]{0.32\linewidth}
\centering
\includegraphics[width=1.8in]{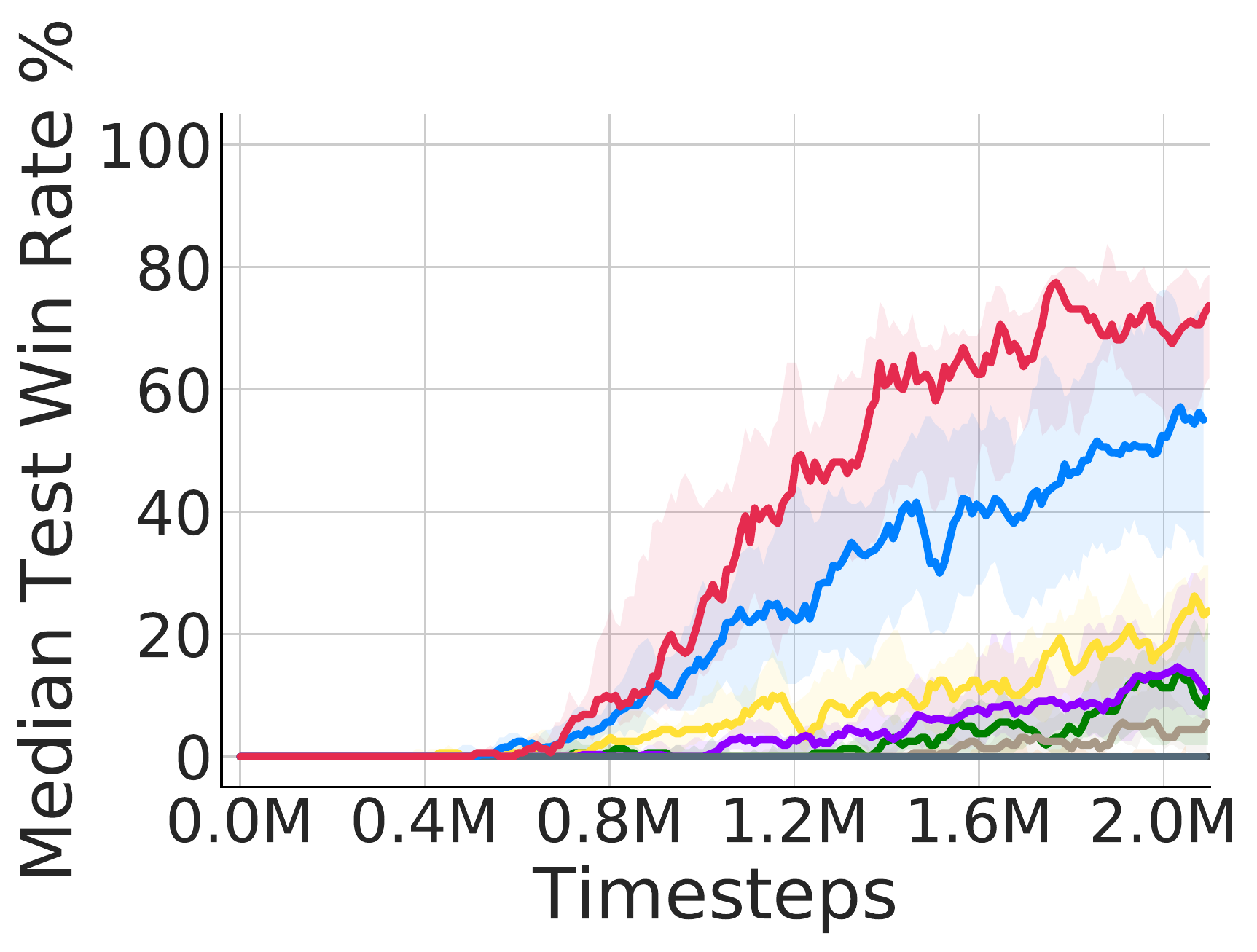}
\end{minipage}%
}%
\subfigure[5s10z]{
\begin{minipage}[t]{0.32\linewidth}
\centering
\includegraphics[width=1.8in]{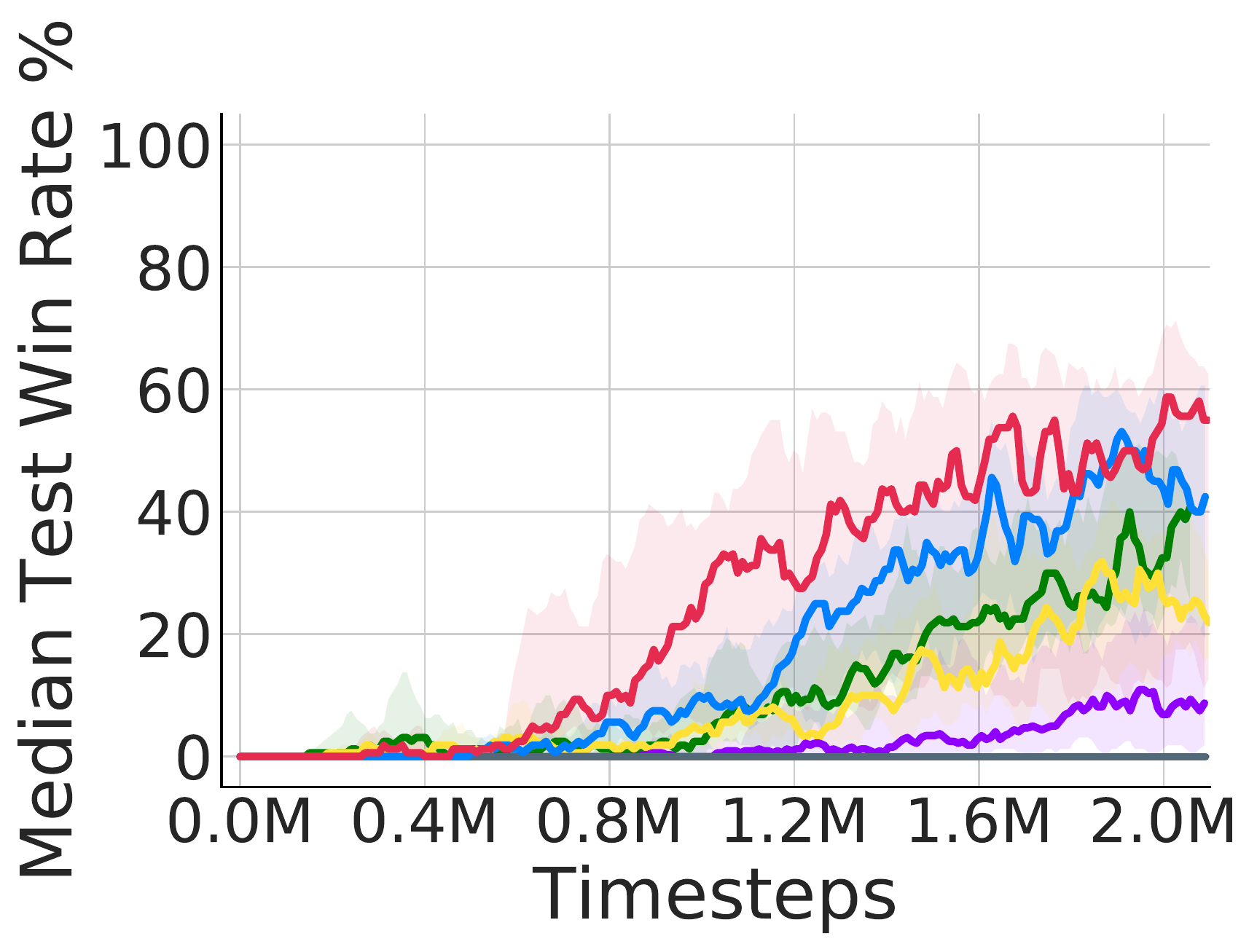}
\end{minipage}%
}%
\subfigure[7s7z]{
\begin{minipage}[t]{0.32\linewidth}
\centering
\includegraphics[width=1.8in]{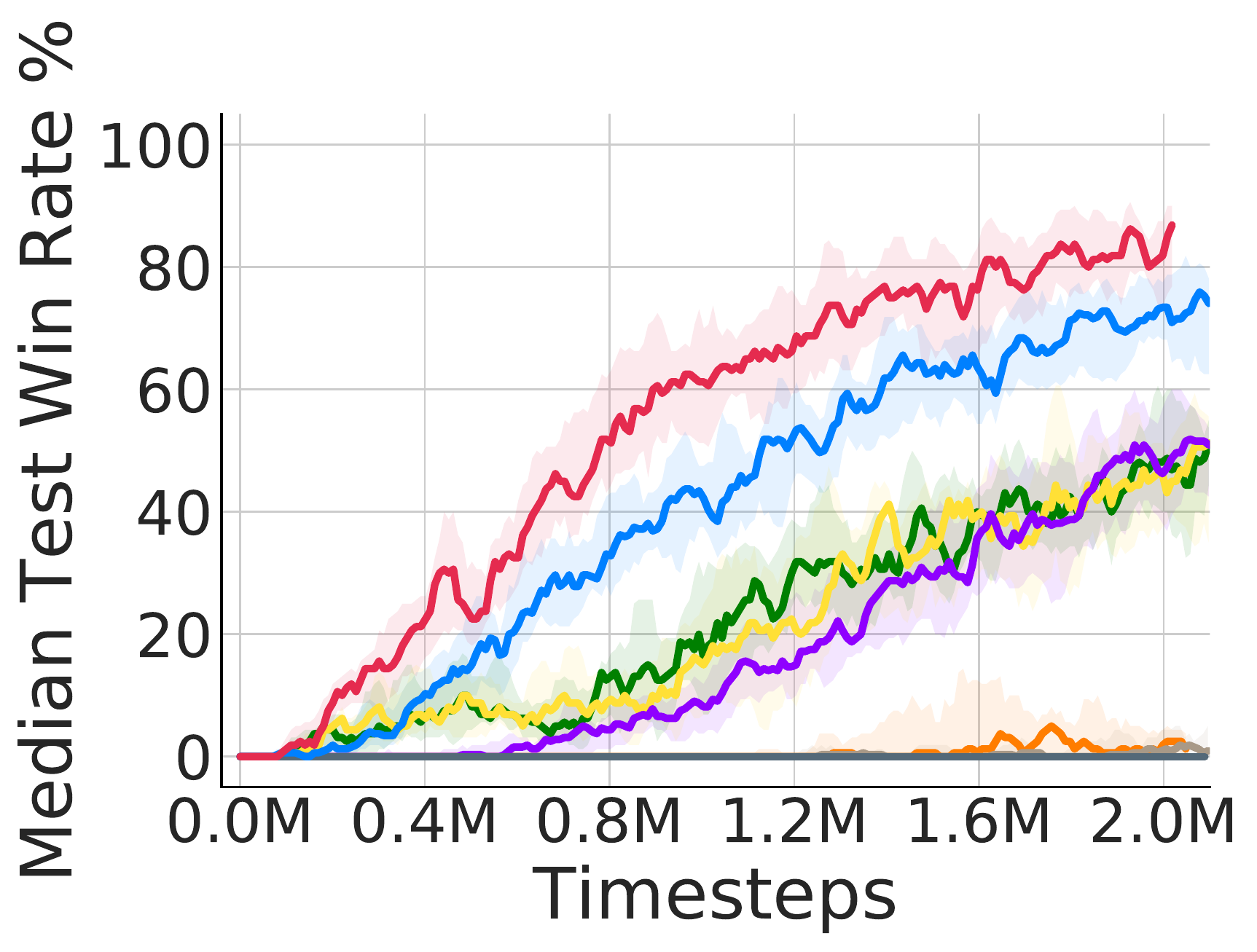}
\end{minipage}
}%
\centering
\caption{Results of super hard maps in SMAC.}
\vspace{-0.2in}
\label{Fig:SmacExpHard}
\end{figure*}

The advantages of our algorithm can be mainly illustrated by the results of the six hard maps which need sufficient exploration shown in Figure \ref{Fig:SmacExpHard}. The three maps in the first row are super hard, and solving them needs efficient, diverse and coordinated exploration. Thus, we can find that the EMC algorithm significantly outperforms other algorithms in \textit{corridor} and \textit{3s5z\_vs\_3s6z}, and also achieves the best performance (equal to RODE) in \textit{6h\_vs\_8z}. To the best of our knowledge, this 
will be the state-of-the-art results in \textit{corridor} and \textit{3s5z\_vs\_3s6z}. For the remaining three maps in the second row ~( \textit{1c3s8z\_vs\_1c3s9z}, \textit{5s10z}, and \textit{7s7z}), where other baselines can also find winning strategies, due to the boost learning process via episodic memory along with efficient exploration, our algorithm EMC still performs the best in the three maps, with fastest learning speed and the highest rates achieved.

\vspace{-0.1in}
\subsection{Ablation Study}
\label{ablation}
To understand the superior performance of EMC, we carry out ablation studies to test the contribution of its two main components: curiosity module and episodic memory. Following methods are included in the evaluation: (i) EMC without curiosity module~(denoted by \textit{EMC-wo-C}); (ii) EMC without episodic memory component~(denoted by \textit{EMC-wo-M}); (iii) QPLEX, which can be considered as EMC without the episodic memory component nor the curiosity module, provides a natural ablation baseline of EMC.

\begin{figure}
\vspace{-0.2in}
\centering
\subfigure[corridor]{
\begin{minipage}[t]{0.32\linewidth}
\centering
\includegraphics[width=1.8in]{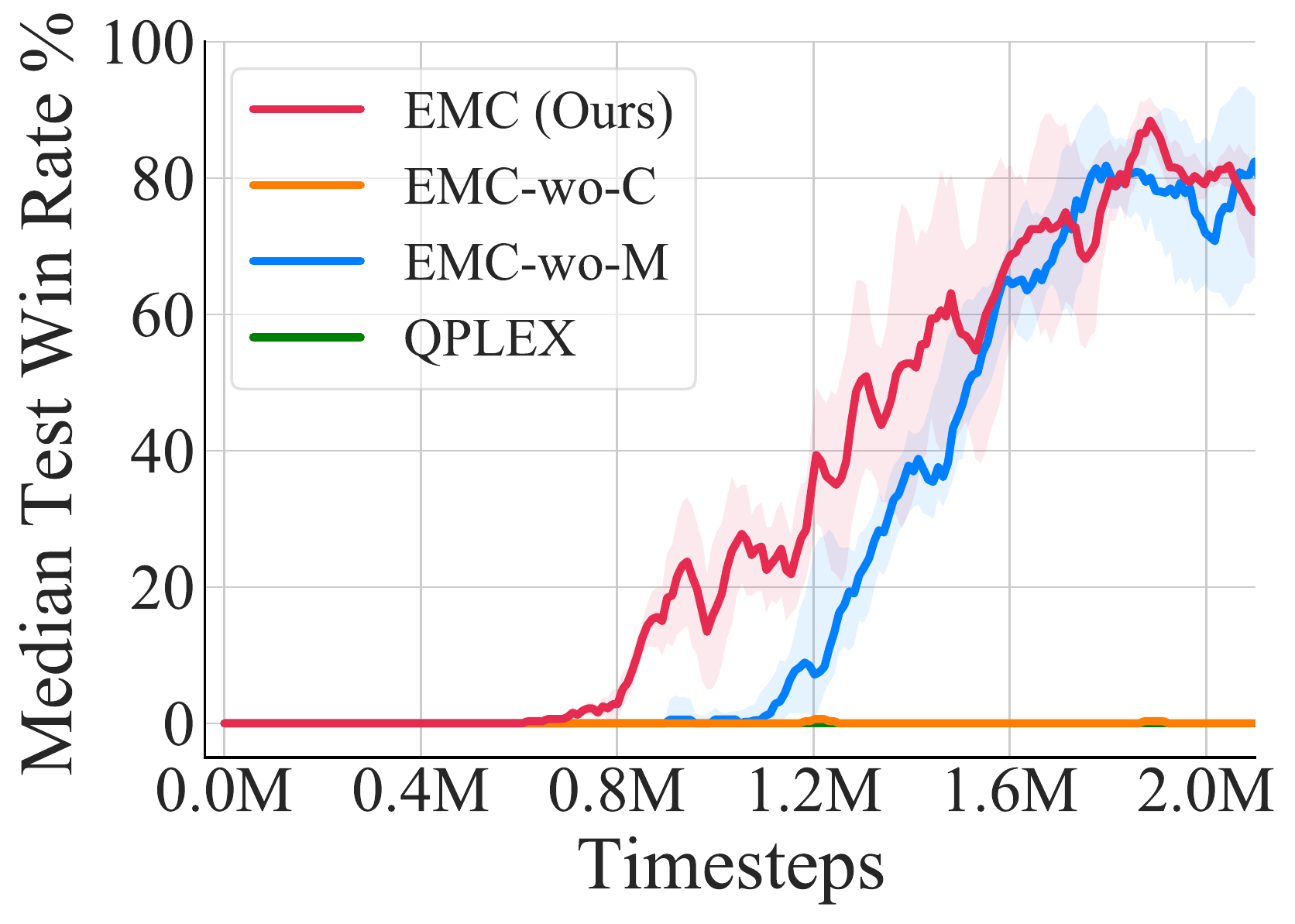}
\end{minipage}%
}%
\subfigure[2s3z]{
\begin{minipage}[t]{0.32\linewidth}
\centering
\includegraphics[width=1.8in]{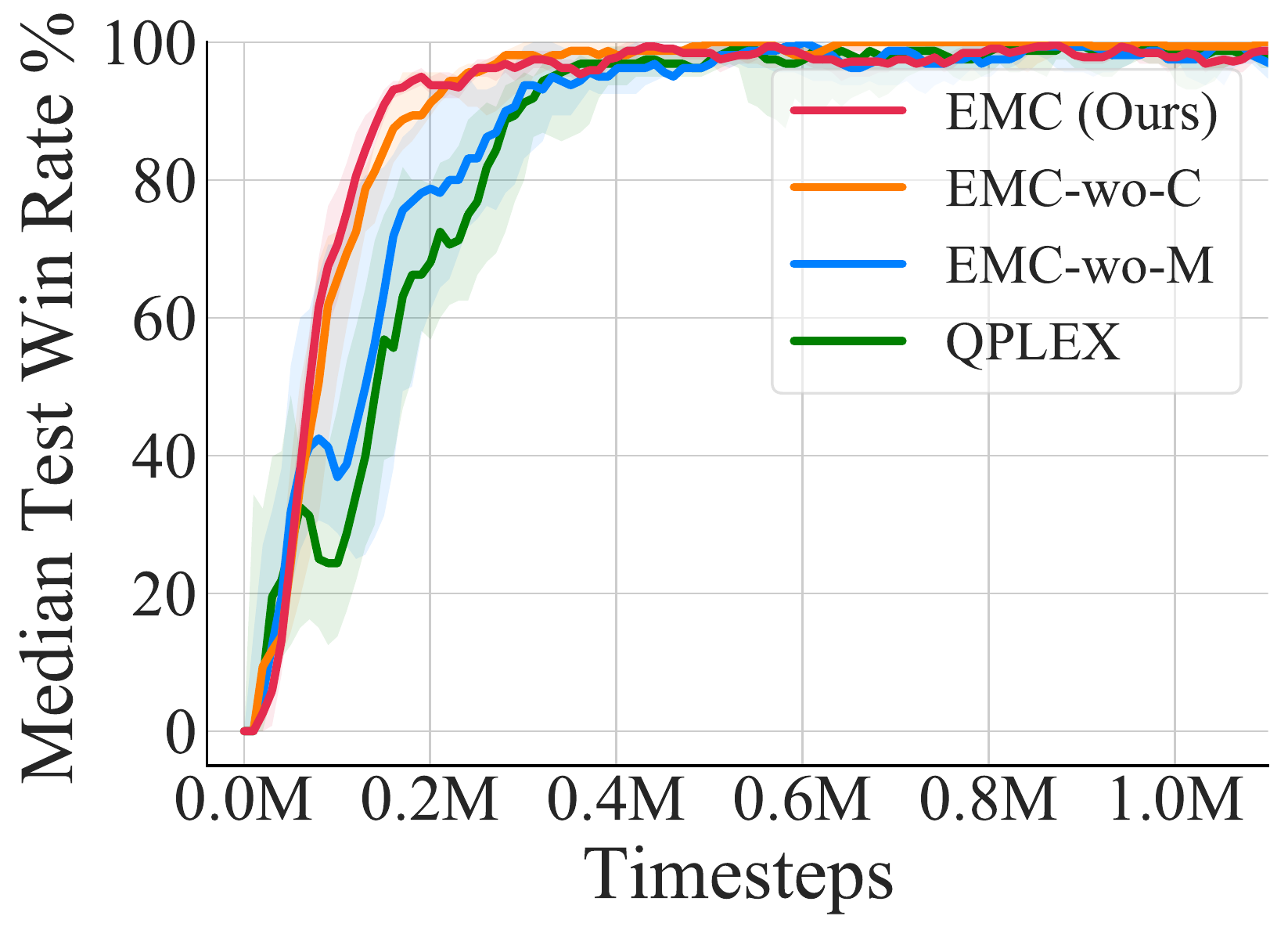}
\end{minipage}%
}%
\subfigure[3s5z]{
\begin{minipage}[t]{0.32\linewidth}
\centering
\includegraphics[width=1.8in]{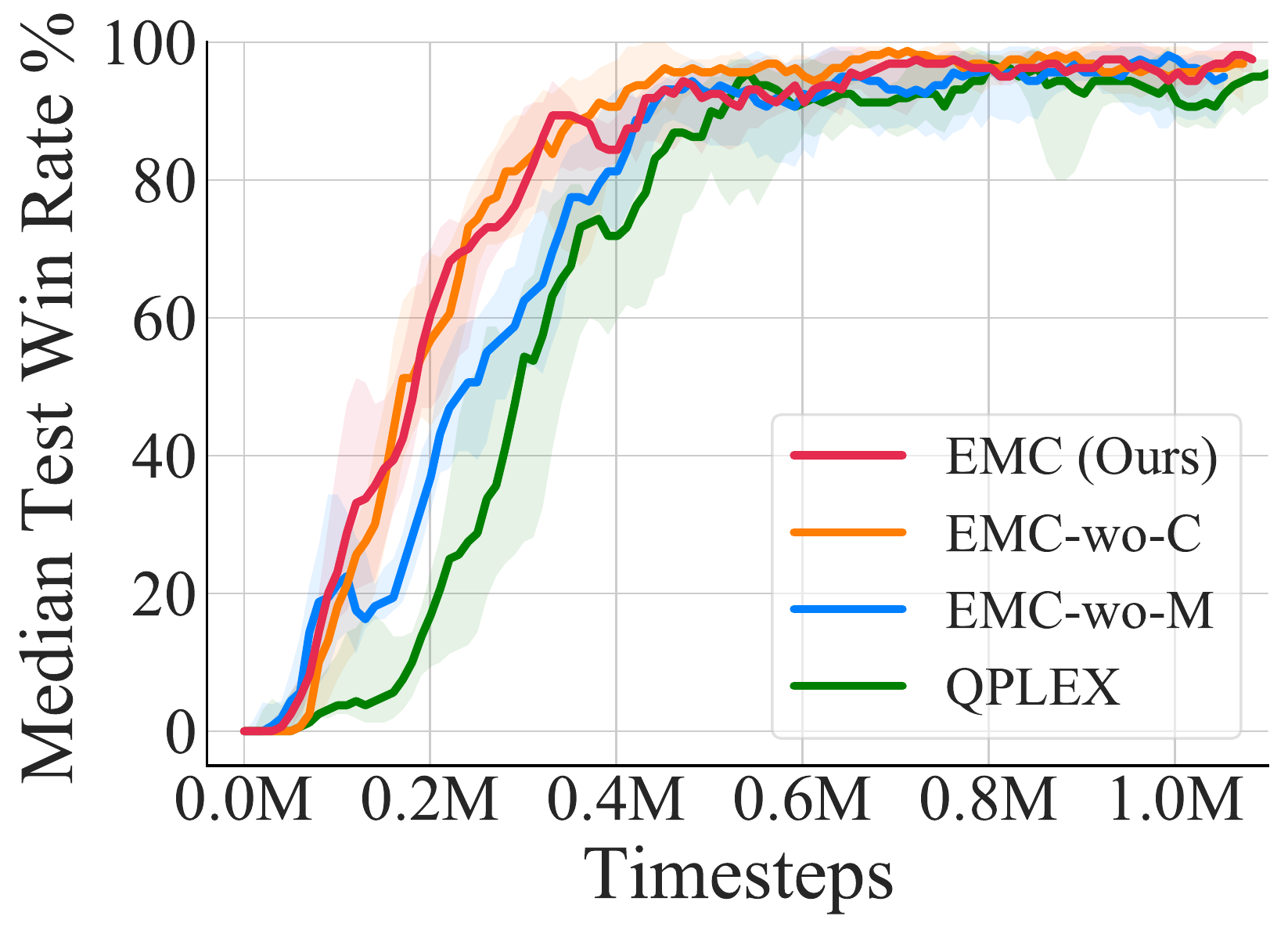}
\end{minipage}%
}%
\centering

\caption{Ablation study on the two major components.}
\label{Fig:Ablation Study}
\vspace{-0.2in}
\end{figure}

Figure~\ref{Fig:Ablation Study}(b-c) shows\ that in easy exploration maps, both EMC and EMC-wo-C achieve the state-of-the-art performance, which implies that in the easy tasks, sufficient exploration can be achieved simply by the popular $\epsilon$-greedy method. However, in super hard exploration maps~(Figure~\ref{Fig:Ablation Study} (a)), EMC-wo-C cannot solve this task but EMC has excellent performance. 
These empirical experiments show that the curiosity module plays a vital role in improving performance when sufficient and coordinated exploration is necessary. On the other hand, making the best use of good trajectories collected by exploration is also essential. As shown Figure~\ref{Fig:Ablation Study}, EMC with episodic memory enjoys better sample efficiency than EMC-wo-M in challenging~(Figure~\ref{Fig:Ablation Study}a) and easy exploration tasks~(Figure~\ref{Fig:Ablation Study}(b-c)). In general, the curiosity module and the episodic memory complement each other, and efficiently using promising exploratory experience trajectories leads to the superior performance of EMC.

Like single-agent curiosity or RND \cite{RND} exploration methods, our approach looks simple yet effective. In addition, its design choices do not look straightforward before we know how to do it right. Therefore we conduct additional ablation studies to demonstrate the effect of our elaborate formulation of curiosity bias. We introduce several baselines and compare them with EMC : (i) using the normalized TD-error of $Q_{total}$ as curiosity rewards, denoted as $\textit{EMC-TD}$; (ii) using the averaged error between the individual utilities and their targets as intrinsic rewards, denoted as $\textit{EMC-Ind}$; (iii) using the TD error of a centralized critic of the controllers which conditions on all agents' histories and actions, denoted as $\textit{EMC-Cen}$; (iv) using the averaged prediction errors of $Q_i^{ext;dec}$ which are trained in a decentralized way, denoted as $\textit{EMC-Dec}$. We aim to investigate the subtle implementation difference between EMC and EMC-TD as well as EMC-Ind, and compare the exploration efficiency of our method with the global curiosity-driven exploration method (EMC-Cen) and local curiosity-driven exploration method (EMC-Dec) empirically.
\begin{figure}
\centering
\subfigure[corridor]{
\begin{minipage}[t]{0.32\linewidth}
\centering
\includegraphics[width=1.8in]{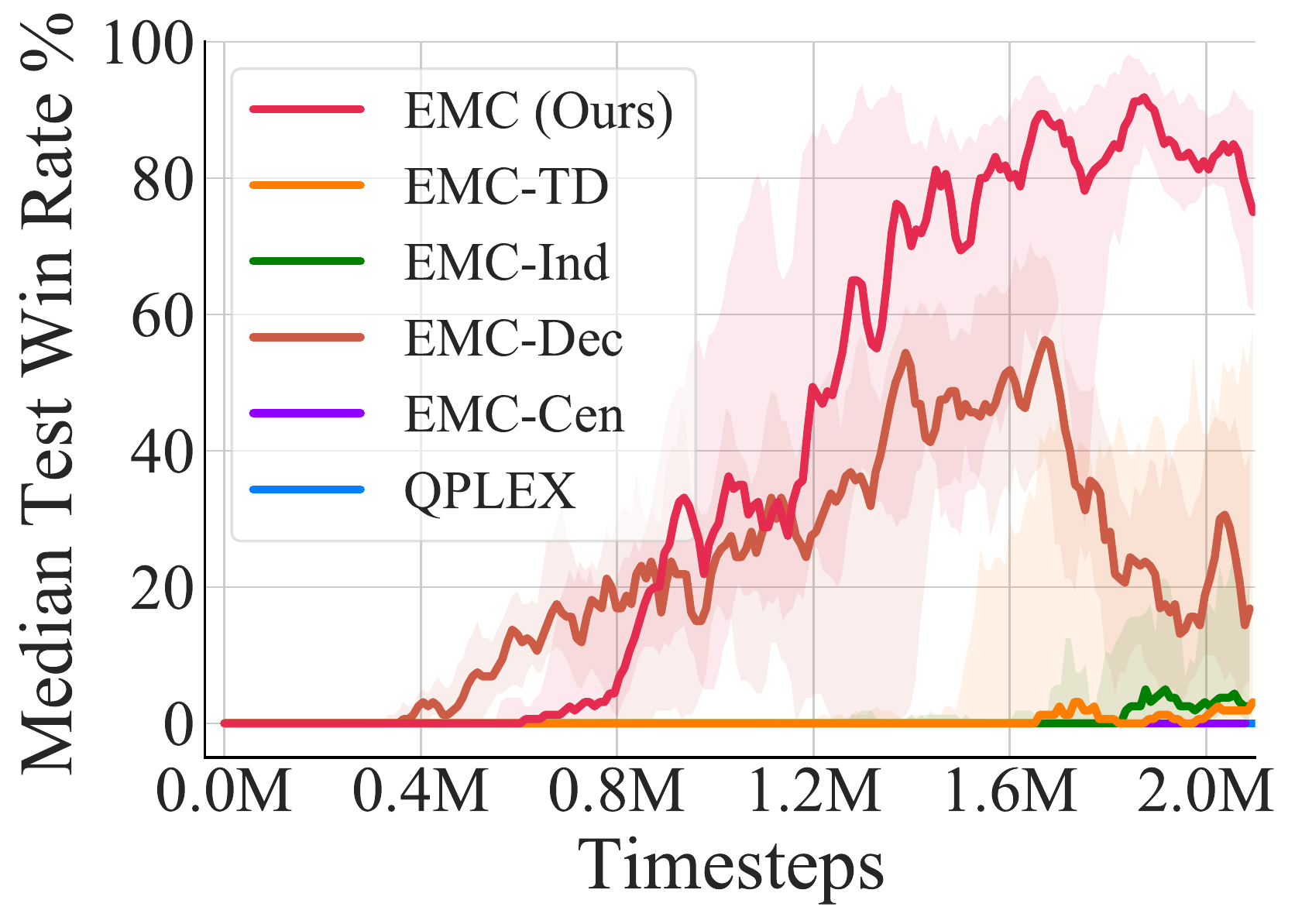}
\end{minipage}%
}%
\subfigure[3s5z\_vs\_3s6z]{
\begin{minipage}[t]{0.32\linewidth}
\centering
\includegraphics[width=1.8in]{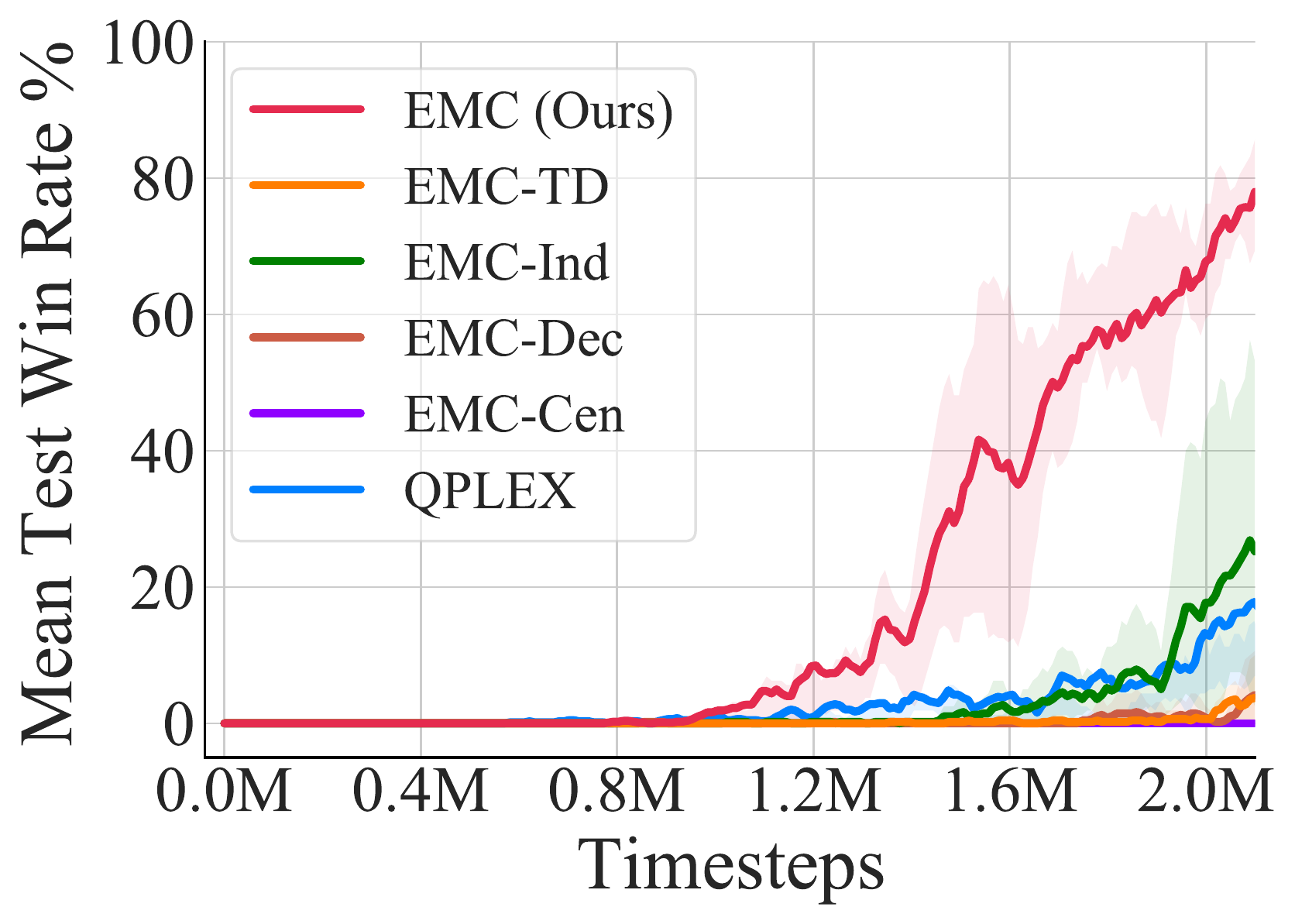}
\end{minipage}%
}%
\subfigure[6h\_vs\_8z]{
\begin{minipage}[t]{0.32\linewidth}
\centering
\includegraphics[width=1.8in]{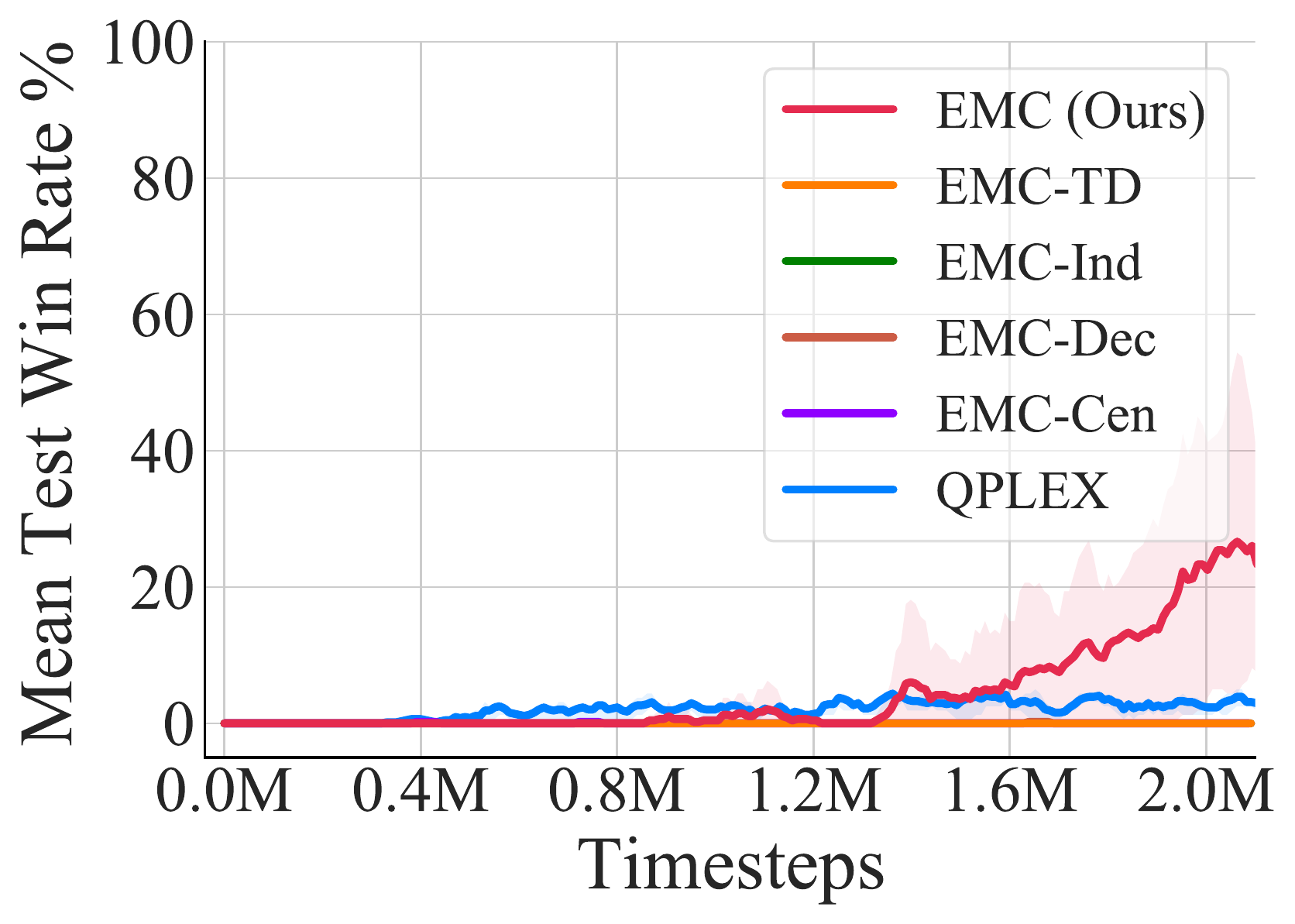}
\end{minipage}%
}%
\centering
\caption{Ablation study on design choice.}
\label{Fig:Ablation Study2}
\vspace{-0.2in}
\end{figure}

We design a variant of the toygame mentioned in section~\ref{toygamesection}, which has an additional random noisy reward region. By visualizations, we demonstrate that the agents of EMC-TD and EMC-Ind tend to get stuck in the noisy-reward region, thus resulting in sub-optimal policy, while our method show superior ability for avoiding such noise-spike problem. On the other hand, since EMC-Cen is based on global curiosity, which encourages agents to explore the whole state space without bias, it may fail in finding sparse but valuable interaction patterns in the exponentially growing space in complex tasks. When comparing with EMC and EMC-Dec, we find that the key difference is the counterfactual baseline (Eq. (\ref{eq:credit_assignment})), which can theoretically reduce the variance of EMC~\cite{DOP}. Therefore, EMC can focus more on the individual specific contribution and achieve the significant improvements.

We test these baselines in SMAC and the results are shown in Figure~\ref{Fig:Ablation Study2}, and our method significantly outperform other baselines. In general, by conducting these ablations, we demonstrate the robustness for noise spikes of our design choice ((i) and (ii)), as well as the efficiency and stability of our method compared with centralized or decentralized curiosity-driven exploration method. More detailed discussions will be deferred to Appendix E.




\vspace{-0.1in}
\section{Conclusions and Future Work}
This paper introduces EMC, a novel episodic multi-agent reinforcement learning algorithm with a curiosity-driven exploration framework that allows for efficient coordinated exploration and boosted policy training by exploiting explored informative experiences. Based on the effective exploration ability, our method shows significant outperformance over state-of-the-art MARL baselines on challenging tasks in the StarCraft II micromanagement benchmark. The limitation of our work lies in the lack of adaptive exploration methods to ensure robustness. Besides, the episodic memory may get problems in  stochastic settings. For future work, we may conduct further research in these directions.


\begin{ack}
We would like to thank the anonymous reviewers for their valuable comments and helpful suggestions.
This work is supported in part by Science and Technology Innovation 2030 – “New Generation Artificial Intelligence” Major Project (No. 2018AAA0100900), a grant from the Institute of Guo Qiang, Tsinghua University, and a grant from Turing AI Institute of Nanjing.
\end{ack}

\bibliographystyle{unsrt}
\bibliography{EMC}
\clearpage
\appendix

\section{Experiment Settings and Implementation Details}
\subsection{StarCraft II}
The benchmark we considered in our paper is the popular combat scenario of StarCraft II unit micromanagement tasks~\cite{SMAC}. In this game, the enemy units are controlled by the built-in AI, and each ally unit is controlled by the reinforcement learning agent. We use the default settings, and the results in our paper use Version SC2.4.6.2.69232. At each time-step, each agent will choose action from the discrete action space, which includes the following actions: no-op, move [direction], attack [enemy id], and stop. By taking actions, agents move and attack in continuous maps. During the game, all agents will receive a global reward equal to the total damage done to enemy units. The team will get additional bonuses of 10 by killing each enemy unit, and bonuses of 200 by winning the combat. Here we briefly introduce each map of the SMAC challenges in Table \ref{table:smac}.
\begin{table*}[ht]\small
	\centering
\begin{tabular}{ccc}
	\toprule 
	Map Name & Ally Units & Enemy Units \\ \hline
	2s3z &  2 Stalkers \& 3 Zealots &  2 Stalkers \& 3 Zealots  \\ 
	3s5z &  3 Stalkers \& 5 Zealots &  3 Stalkers \& 5 Zealots \\ 
	1c3s5z & 1 Colossus, 3 Stalkers \& 5 Zealots &  1 Colossus, 3 Stalkers \& 5 Zealots \\ \hline
	5m\_vs\_6m & 5 Marines & 6 Marines \\
	10m\_vs\_11m & 10 Marines & 11 Marines \\
	27m\_vs\_30m & 27 Marines & 30 Marines \\
	3s5z\_vs\_3s6z & 3 Stalkers \& 5 Zealots &  3 Stalkers \& 6 Zealots \\
	MMM2 & 1 Medivac, 2 Marauders \& 7 Marines & 1 Medivac, 2 Marauders \& 8 Marines \\ \hline
	2s\_vs\_1sc & 2 Stalkers & 1 Spine Crawler \\
	3s\_vs\_5z & 3 Stalkers & 5 Zealots \\
	6h\_vs\_8z & 6 Hydralisks & 8 Zealots \\
	bane\_vs\_bane & 20 Zerglings \& 4 Banelings & 20 Zerglings \& 4 Banelings \\
	2c\_vs\_64zg & 2 Colossi & 64 Zerglings \\ 
	corridor & 6 Zealots & 24 Zerglings \\ \hline
	5s10z & 5 Stalkers \& 10 Zealots &  5 Stalkers \& 10 Zealots \\ 
	7sz & 7 Stalkers \& 7 Zealots &  7 Stalkers \& 7 Zealots \\ 
	1c3s8z\_vs\_1c3s9z & 1 Colossus, 3 Stalkers \& 8 Zealots &  1 Colossus, 3 Stalkers \& 9 Zealots \\ \toprule
\end{tabular}
	\caption{SMAC challenges.}\label{table:smac}
\end{table*}
\subsection{Didactic Examples}
Figure \ref{Fig:gridworld} shows the referred didactic example in section 5.1, which is a $11 \times 12$ grid world game. The blue agent and red agent can choose one of the five actions: \textit{[up, down, left, right, stay]} at each time step. The two agents is isolated by the wall, and they cannot be observed by the other one until they get into the $5\times 6$ light shaded area. They will receive a global positive reward of 10 if and only if they arrive at the dark shaded grid at the same time. If only one arrives, the incoordination will be punished by a negative reward of $-p$.

\subsection{Implementation Details}\label{sec:implmentation}
We adopt the PyMARL \citep{SMAC} implementation of state-of-the-art baselines: RODE \cite{RODE}, QPLEX \cite{qplex}, MAVEN ~\cite{MAVEN}, Qtran~ \cite{QTRAN},VDN~\cite{VDN},QMIX~\cite{QMIX} and Weighted-QMIX \cite{WQMIX}. The hyper-parameters of these algorithms are the same as that in SMAC \cite{SMAC} and referred in their source codes. Our method is also based on QPLEX, and the hyper-parameters are the same referred in its source codes. While the special hyper-parameters are illustrated in Table \ref{table:configuration_EMC} and other common hyper-parameters are adopted by the default implementation of PyMARL \cite{SMAC}. 

We conduct experiments on an NVIDIA Tesla P100 GPU, and each task in SMAC needs to train about 20 hours to 30 hours, depending on the number of agents and episode length limit of each map.  We evaluate 32 episodes with decentralized greedy action selection without $\epsilon-greedy$ strategy every 10k timesteps for each algorithm. The test win rate shows the percentage of episodes in which agents defeat all enemy units within the time limit. Besides, since the intrinsic rewards need to vanish as the policy converges, we use a decaying weighting term to scale the intrinsic rewards: $\tilde{r}_t^{int}=\eta_t*r_t^{int},\eta_{t+200k}=0.9*\eta_{t}$.
In the three super hard maps: corridor, 3s5z\_vs\_3s6z, 6h\_vs\_8z, we set $\eta=0.05$ while set $\eta=0.0001$ in other maps.

\begin{table}
	\centering
\begin{tabular}{ccc}
	\toprule 
	EMC's architecuture configurations & Value \\ \hline
	soft update weight & 0.05\\
	weighting term $\lambda$ of episodic loss &0.1 or 0.01 \\
	episodic memory capacity &1M \\
	episodic latent dim & 4 \\
	\toprule
\end{tabular}
	\caption{The hyper-parameters of EMC’s architecture.}\label{table:configuration_EMC}
\end{table}

 \section{Experiments on StarCraftII}

\begin{figure*}[ht]
\centering
\begin{minipage}[t]{\linewidth}
\centering
\includegraphics[width=0.95\textwidth]{figure/legend.pdf}
\end{minipage}%

\subfigure[2s\_vs\_1sc]{
\begin{minipage}[t]{0.32\linewidth}
\centering
\includegraphics[width=1.8in]{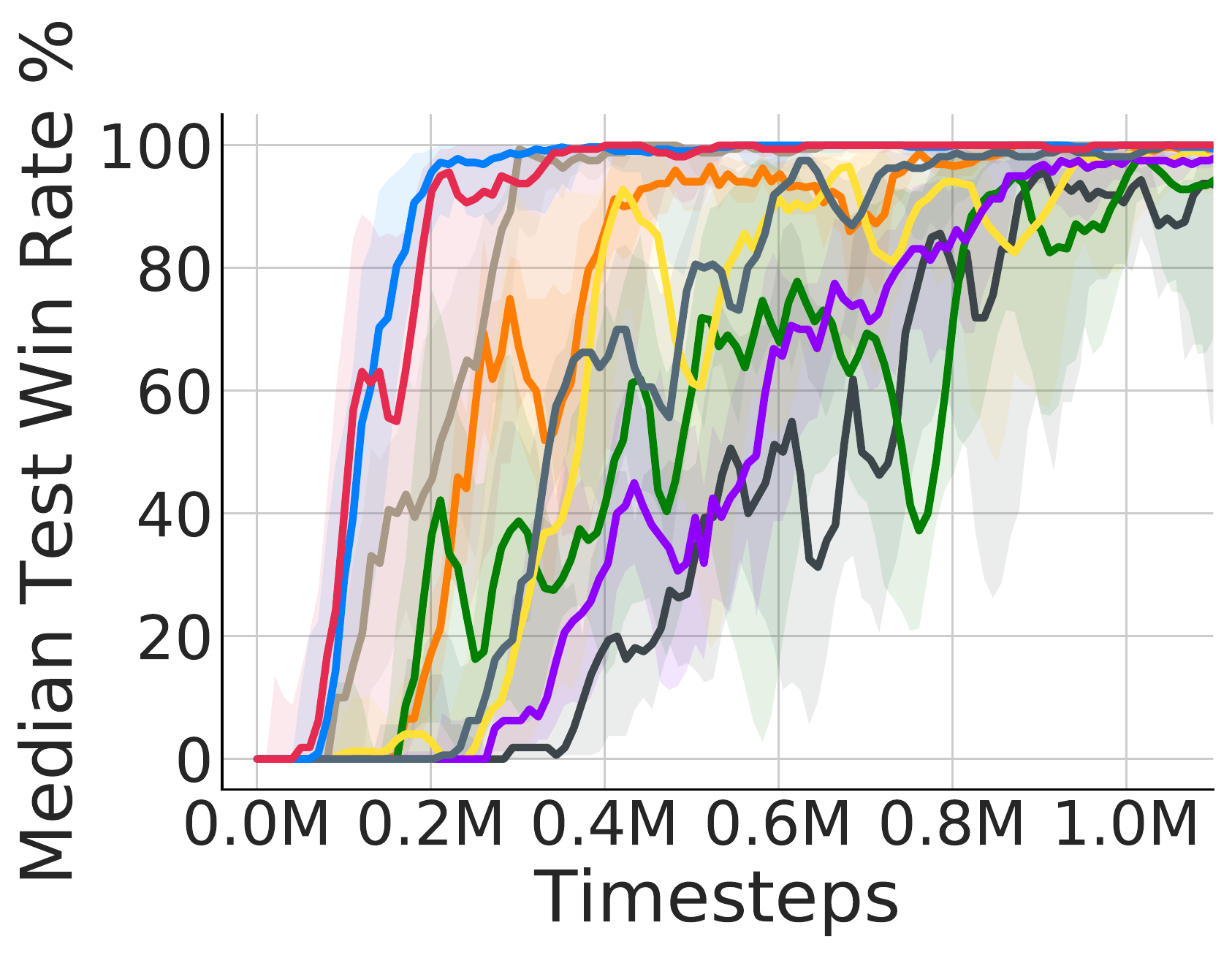}
\end{minipage}%
}%
\subfigure[2s3z]{
\begin{minipage}[t]{0.32\linewidth}
\centering
\includegraphics[width=1.8in]{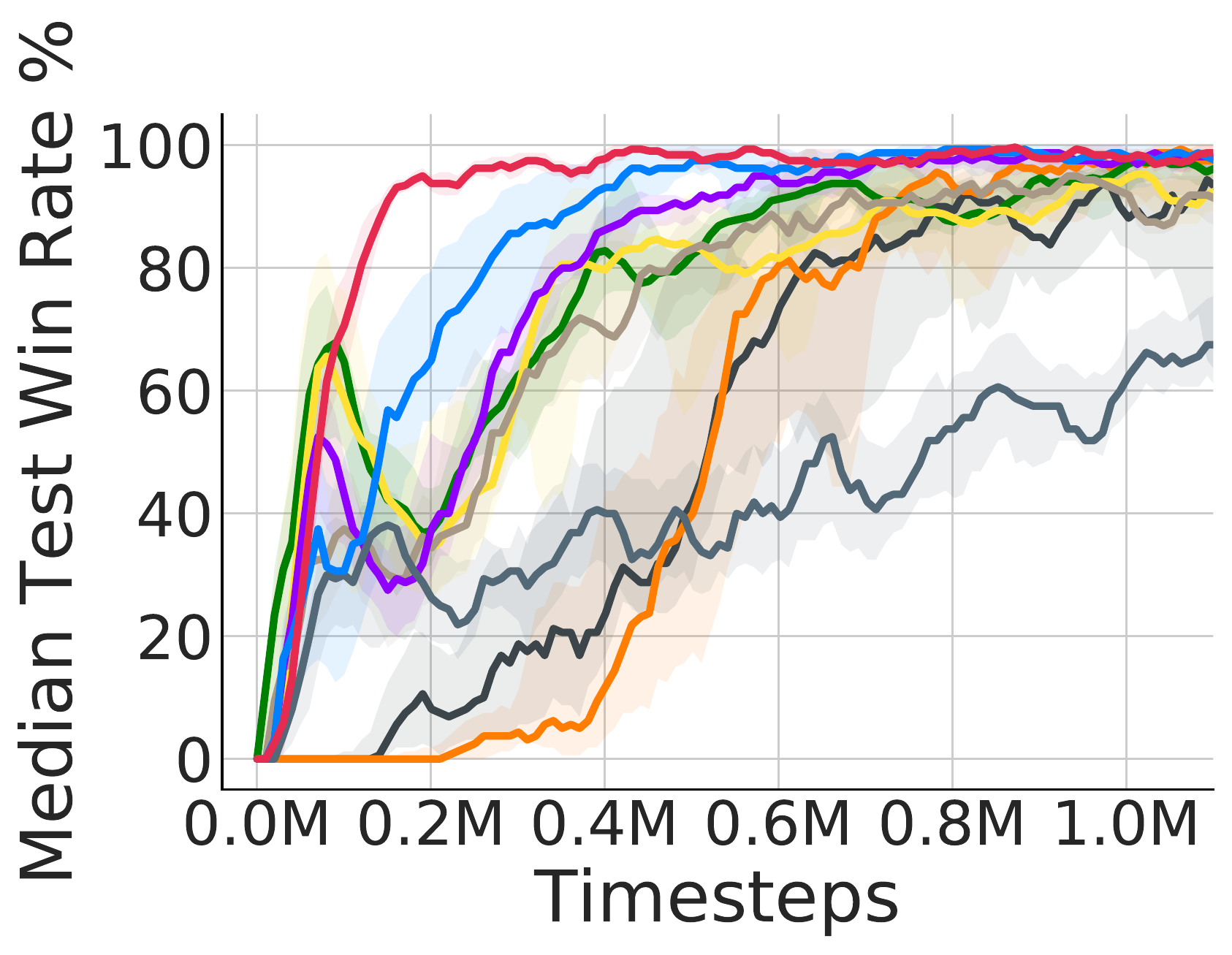}
\end{minipage}%
}%
\subfigure[3s5z]{
\begin{minipage}[t]{0.32\linewidth}
\centering
\includegraphics[width=1.8in]{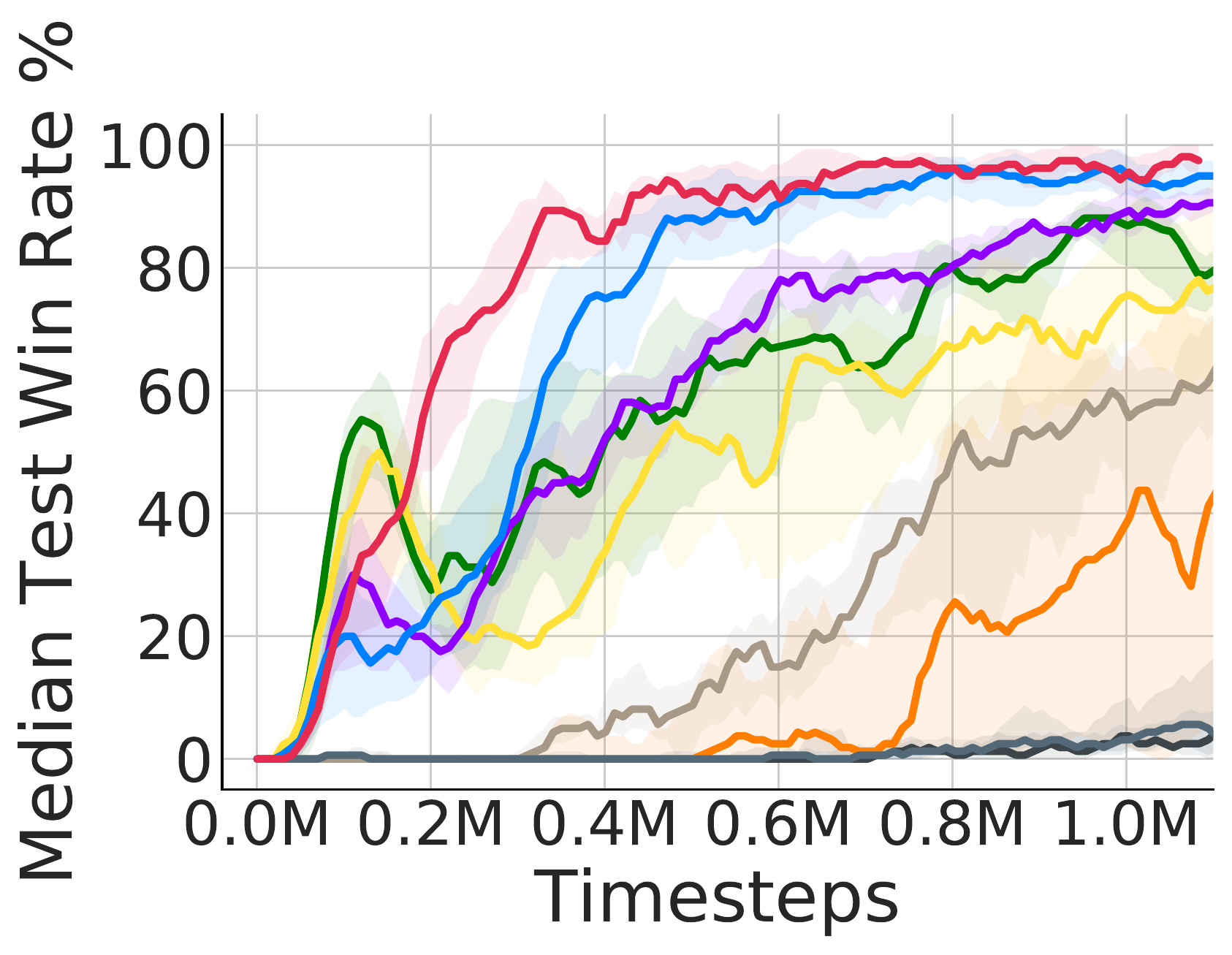}

\end{minipage}
}%

\subfigure[1c3s5z]{
\begin{minipage}[t]{0.32\linewidth}
\centering
\includegraphics[width=1.8in]{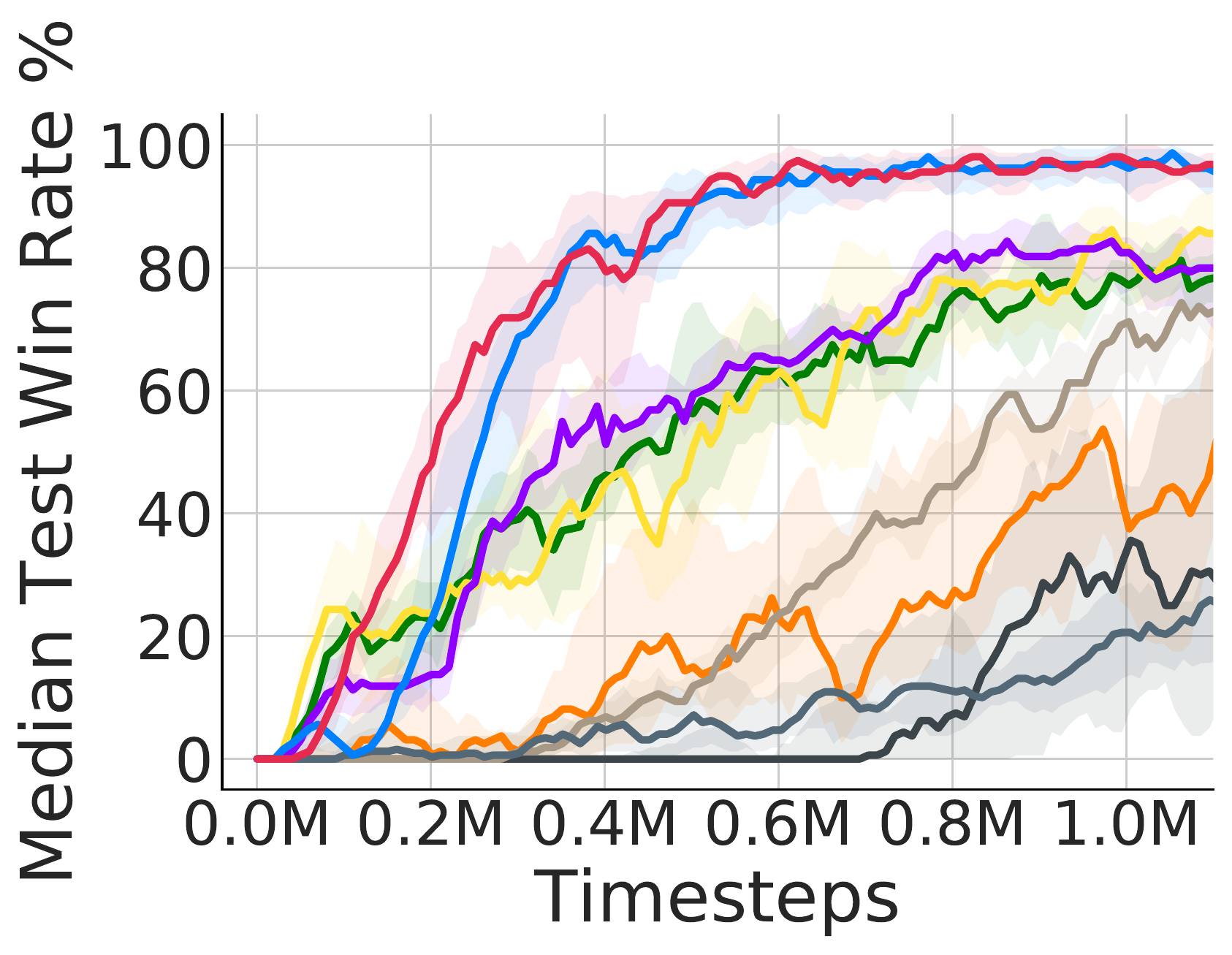}
\end{minipage}%
}%
\subfigure[3s\_vs\_5z]{
\begin{minipage}[t]{0.32\linewidth}
\centering
\includegraphics[width=1.8in]{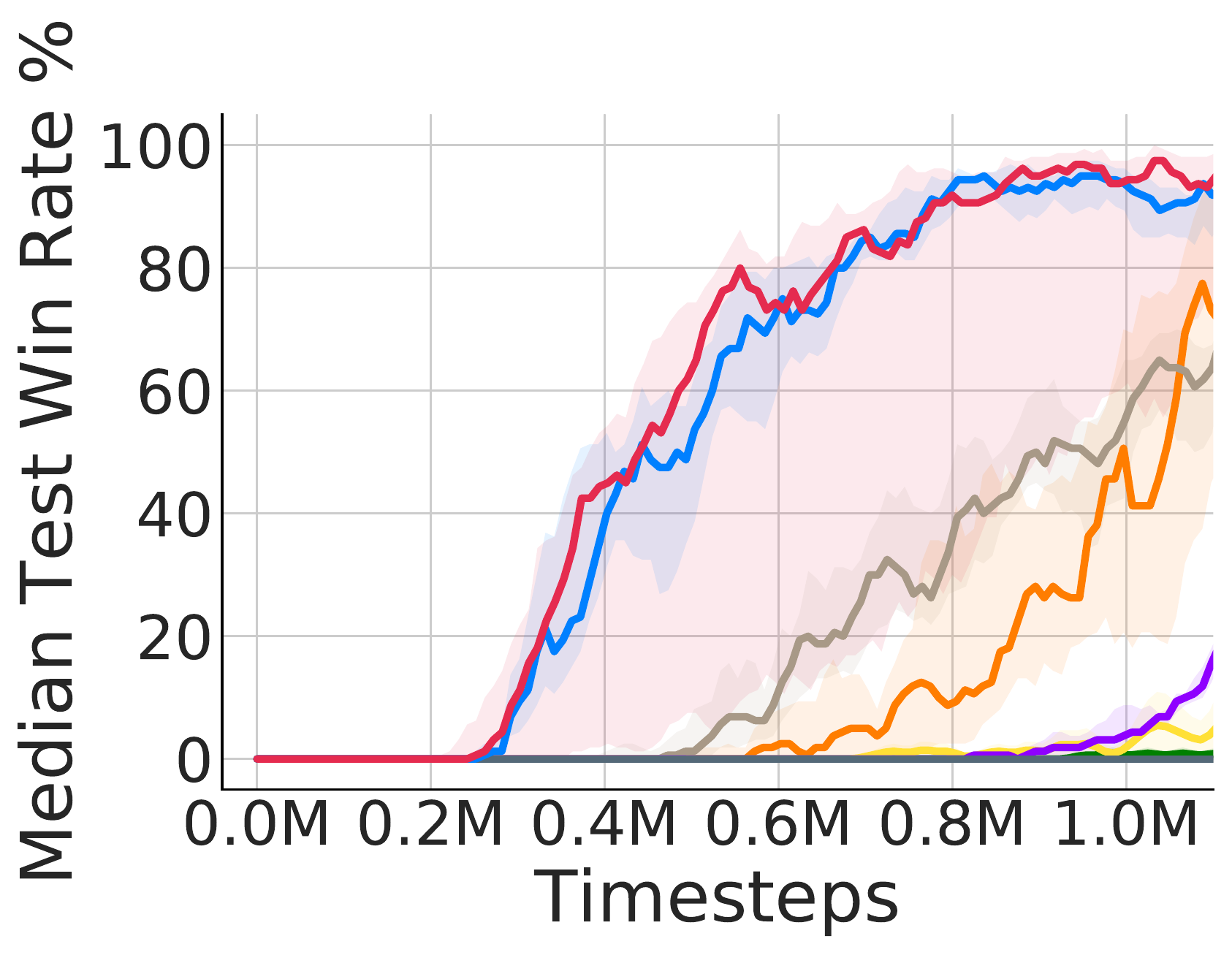}
\end{minipage}%
}%
\subfigure[bane\_vs\_bane]{
\begin{minipage}[t]{0.32\linewidth}
\centering
\includegraphics[width=1.8in]{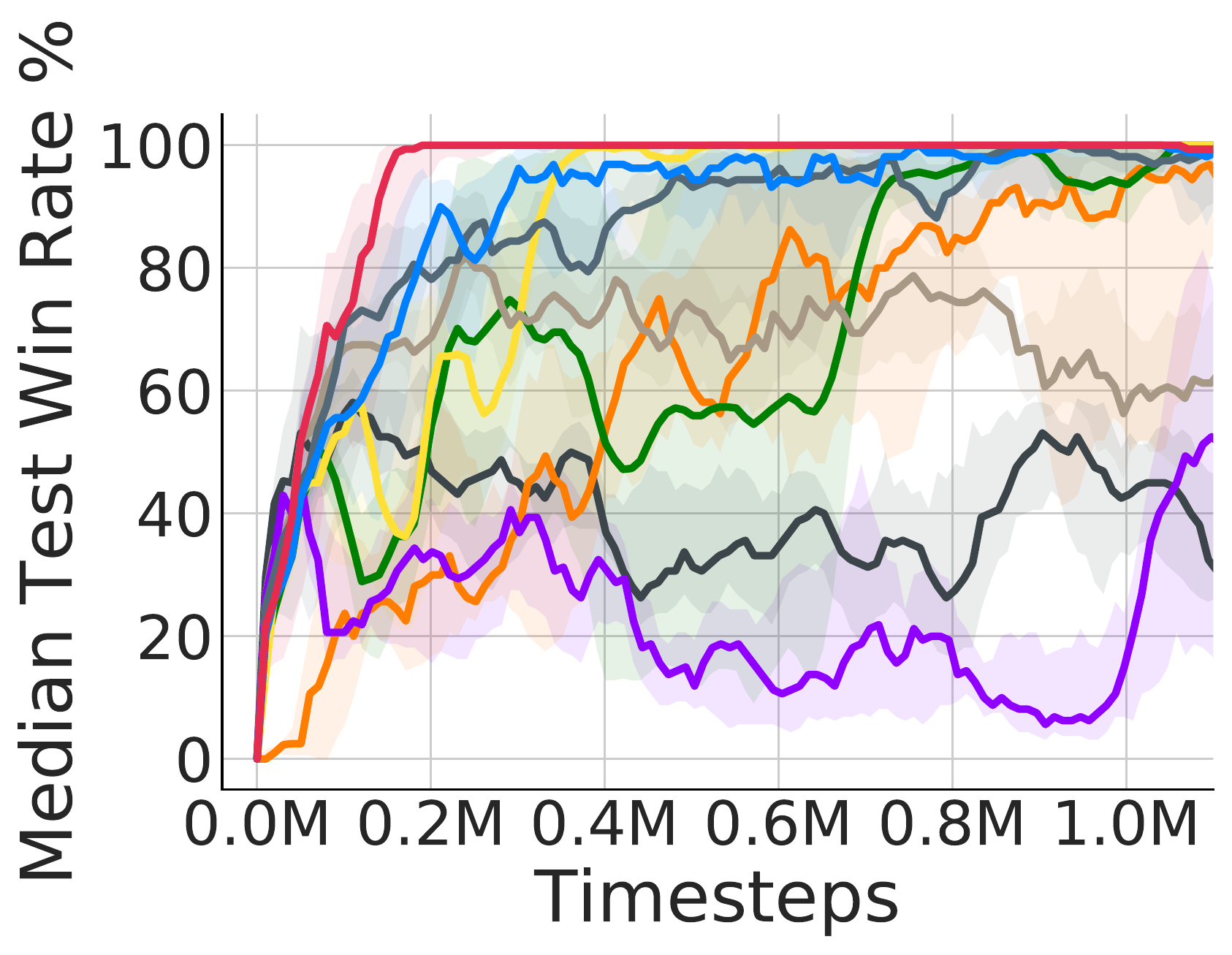}
\end{minipage}
}%

\centering

\caption{Results of the $6$ easy maps in the SMAC experiments. \label{Fig:SmacExpEasy}}

\end{figure*}

Figure \ref{Fig:SmacExpEasy} shows the performance of  $6$ easy maps in SMAC. It can be found that our algorithm perform the best in $5$ of the $6$ easy maps. In the map \textit{2s\_vs\_1sc}, although EMC is the second best, the performance gap between EMC and the QPLEX algorithm is very subtle. The advantage of EMC over the other algorithms can be found in Figures \ref{Fig:SmacExpEasy}(b), \ref{Fig:SmacExpEasy}(c), and \ref{Fig:SmacExpEasy}(f), where it converges much faster than the second best algorithm QPLEX. For example, in the \textit{bane\_vs\_bane} tasks, the EMC algorithm reaches a $100\%$ win rate in fewer than $0.2$M steps, while the QPLEX algorithm converges at the time step of $0.3$M. Futhermore, the win rate of QPLEX does not reach $100\%$ in this map and its learning process is not as stable as that of EMC. Therefore, although the state-of-the-art algorithms such as QPLEX performs sufficiently well in these easy maps, the coordinated exploration mechanism and episodic-memory control equipped by EMC can further enhance the performance a learning algorithm.

\begin{figure*}[ht]
\centering
\begin{minipage}[t]{\linewidth}
\centering
\includegraphics[width=0.95\textwidth]{figure/legend.pdf}
\end{minipage}%

\subfigure[2c\_vs\_64zg]{
 \begin{minipage}[t]{0.32\linewidth}
 \centering
\includegraphics[width=1.8in]{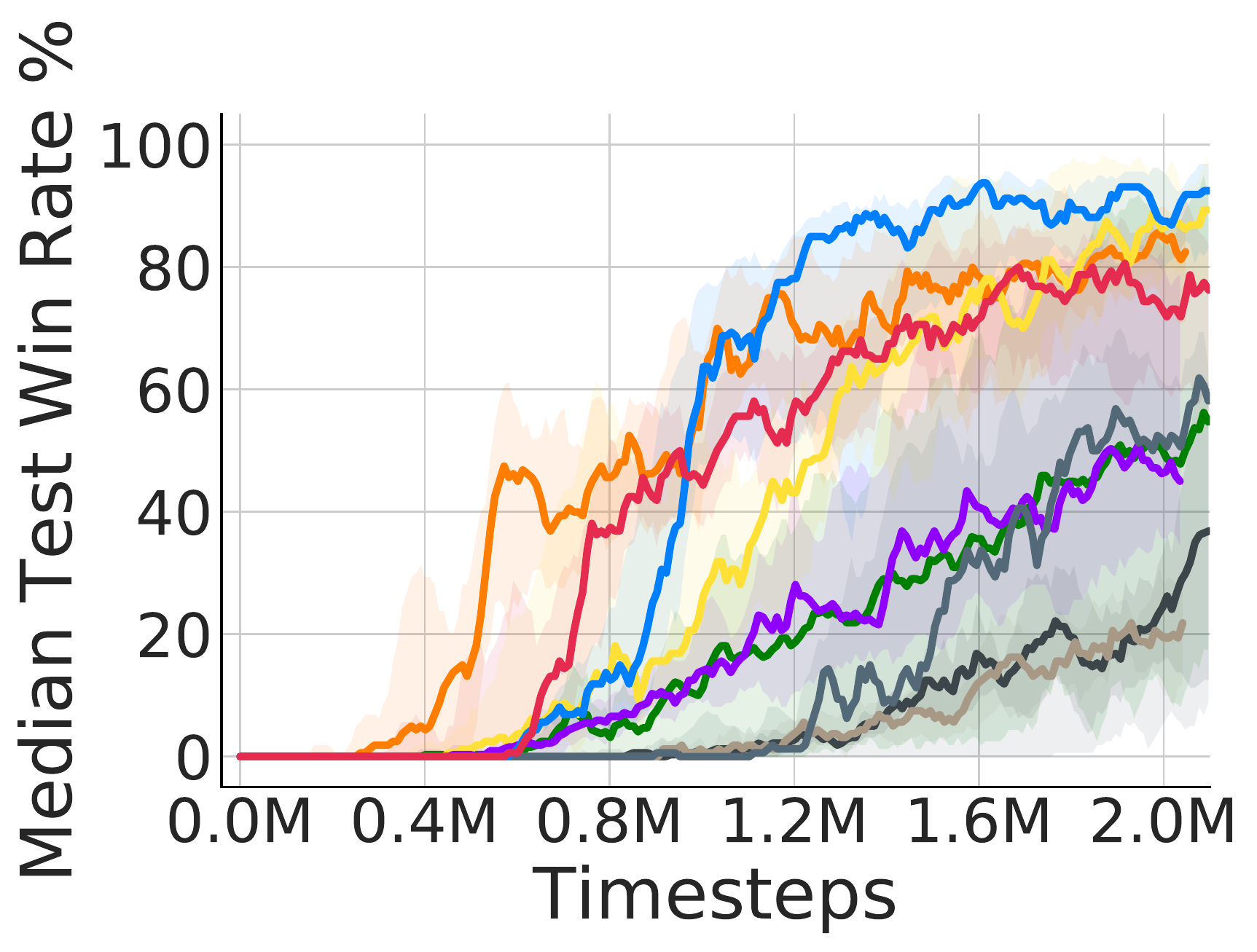}
 \end{minipage}%
}%
 \subfigure[27m\_vs\_30m]{
 \begin{minipage}[t]{0.32\linewidth}
 \centering
 \includegraphics[width=1.8in]{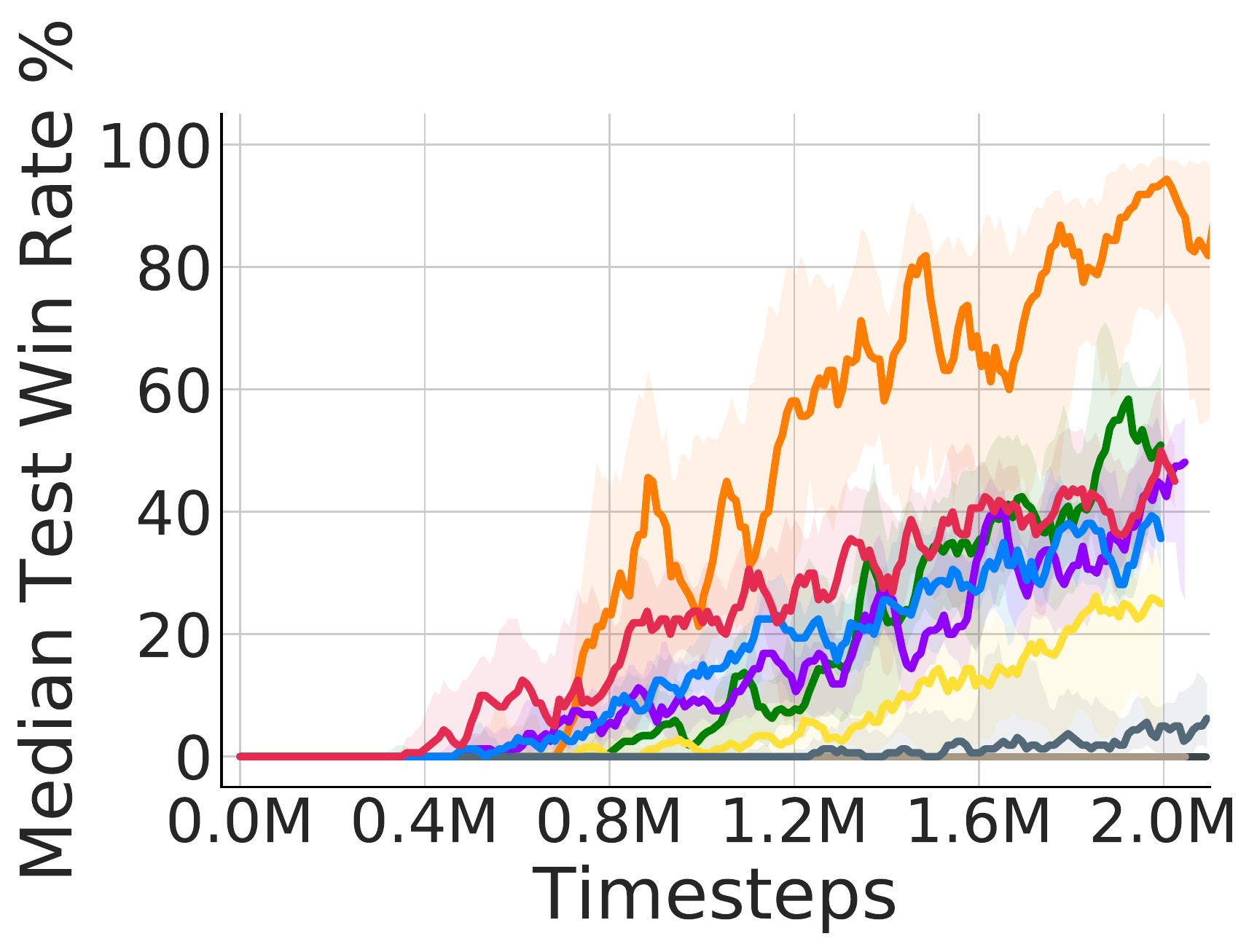}
\end{minipage}
 }%
 \subfigure[MMM2]{
 \begin{minipage}[t]{0.32\linewidth}
 \centering
 \includegraphics[width=1.8in]{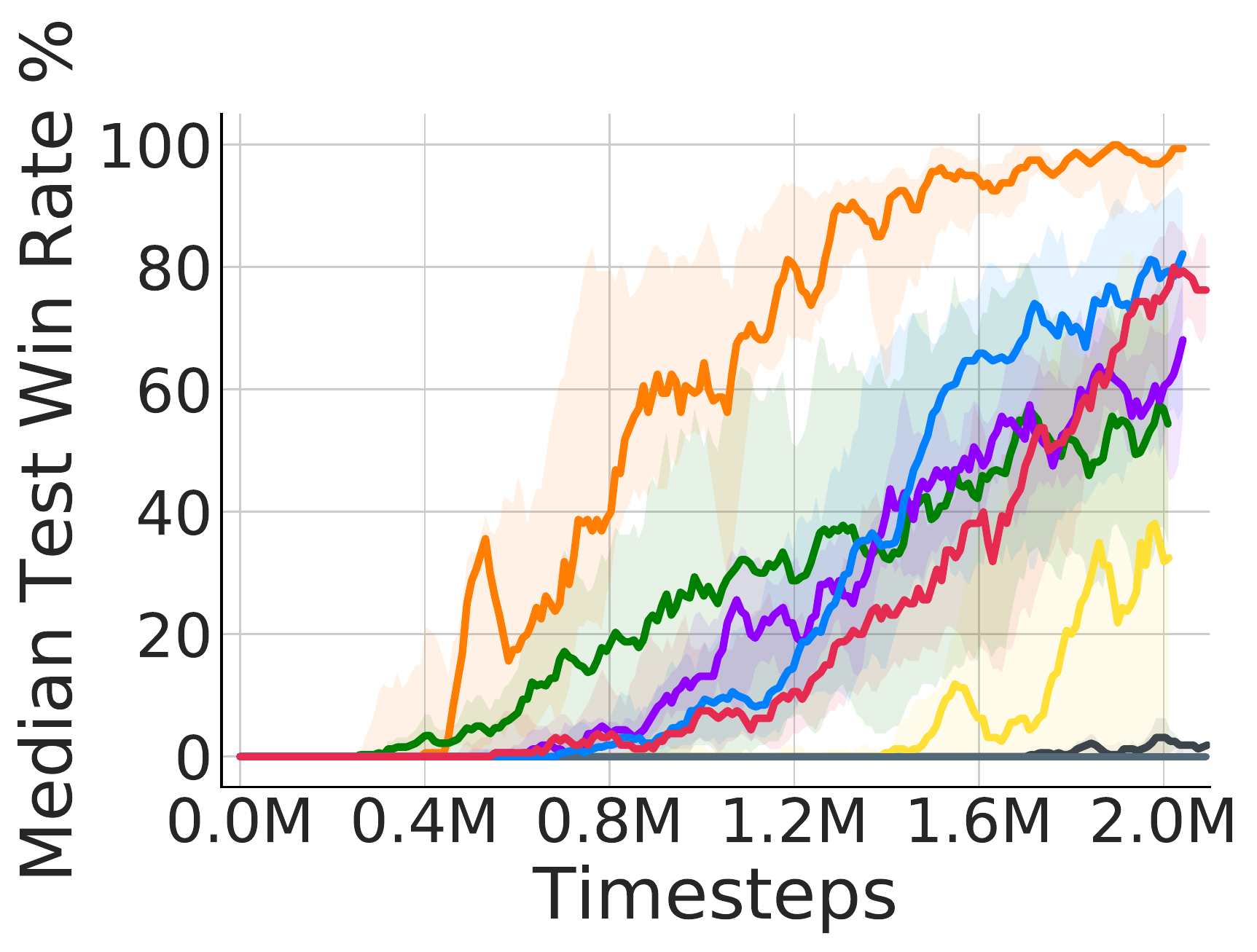}
 \end{minipage}%
 }%

 \subfigure[5m\_vs\_6m]{
 \begin{minipage}[t]{0.32\linewidth}
 \centering
 \includegraphics[width=1.8in]{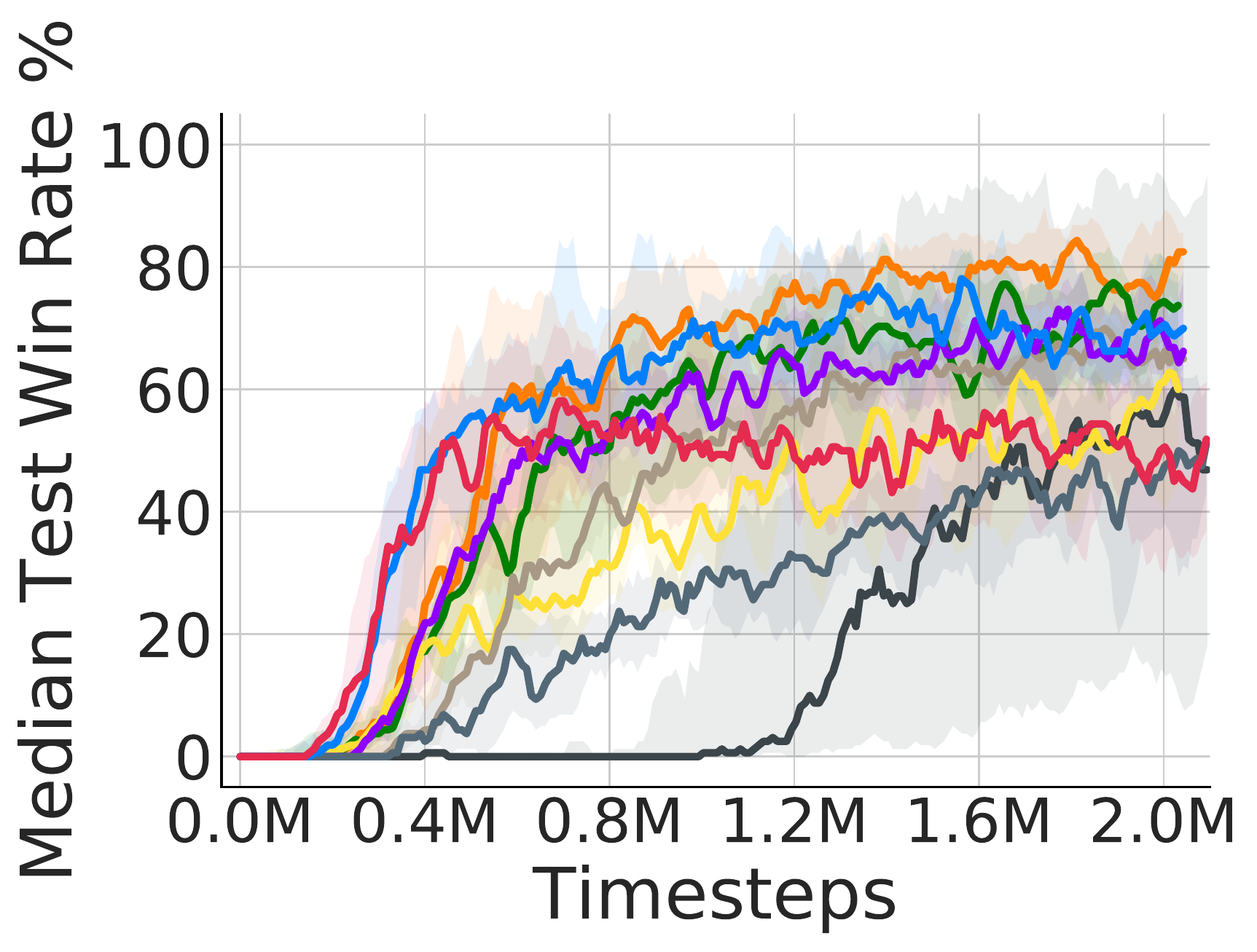}
 \end{minipage}%
 }%
 \subfigure[10m\_vs\_11m]{
 \begin{minipage}[t]{0.32\linewidth}
 \centering
 \includegraphics[width=1.8in]{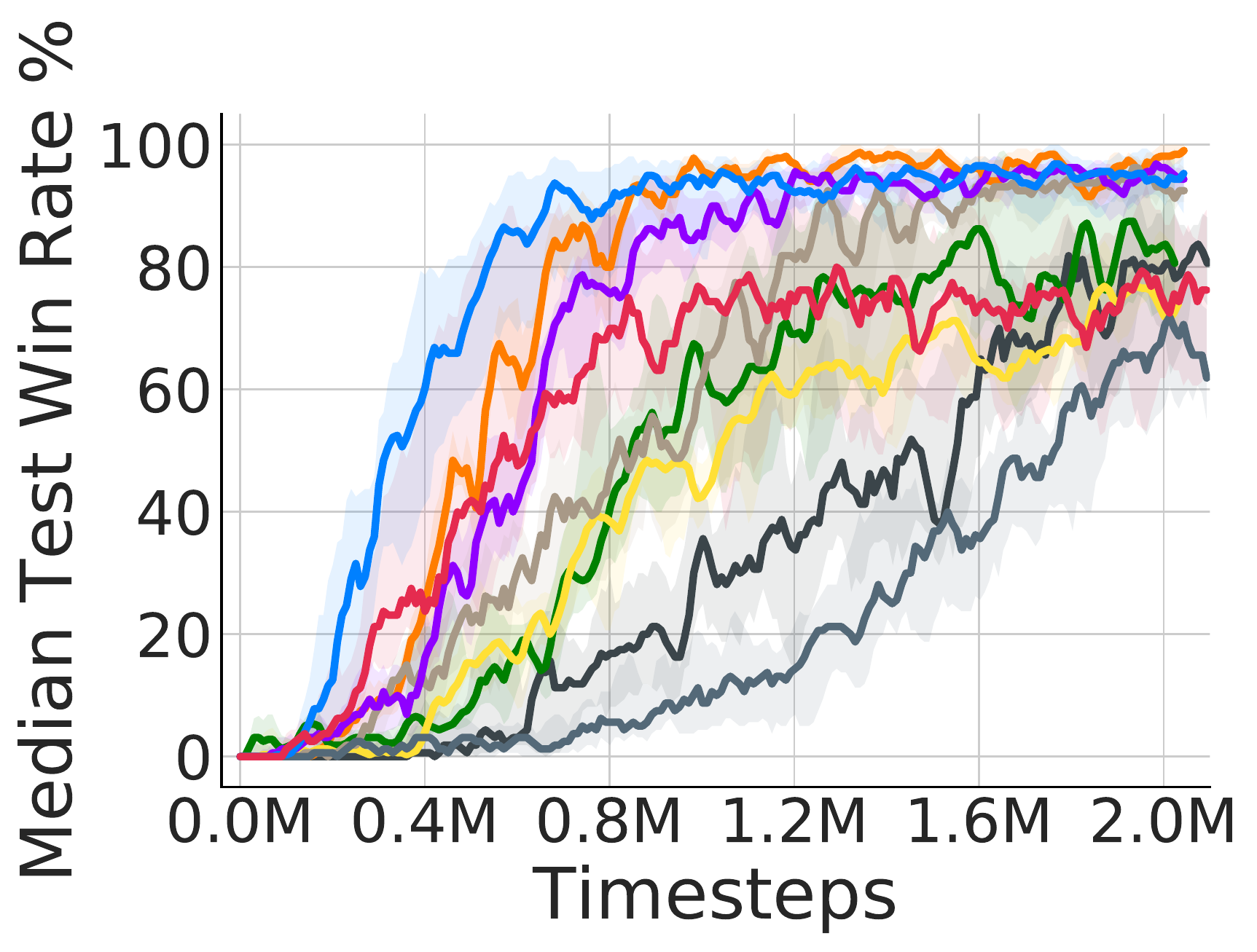}
 \end{minipage}%
  }%

 \centering
 \caption{Results of the remaining $5$ maps in the SMAC experiments  \label{Fig:SmacExpComparable}.}
 \end{figure*}

  The results in Figure \ref{Fig:SmacExpComparable} shows that the EMC algorithm achieves comparble performance with the state-of-the-art baseline algorithms in the $5$ maps \textit{2c\_vs\_64zg}, \textit{27m\_vs\_30m}, \textit{MMM2}, \textit{5m\_vs\_6m}, and \textit{10m\_vs\_11m}. Moreover, our method may not perform pretty well in few maps, and we hypothesis that the training process of EMC may get stuck in local optimal due to the episodic control mechanism. 
 It also should be noted that the MAVEN algorithm, which is specially designed for the multi-agent exploration problem, performs the worst in the these maps. And our algorithm, which is also equipped with an exploration mechanism, outperforms MAVEN drastically.

 \section{Experiments on the Coordinated Toygame}

  \begin{figure*}[ht]
\centering
\begin{minipage}[t]{\linewidth}
\centering
\includegraphics[width=0.99\textwidth]{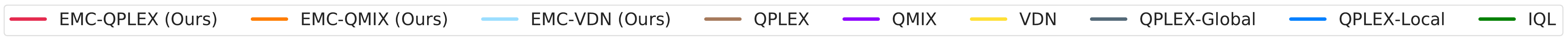}
\end{minipage}%

\subfigure[\(p=0\)]{
 \begin{minipage}[t]{0.32\linewidth}
 \centering
\includegraphics[width=1.8in]{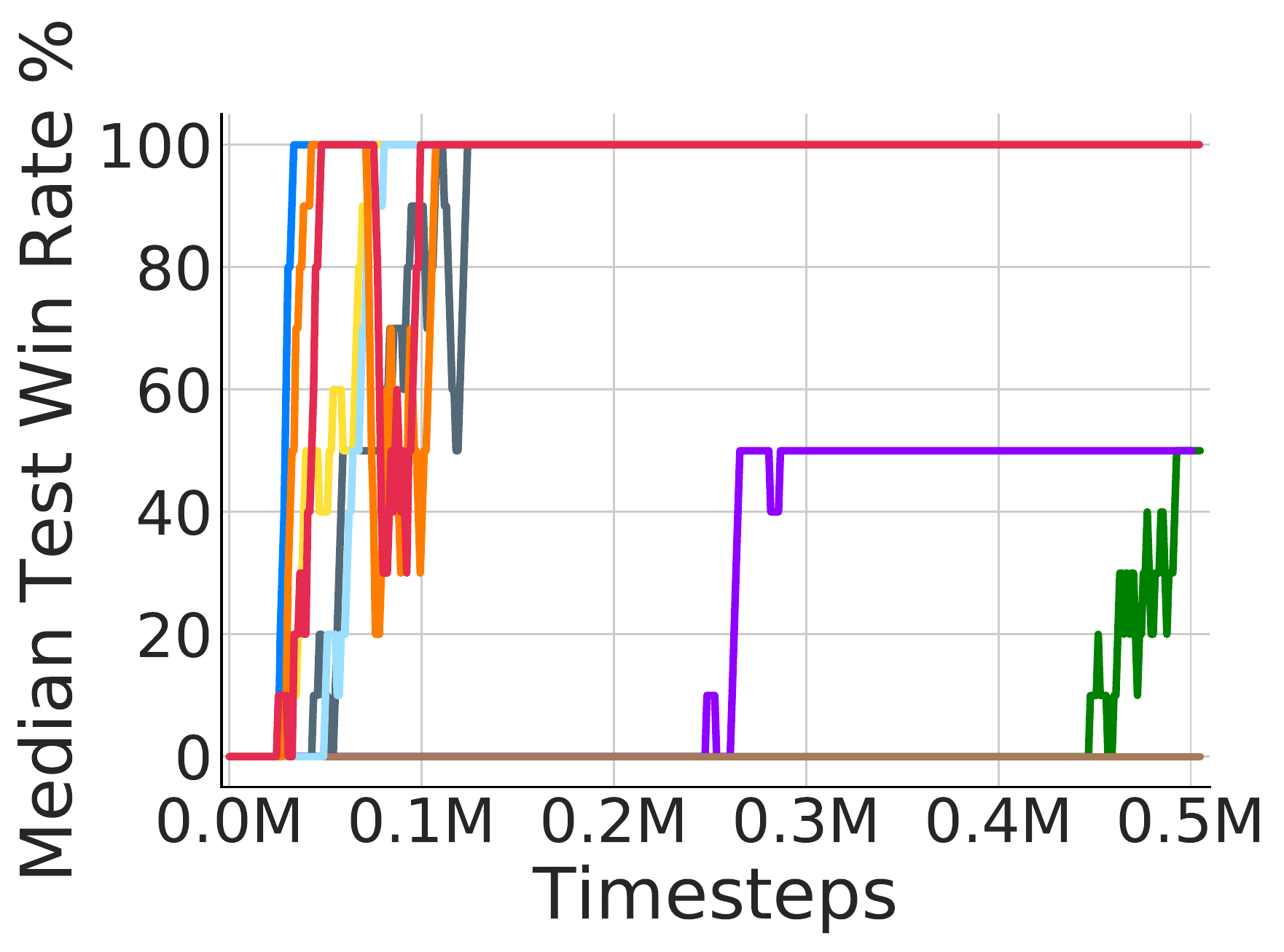}
 \end{minipage}%
}%
 \subfigure[\(p=0.5\)]{
 \begin{minipage}[t]{0.32\linewidth}
 \centering
 \includegraphics[width=1.8in]{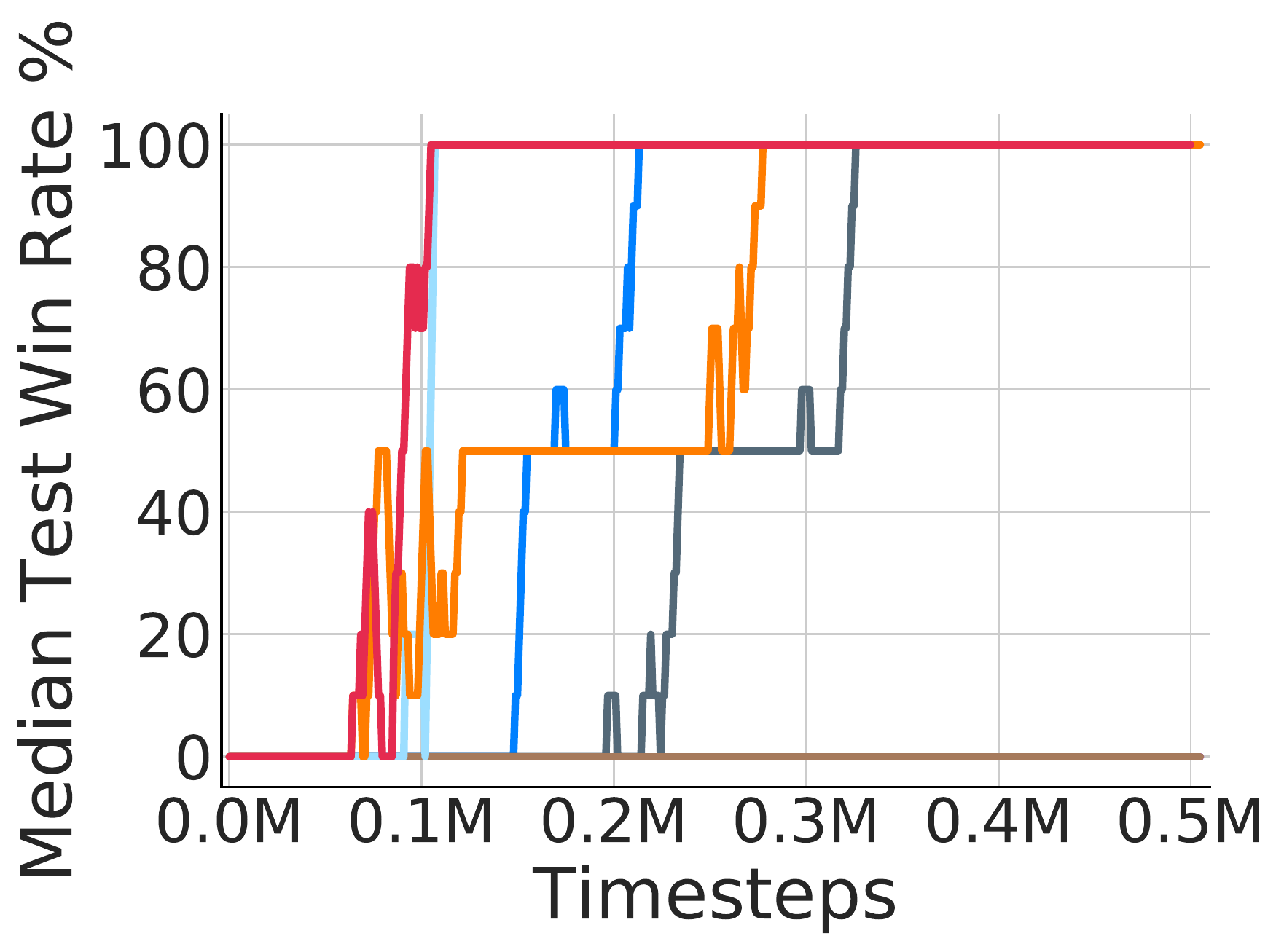}
\end{minipage}
 }%
 \subfigure[\(p=1\)]{
 \begin{minipage}[t]{0.32\linewidth}
 \centering
 \includegraphics[width=1.8in]{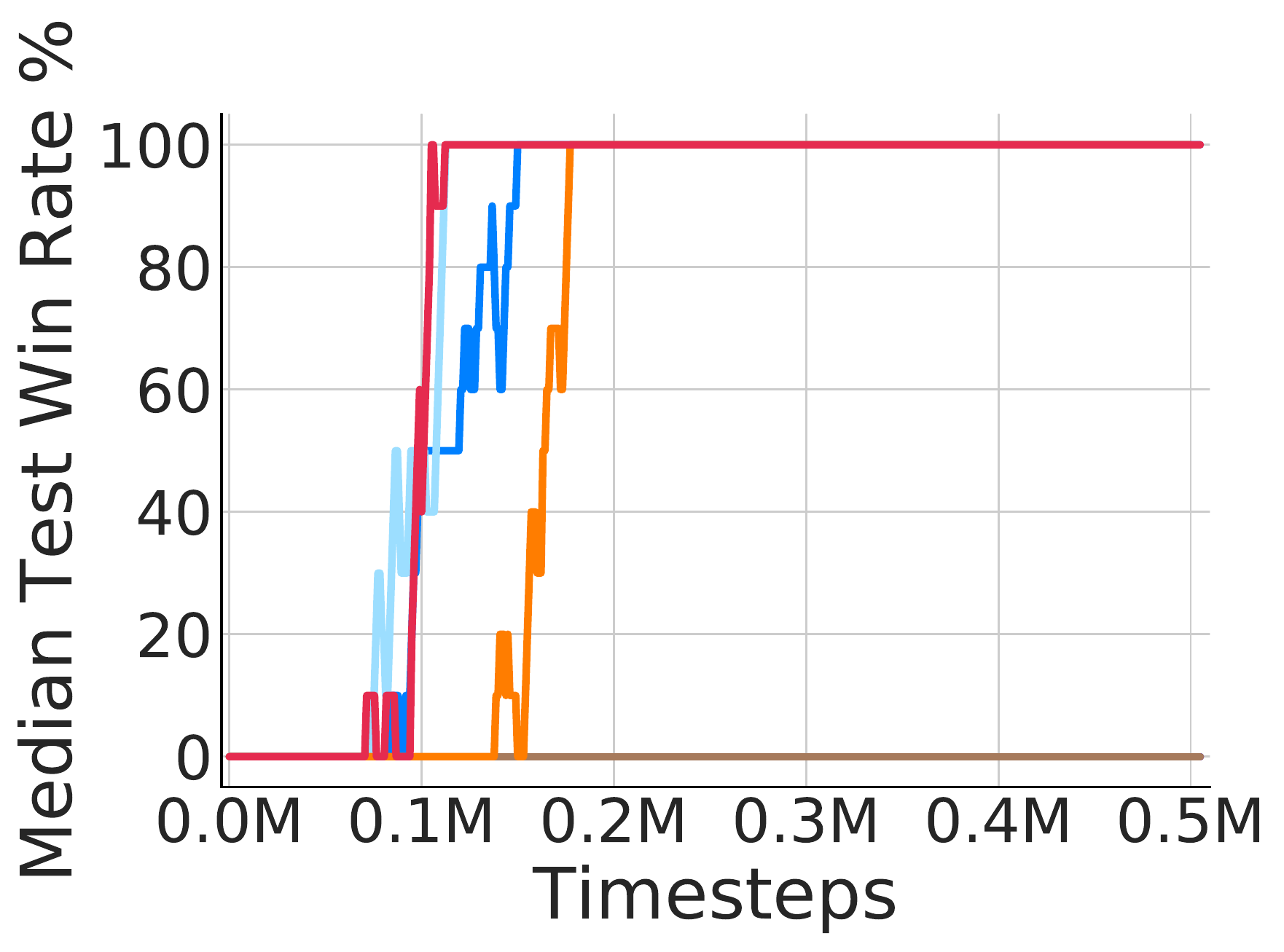}
 \end{minipage}%
 }%
 
 \subfigure[\(p=1.5\)]{
 \begin{minipage}[t]{0.32\linewidth}
 \centering
 \includegraphics[width=1.8in]{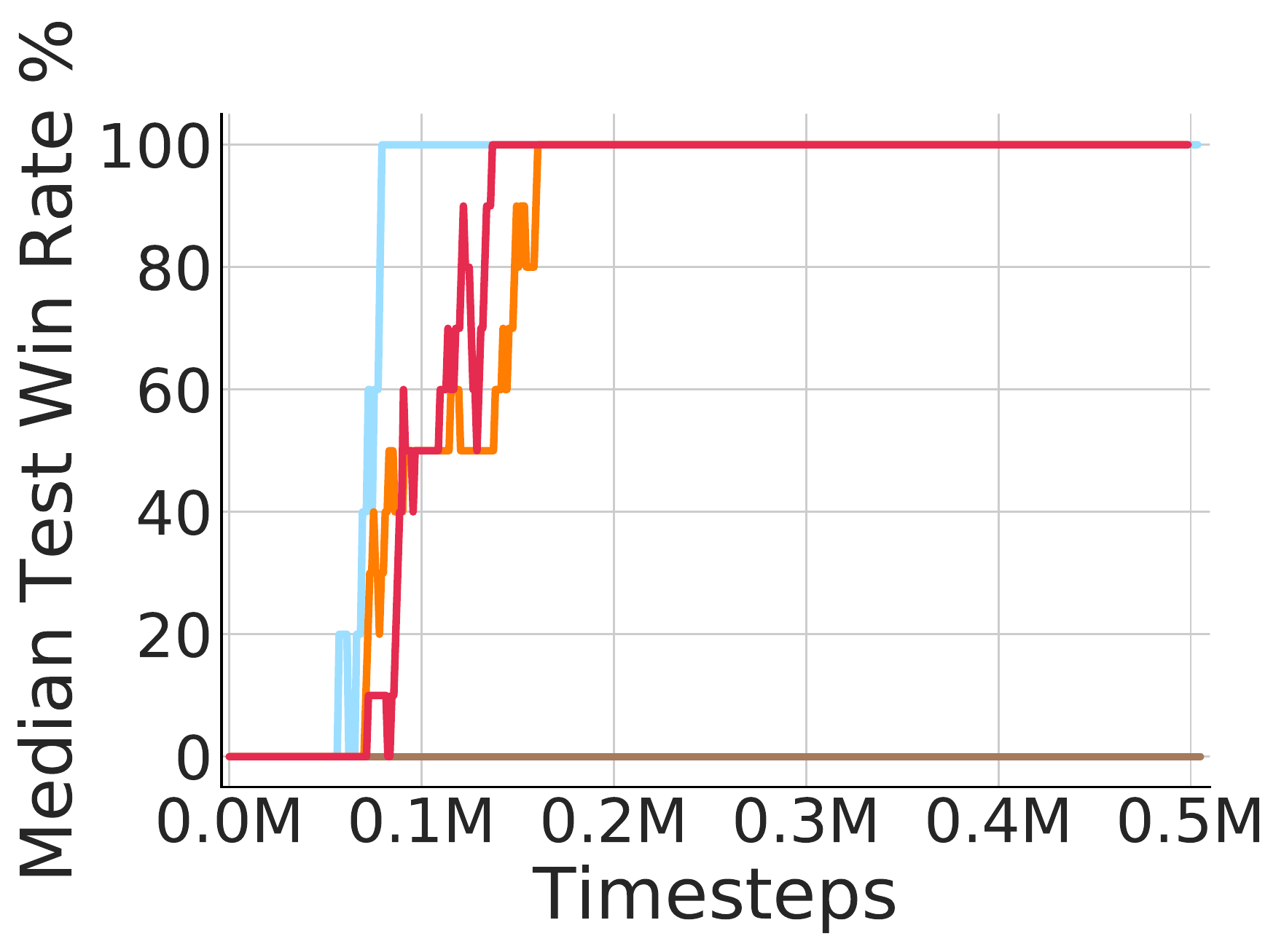}
 \end{minipage}%
 }%
  \subfigure[\(p=2\)]{
 \begin{minipage}[t]{0.32\linewidth}
 \centering
 \includegraphics[width=1.8in]{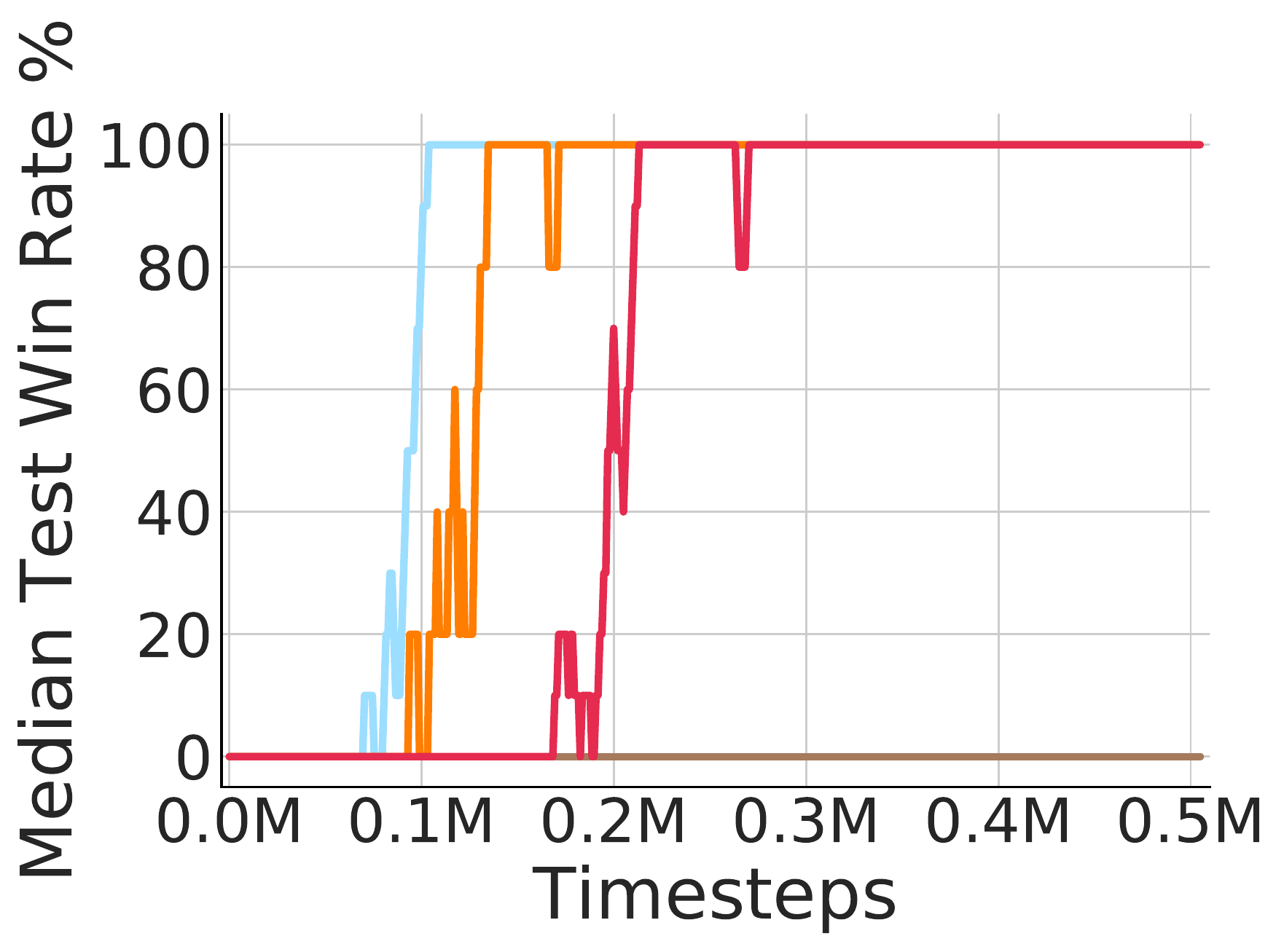}
 \end{minipage}%
 }%
 \centering
 \caption{Results of the coordinated toygame with different punishment \(p\) \label{Fig:toygamemore}.}
 \end{figure*}
 To compare EMC's ability of coordinated exploration with other algorithms, we conduct several experiments with different punishment degrees, i.e., different $p$, in the coordinated toygame~(see Section \ref{toygamesection}). Figure~\ref{Fig:toygamemore} shows, when $p=0$, all algorithms, except for QPLEX and IQL, can find winning strategies quickly since the game is easy. However, under the incoordination penalty~ ($p>0$), algorithms without intrinsic motivation fail to win the game (Figure~\ref{Fig:toygamemore} (b)), since sufficient coordinated exploration needs to be addressed. Moreover, as expected, QPLEX-Local has an advantage over sample efficiency compared with QPLEX-Global~(Figure~\ref{Fig:toygamemore} (b-c)) because of the decentralized exploration, which can avoid searching the whole state space. As $p$ increases~(Figure~\ref{Fig:toygamemore} (d-e)), thanks to the  biased and efficient exploration by predicting individual Q-values, only our methods can solve the task. By these experiments, we can conclude that neither centralized (global) curiosity nor decentralized (local) curiosity is practical for exploration in MARL. In contrast,  predicting individual Q-values can capture the sparse and valuable interactions by leading agents to explore the areas where Q-values are more dynamic, thus achieve coordinated exploration effectively.

 \section{Ablation Study of Coefficient Term}
\begin{figure*}[ht]
\centering
\subfigure[3s5z\_vs\_3s6z]{
 \begin{minipage}[t]{0.32\linewidth}
 \centering
 \includegraphics[width=1.8in]{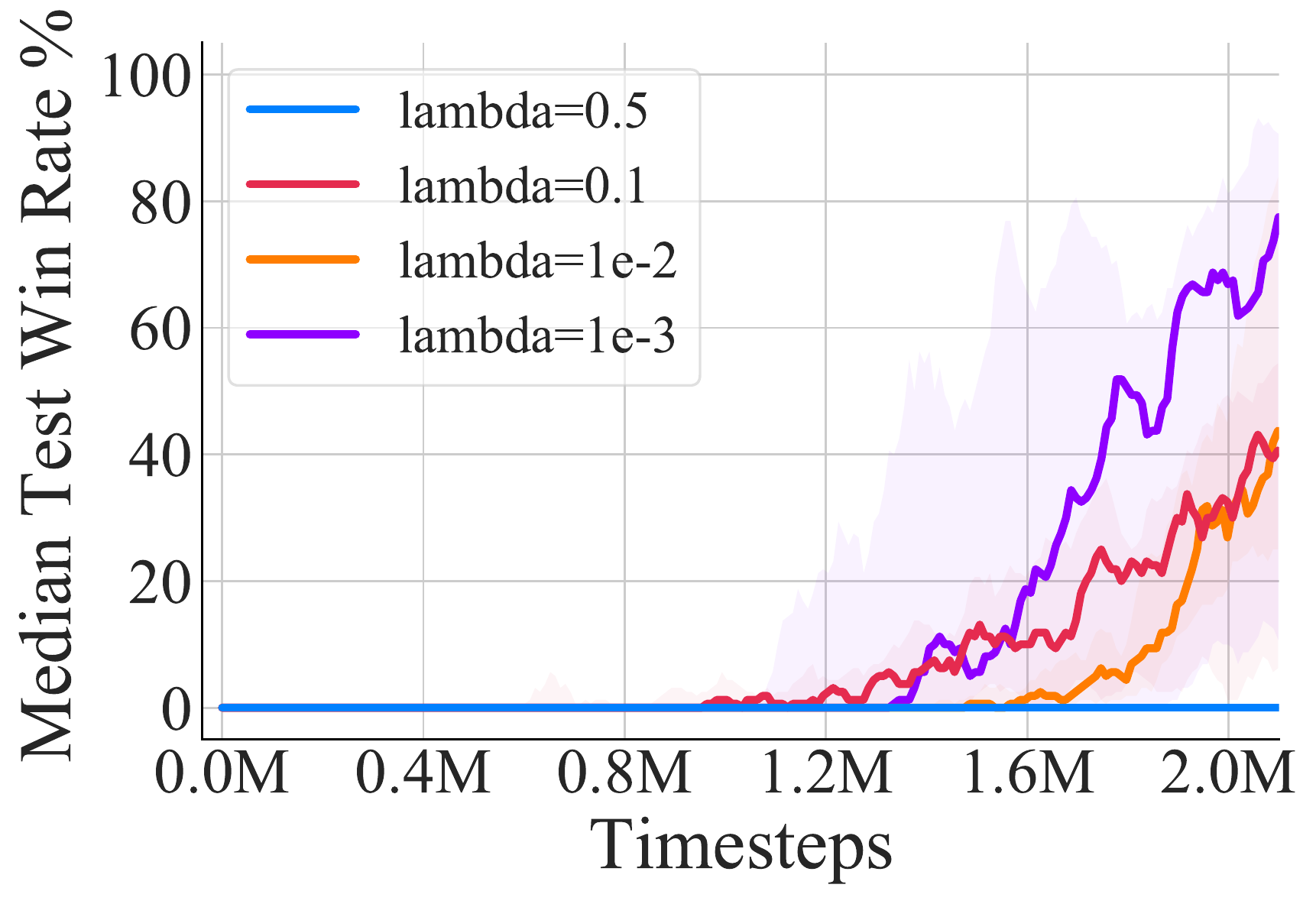}
\end{minipage}
 }%
 \subfigure[2s3z]{
 \begin{minipage}[t]{0.32\linewidth}
 \centering
 \includegraphics[width=1.8in]{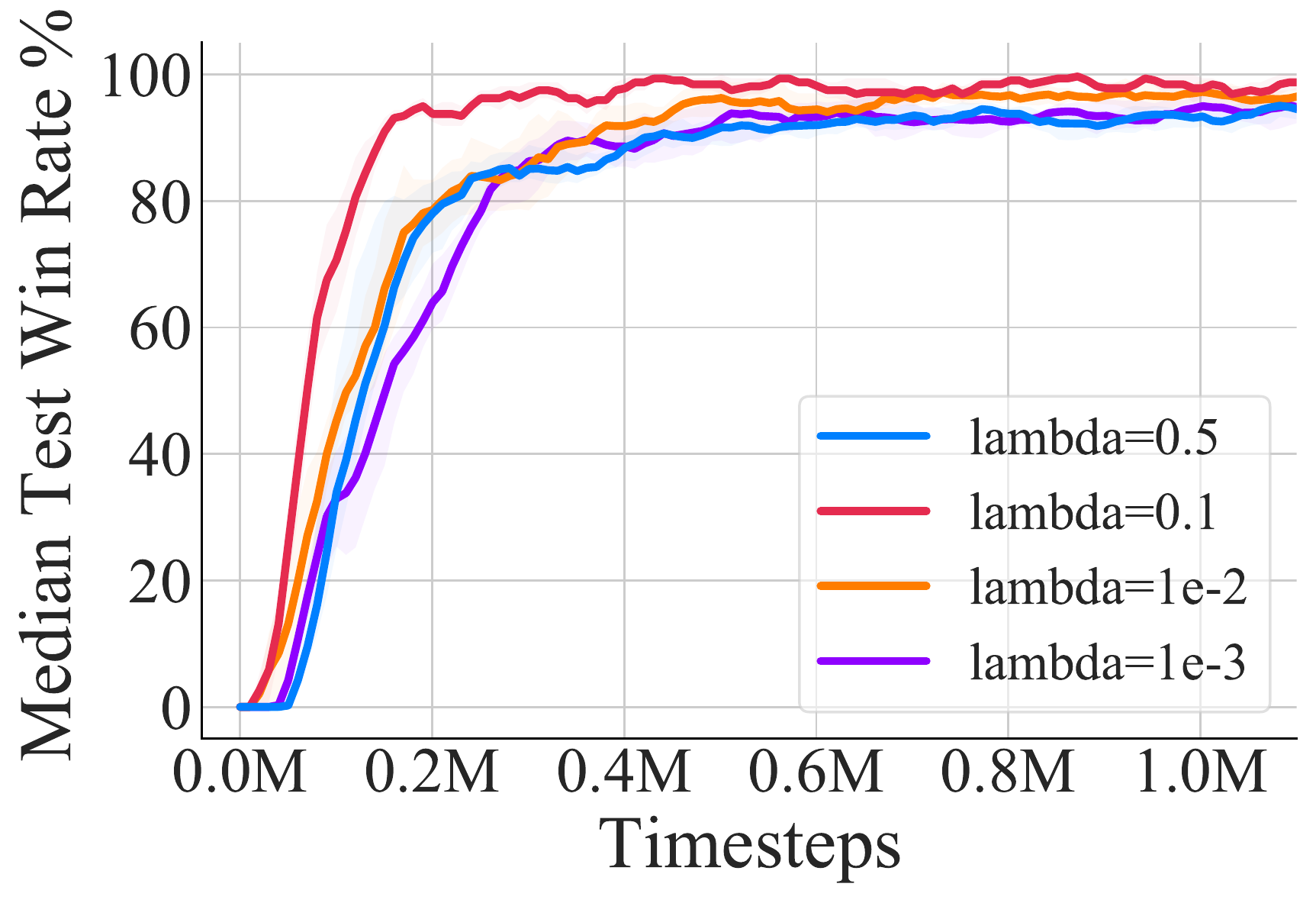}
 \end{minipage}%
 }%
 \subfigure[3s\_vs\_5z]{
 \begin{minipage}[t]{0.32\linewidth}
 \centering
 \includegraphics[width=1.8in]{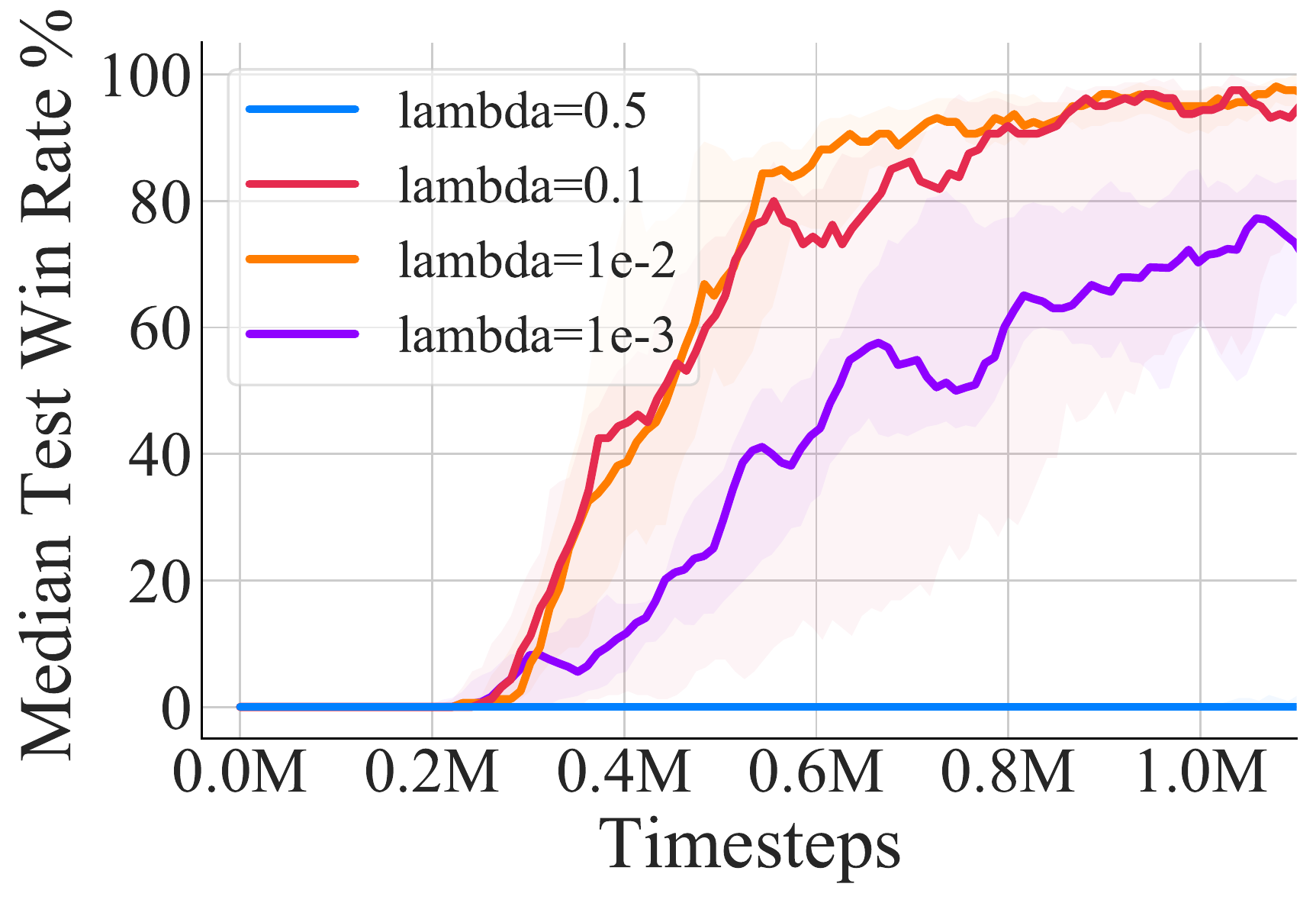}
 \end{minipage}%
  }%
 \centering
 \caption{Ablation study of coefficient term $\lambda$. \label{Fig:Sensitivity}}
 \end{figure*}

In this section, we study the different coefficient term~$\lambda$ in Eq. (8) to visualize the robustness of our hyperparameters. The weighting term $\lambda$ of memory TD loss (see Section~5) was selected in \{0.001, 0.01, 0.1, 0.5\}. We study the influence of different $\lambda$ in several maps, i.e., \textit{2s3z}, \textit{3s5z}, and \textit{3s5z\_vs\_3s6z}. Figure~\ref{Fig:Sensitivity} shows that EMC with $\lambda=0.01 \text{ or } 0.1$ can achieve the state-of-the-art performance. From these empirical experiments, we find that in general, $\lambda$ is not sensitive for most maps when chosen from $0.01\sim 0.1$. However, the performance may be degenerated if $\lambda$ is too large~(e.g., $\lambda=0.5$ in \textit{3s\_vs\_5z}) since the best memorized return of our episodic memory may bring learning into local optimum.


\section{Ablation Study of Design Choice}

The proposed intrinsic reward is the average of individual MSE (Eq. (\ref{intrinsic_reward})) between utilities "Targets" and "Predictors", thus seems to be similar with the average of the Target-utilities individual error (which is just the normalized VDN TD-error). So we take a closer look and investigate the subtle difference between these two implementations. We first discuss the similarities between prediction error~(EMC) and TD error, then we analyze their fundamental (high-level) difference and technical difference, and, finally, we provide empirical results to support our claims.

\textbf{Similarities}

There are some similar properties between prediction error and TD error: 1) they both converge when the policy converges; 2) they are common metrics that show promising results for exploration \citep{RND,ICM,burda2018large,kim2018emi} and exploitation \citep{horgan2018distributed,schaul2015prioritized,hessel2018rainbow}, respectively. These similarities motivate us to study the effects of EMC's each module by comparing with the ablation study using TD error.

\textbf{Fundamental High-level Difference}

A fundamental (high-level) difference is that using TD error as an intrinsic reward can hurt performance since the objective of this intrinsic reward (maximize the TD error) is totally the opposite of the original objective of VDN (minimize the TD error). Intuitively, the metric of TD error does not match the TD-learning framework in the perspective of objective functions. In contrast, we use prediction errors of individual Q-values (i.e., the embeddings of local action-observation histories) as intrinsic rewards for coordinated exploration. To stabilize training, we utilize a slowly updated target network for individual Q-values as a moving target. Using prediction error as a curiosity for exploration has been a long-standing topic in the single-agent deep RL literature \citep{RND,ICM,burda2018large,kim2018emi}, and one important related work is \citep{kim2018emi}, which measures the prediction error in the learnable representation space (i.e., this latent space is also a moving target). The literature \citep{RND,ICM,burda2018large,kim2018emi} implies that the idea of using prediction error is empirically effective, but the novelty of our EMC method is to situate this idea in multi-agent reinforcement learning by exploiting its factorization structure.   
 
\textbf{Technical Difference}
 
More concretely, we will clarify the technical difference between our method (EMC) and the baseline using TD error as intrinsic rewards, as the reviewer suggested. As stated in Eq. (\ref{intrinsic_reward}) in Section 4, EMC (our method) uses the prediction error of the individual utilities as below:
\begin{equation}\label{eq_td_1}
    r^{int}_{EMC}=\frac{1}{N}\sum_{i=1}^{N}{\left \|\widetilde{Q}_{i}(\tau_i,\cdot) - {Q}^{target}_i(\tau_i,\cdot)\right \|}_2.
\end{equation}

While the normalized VDN TD-error (denoted as \textit{EMC-TD}) can be formulated by:

\begin{equation}\label{eq_td_2}
\begin{aligned}
     r^{int}_{TD}=&\frac{1}{N}{\left \|{Q}^{tot}(\bm{\tau},\bm{a}) - \left(r^{ext}+\gamma\max_{\bm{a}'} {Q}_{tot}^{target}(\bm{\tau}',\bm{a}')\right)\right \|}_2 \\
     =&\frac{1}{N}{\left \|\sum_{i=1}^{N}\left({Q}_{i}(\tau_i,\cdot) - \left(r^{ext}+\gamma\max_{a_i'} {Q}^{target}_i(\tau_i',a_i')\right)\right)\right \|}_2
\end{aligned}
\end{equation}
where $\bm{\tau}'$ denotes the joint history on the next state.
If we ignore the difference in the summation operator, the major difference in Eq. (\ref{eq_td_1}) and Eq. (\ref{eq_td_2}) is two-fold:
\begin{itemize}
    \item [a.] TD error uses a one-step temporal difference which is involved with the immediate reward $r^{ext}$.
    \item [b.] TD error uses the VDN utility functions ${Q}_{i}(\tau_i,\cdot)$, while EMC uses the predictors $\widetilde{Q}_{i}(\tau_i,\cdot)$.
\end{itemize}

Therefore, using TD errors as intrinsic rewards may result in the following several issues:
\begin{itemize}
    \item Since it uses the one-step TD target with immediate reward $r^{ext}$, it can be sensitive to noise spikes (e.g. when rewards are stochastic), which can be exacerbated by bootstrapping, where approximation errors appear as another source of noise \citep{schaul2015prioritized}.
    \item Instead of predictors $\widetilde{Q}_{i}(\tau_i,\cdot)$ which are optimized end to end with the targets, VDN utility functions $Q_i$ are learned by one-step reward backpropagation, resulting in that the errors shrink slowly and the agents tend to be stuck in early trajectories. As discussed similarly in \citep{schaul2015prioritized}, the lack of diversity will make the system prone to over-fitting.
\end{itemize}

\textbf{Ablation Study}
 
To investigate the difference between EMC and EMC-TD, we carry out an ablation study on SMAC and a gridworld game. To study the major difference in Eq. (\ref{eq_td_1}) and Eq. (\ref{eq_td_2}) discussed above (i.e., (a) and (b) bullets), we introduce another baseline using $r^{int}_{ind}$ (denoted as \textit{EMC-Ind}), which uses the averaged error between the individual utilities and their targets as intrinsic rewards.
\begin{equation}
    r_{Ind}^{int}=\frac{1}{N}\sum_{i=1}^{N}{\left \|{Q}_{i}(\tau_i,\cdot) - {Q}^{target}_i(\tau_i,\cdot)\right \|}_2.
\end{equation}

EMC-Ind aligns the individual utilities $Q_{i}(\tau_i,\cdot)$ and their target ${Q}_i^{target}(\tau_i,a_i)$ in the same temporal steps and does not include the predictor network $\widetilde{Q}_{i}(\tau_i,\cdot)$. Comparing EMC-Ind with EMC-TD and EMC, we can provide the ablation studies for the effect of (a) and (b), respectively. In the gridworld game, we combine the different curiosity methods (i.e., EMC, EMC-TD, and EMC-Ind) with the VDN learning algorithm and conduct the baseline VDN to demonstrate the effect of these intrinsic rewards in exploration. The experiments are listed as follows.

\begin{figure}[H]
\centering 
\vspace{-0.2in}
\includegraphics[width=0.4\textwidth]{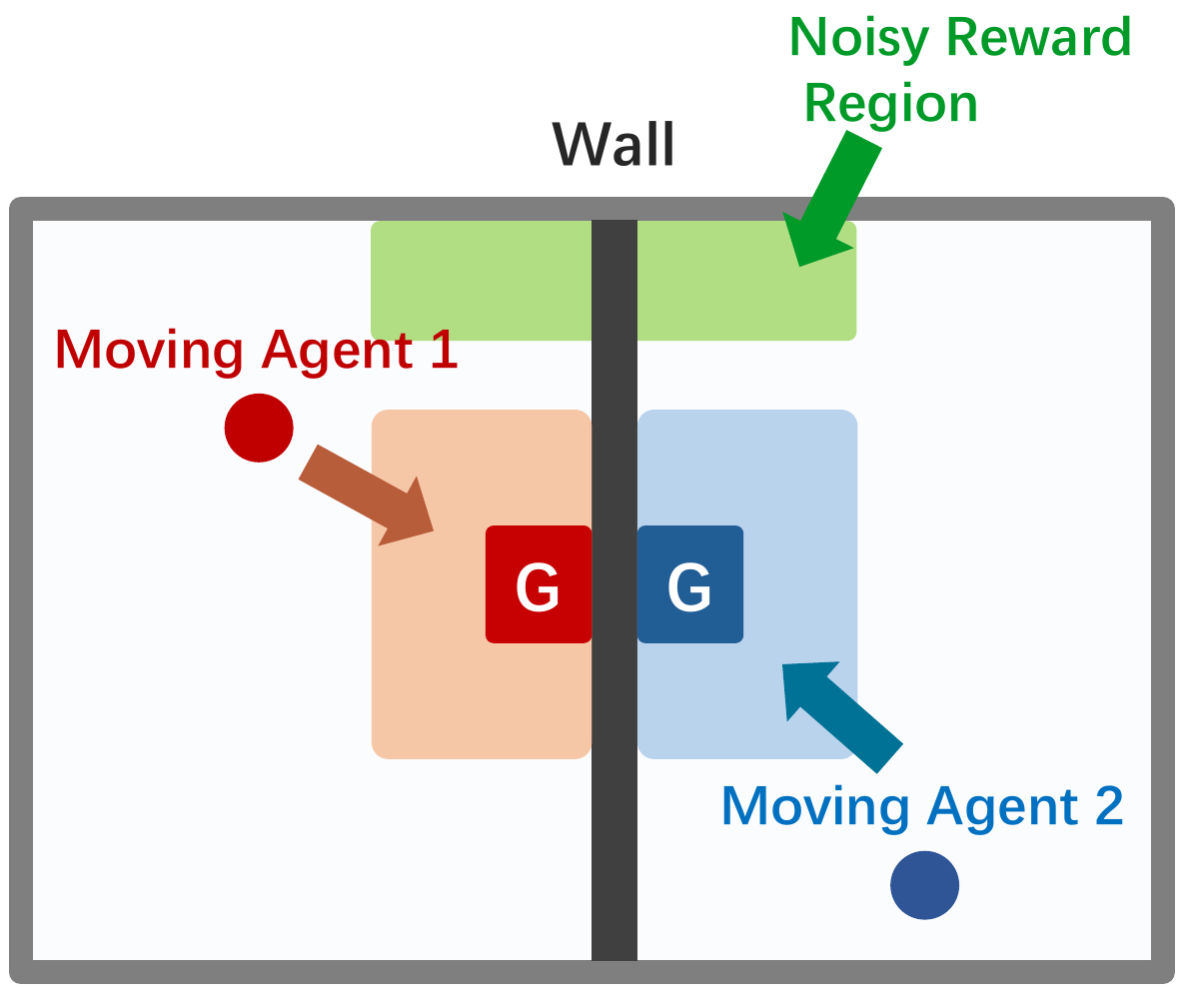}
\caption{New grid world.} 
\label{Fig:New grid world} 
\vspace{-0.2in}
\end{figure}

To better demonstrate the issues caused by (a) and (b), we introduce a new variant of the original gridworld task in the paper (Figure \ref{Fig:New grid world}), which adds a noisy reward region above the shaded area, and two agents will receive a random Gaussian ($\mu=0$, $\sigma=\{0,0.0001,0.01,0.25\}$) noisy reward here. This noisy reward area is used to represent noise spikes in the reward function. The two agents need to jump out the local optimal area (the noisy reward region) and arrive at the goal grid at the same time.

\begin{figure*}[ht]
\centering
\begin{minipage}[t]{\linewidth}
\centering
\includegraphics[width=0.6\textwidth]{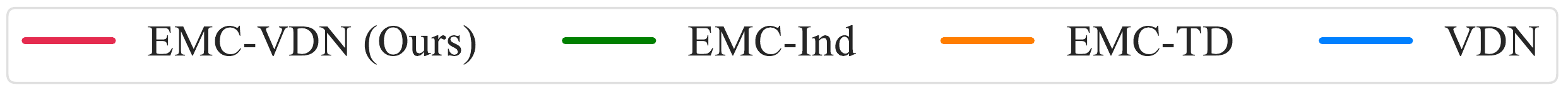}
\end{minipage}%

\subfigure[\(\sigma=0\)]{
 \begin{minipage}[t]{0.32\linewidth}
 \centering
\includegraphics[width=1.8in]{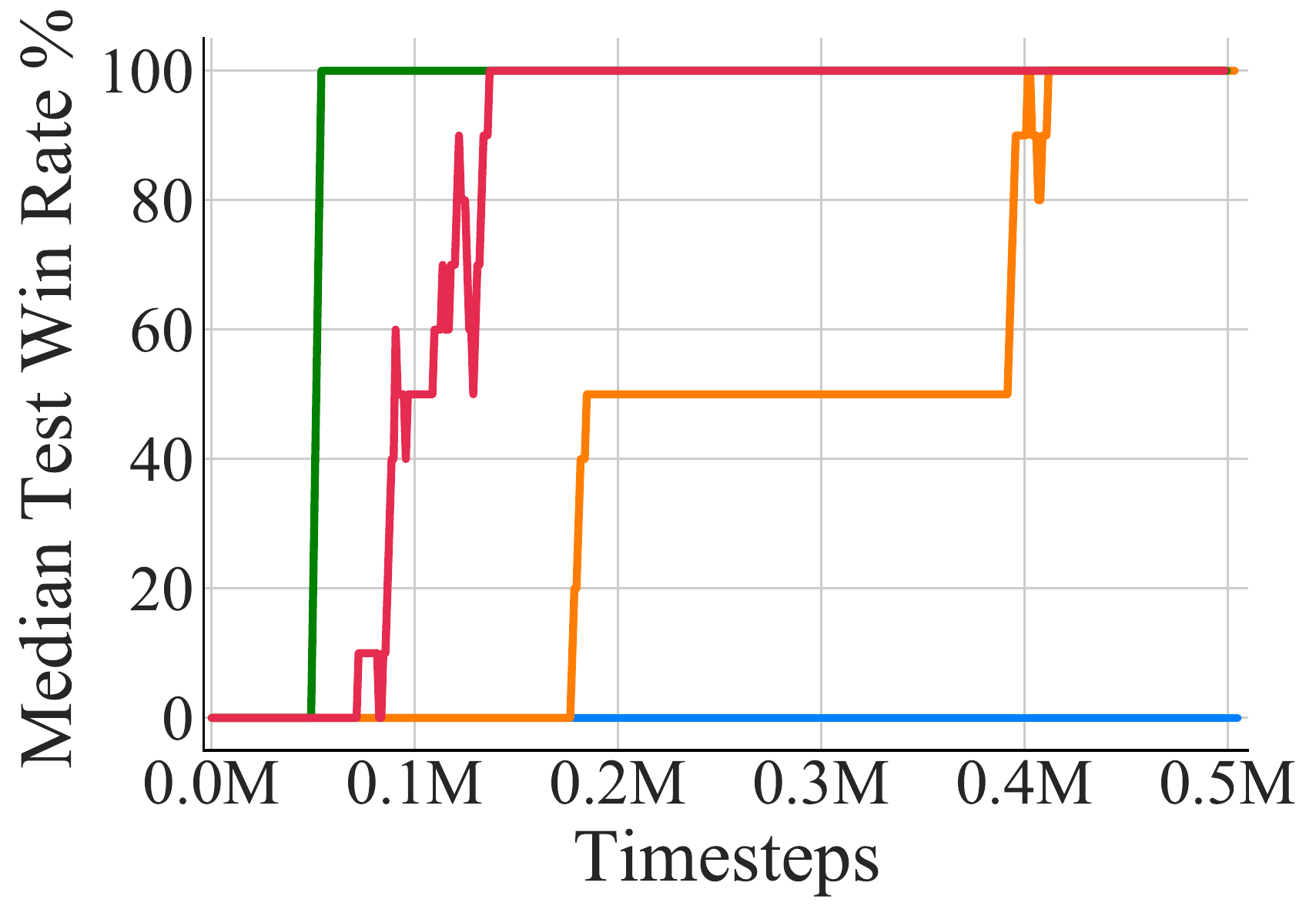}
 \end{minipage}%
}%
 \subfigure[\(\sigma=0.0001\)]{
 \begin{minipage}[t]{0.32\linewidth}
 \centering
 \includegraphics[width=1.8in]{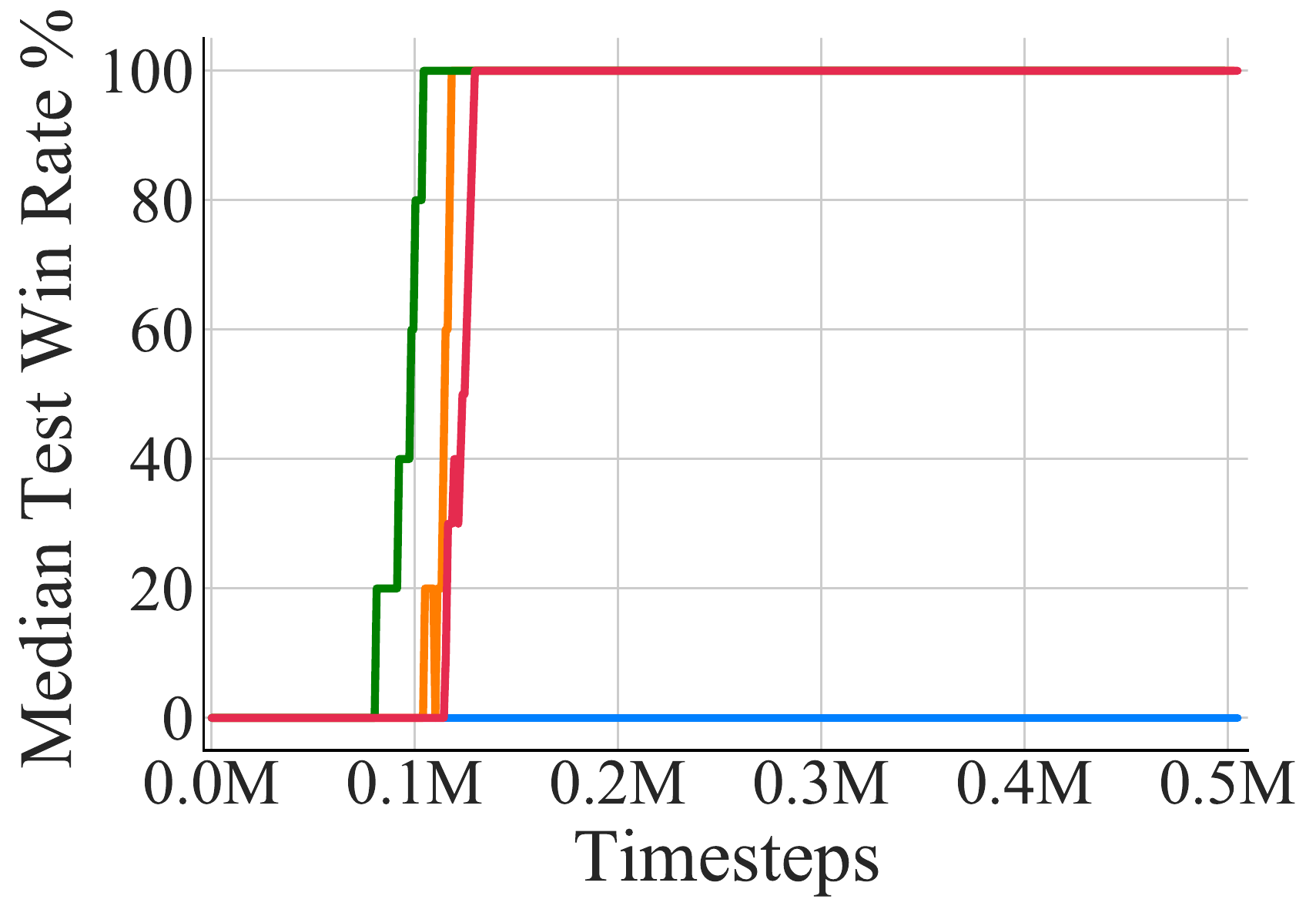}
\end{minipage}
 }%
 \subfigure[\(\sigma=0.01\)]{
 \begin{minipage}[t]{0.32\linewidth}
 \centering
 \includegraphics[width=1.8in]{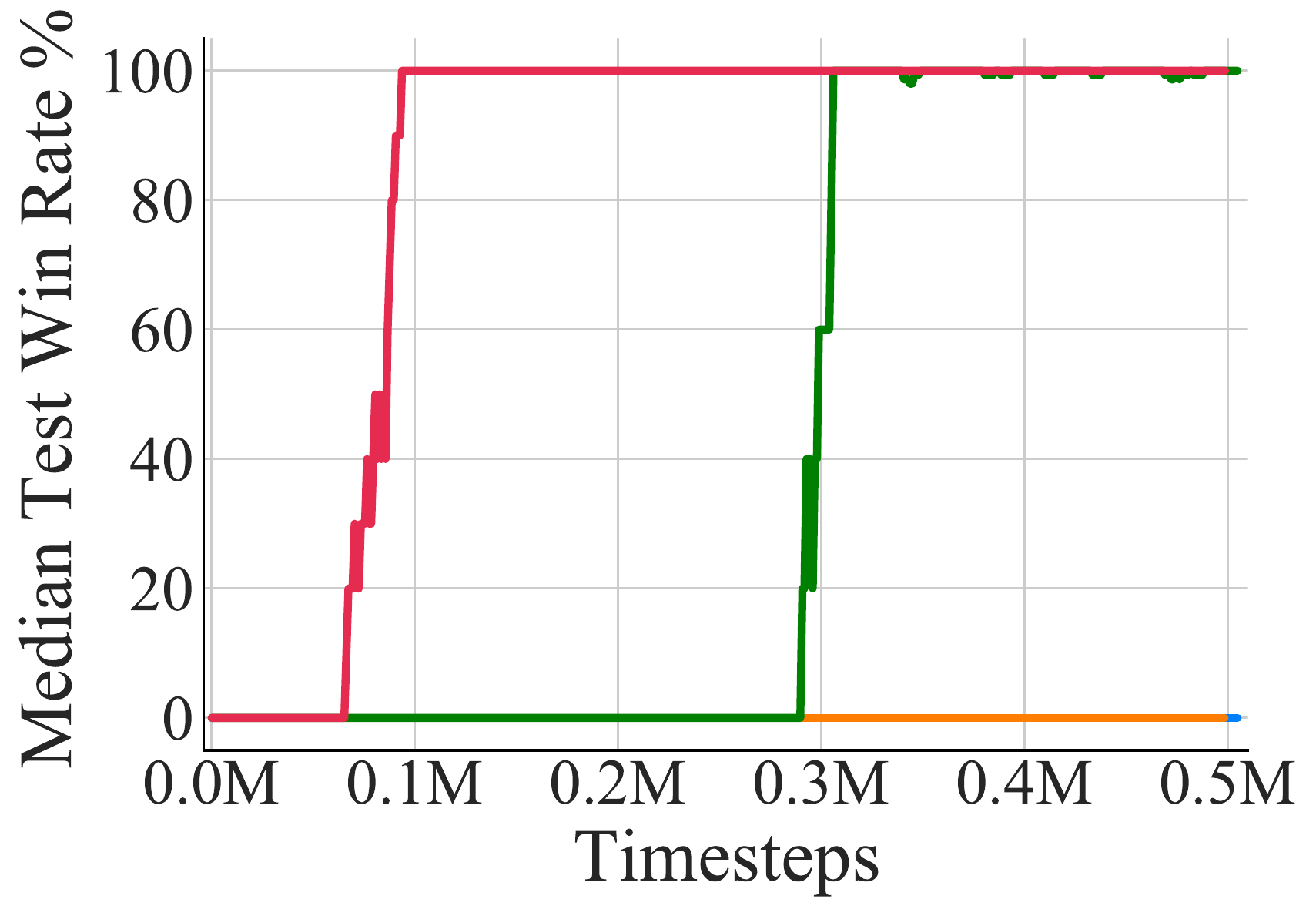}
 \end{minipage}%
 }%
 
 \subfigure[\(\sigma=0.25\)]{
 \begin{minipage}[t]{0.32\linewidth}
 \centering
 \includegraphics[width=1.8in]{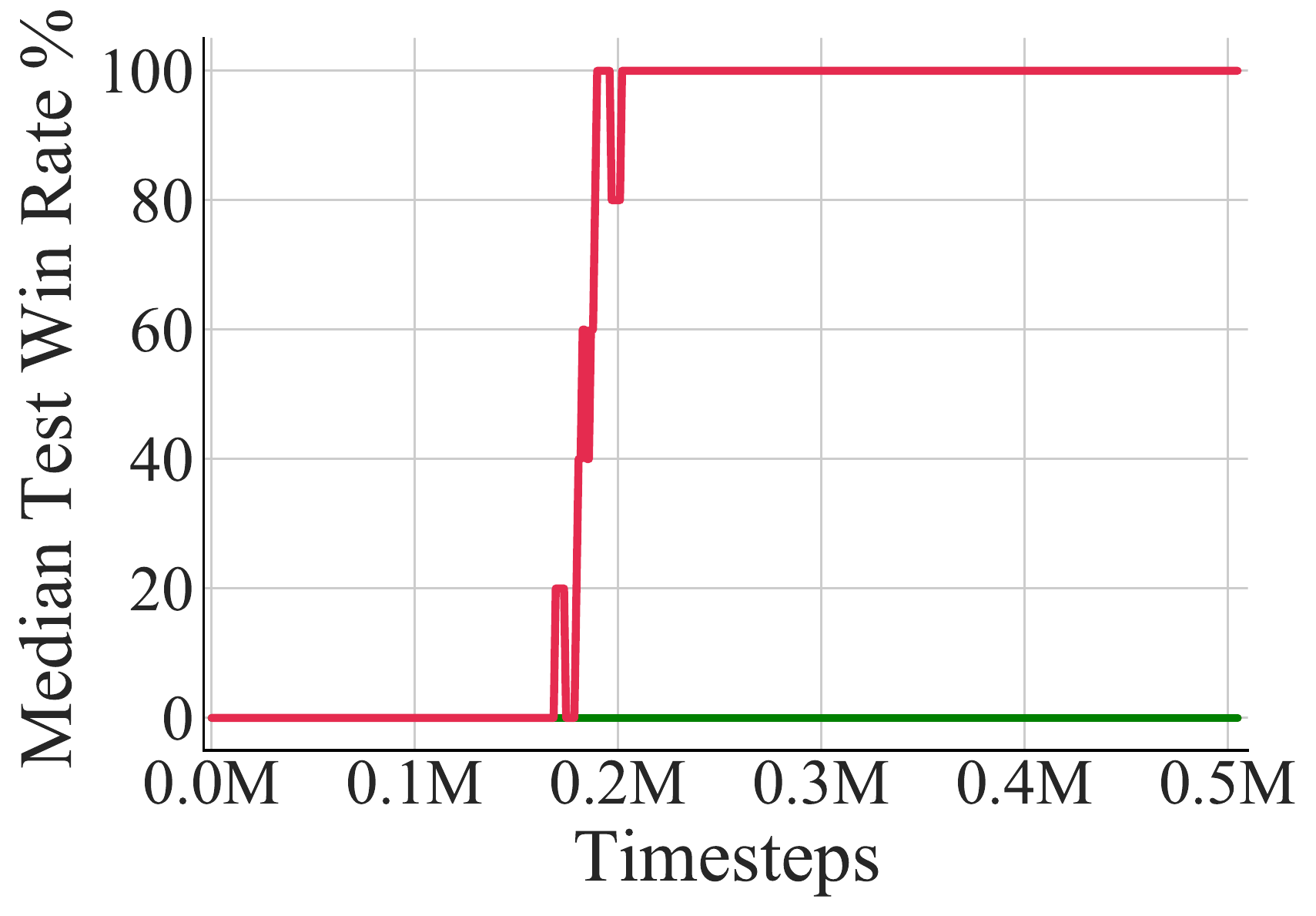}
 \end{minipage}%
 }%
 \centering
 \caption{Results of the coordinated toygame with different scale of noisy reward \label{Fig:toygamemore2}.}
 \end{figure*}

The results in Figure~\ref{Fig:toygamemore2}(a-d) show that EMC-VDN has achieved the best performance and EMC-Ind performs better than EMC-TD. The comparison between EMC-Ind and EMC-TD indicates that one step TD target is sensitive to noise spikes. The reason why EMC-Ind underperforms EMC-VDN is that EMC uses predictors $\tilde{Q}_i$
which are optimized end to end with the targets $Q_i^{target}$, while EMC-Ind uses utility functions $Q_i$ which are optimized with the one-step TD target. Therefore, the intrinsic reward of EMC-VDN will decay as the frequency of visiting the state-action pairs increases (i.e., capturing the novelty of states). In contrast, for EMC-Ind, the optimization of $Q_i$ is influenced by the one-step TD target which is softly updated by a fixed rate, thus the intrinsic rewards cannot vanish along with the number of training steps on the corresponding states (i.e., cannot capture the novelty of states). In other words, the prediction errors of EMC-Ind (i.e., the intrinsic rewards) depend on the update frequency rather than the novelty of visited states. Thus as the scale of noise increases, EMC-Ind and EMC-TD both fail in finding the optimal policy, and EMC-VDN significantly outperforms these two baselines. VDN cannot solve these problems, which indicates that the intrinsic reward introduced by EMC-VDN, EMC-TD, and EMC-Ind is effective for hard exploration problems.
 
\begin{figure}[ht]
\centering
\subfigure[noisy-reward area]{
\begin{minipage}[t]{0.4\linewidth}
\centering
\includegraphics[width=2.5in]{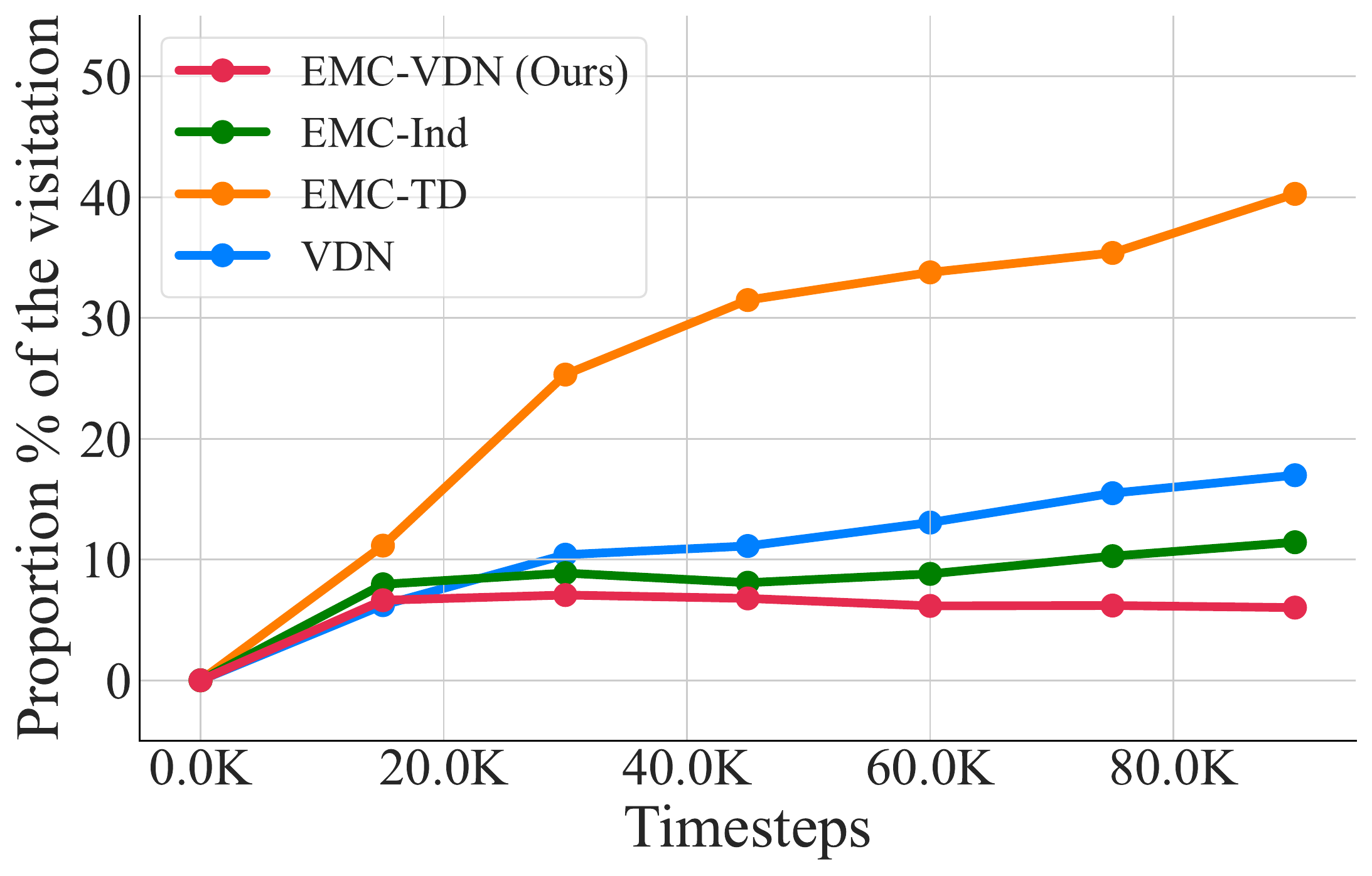}
\end{minipage}%
}%
\hspace{0.5in}
\subfigure[goal grid area]{
\begin{minipage}[t]{0.4\linewidth}
\centering
\includegraphics[width=2.5in]{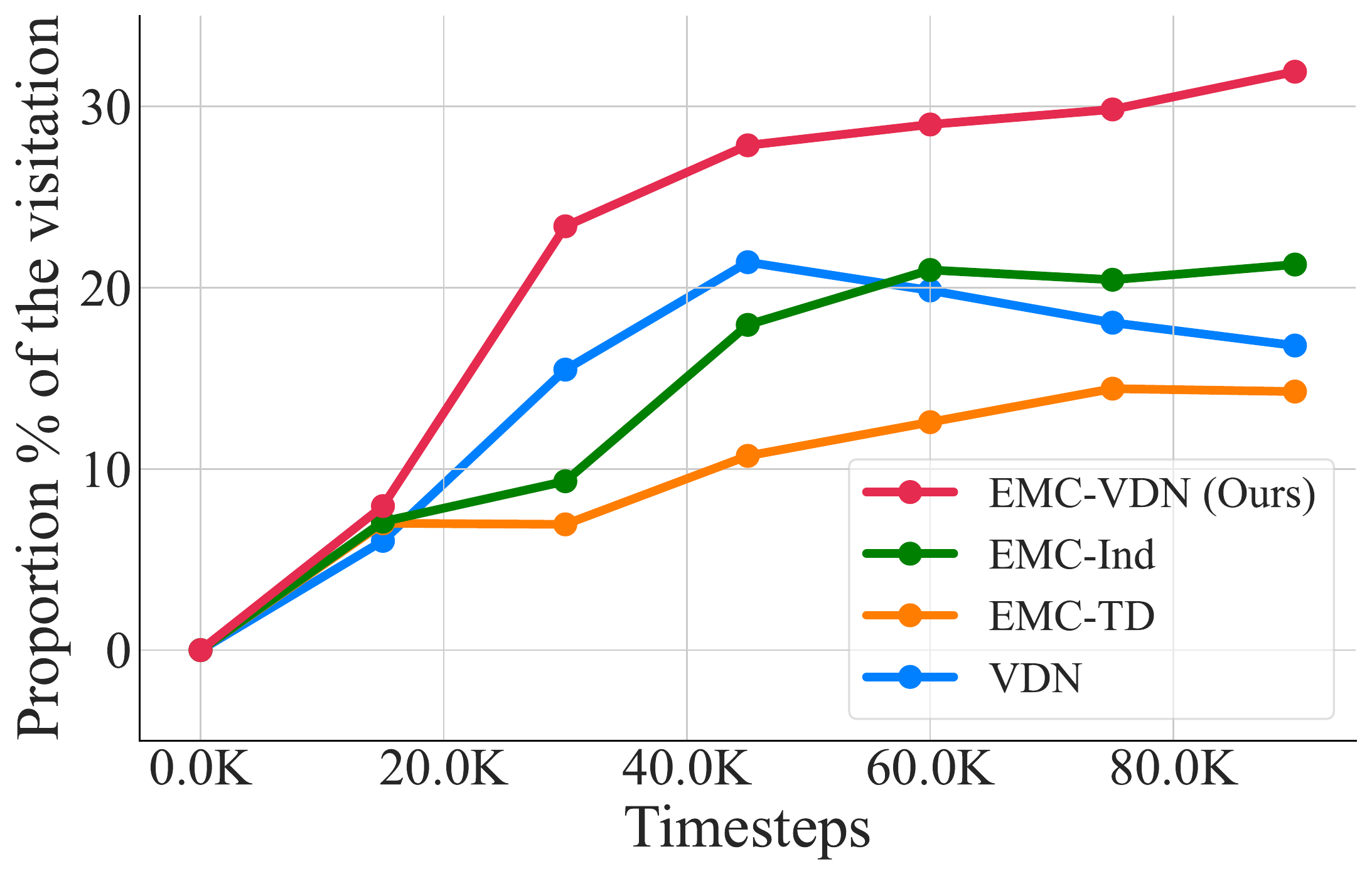}
\end{minipage}%
}%
\centering
\caption{Proportion \% of the visitation in the noisy-reward area and around the goal grid area ($\sigma=0.25$), respectively.}\label{fig_visitation}
\end{figure}

For more clear clarifications, we provide the proportion of the visitation in the areas of noisy-reward and goal grid in the gridworld with $\sigma=0.25$, respectively, and the results are shown in Figure~\ref{fig_visitation}. As we expected, the results show that, due to reasons discussed above, the agents of EMC-Ind or EMC-TD tend to get stuck in this noisy-reward region, and EMC-VDN can jump out of the local optimal area and reach the goal grid. Compared with EMC-Ind, the ability to capture state novelty provides EMC-VDN with more efficient curiosity-driven exploration. On the other hand, especially compared with VDN, EMC-TD will be stuck in the noisy-reward region longer, which shows that TD-error cannot perform well under higher noise spikes.


\begin{figure}[ht]
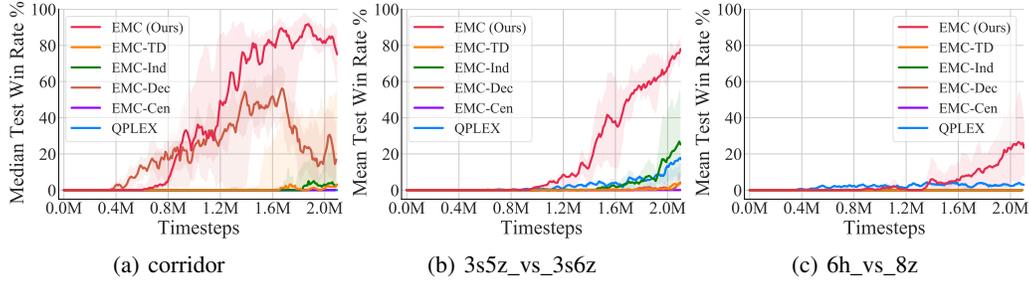

\centering
\subfigure[corridor]{
\begin{minipage}[t]{0.32\linewidth}
\centering
\includegraphics[width=1.8in]{figure/corridor_ablationTD.pdf}
\end{minipage}%
}%
\subfigure[3s5z\_vs\_3s6z]{
\begin{minipage}[t]{0.32\linewidth}
\centering
\includegraphics[width=1.8in]{figure/3s5z_vs_3s6z_ablationTD.pdf}
\end{minipage}%
}%
\subfigure[6h\_vs\_8z]{
\begin{minipage}[t]{0.32\linewidth}
\centering
\includegraphics[width=1.8in]{figure/6h_vs_8z_ablationTD.pdf}
\end{minipage}%
}%
\centering
\caption{Ablation study on design choice.}
\label{Fig:Ablation Study3}
\end{figure}

Figure~\ref{Fig:Ablation Study3} show the results of SMAC. We can see that EMC-Ind shows advantages over EMC-TD in corridor and 3s5z\_vs\_3s6z map, demonstrating that using one-step difference may harm performance. However, the winning rate of EMC-Ind shows relatively low compared with EMC (ours), indicating that it may be stuck in local optima.

As discussed above, we choose to use prediction errors as intrinsic rewards instead of using TD error, which can capture the dynamics of $Q_i$ quickly, avoid the noise spikes problem, and jump out of the local optima effectively. Empirical results show that EMC can encourage the agents to visit novel and promising states efficiently.

\newpage
\section{Ablation Study of Episodic Memory}
\begin{figure*}[ht]
\centering
\begin{minipage}[t]{\linewidth}
\centering
\includegraphics[width=0.9\textwidth]{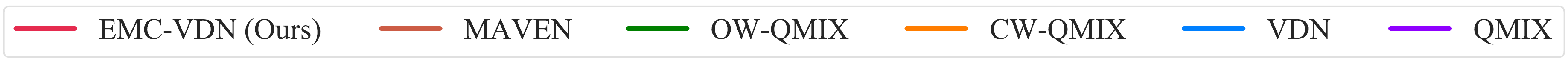}
\end{minipage}%

\subfigure[\(p=2,\xi=0\)]{
 \begin{minipage}[t]{0.32\linewidth}
 \centering
\includegraphics[width=1.8in]{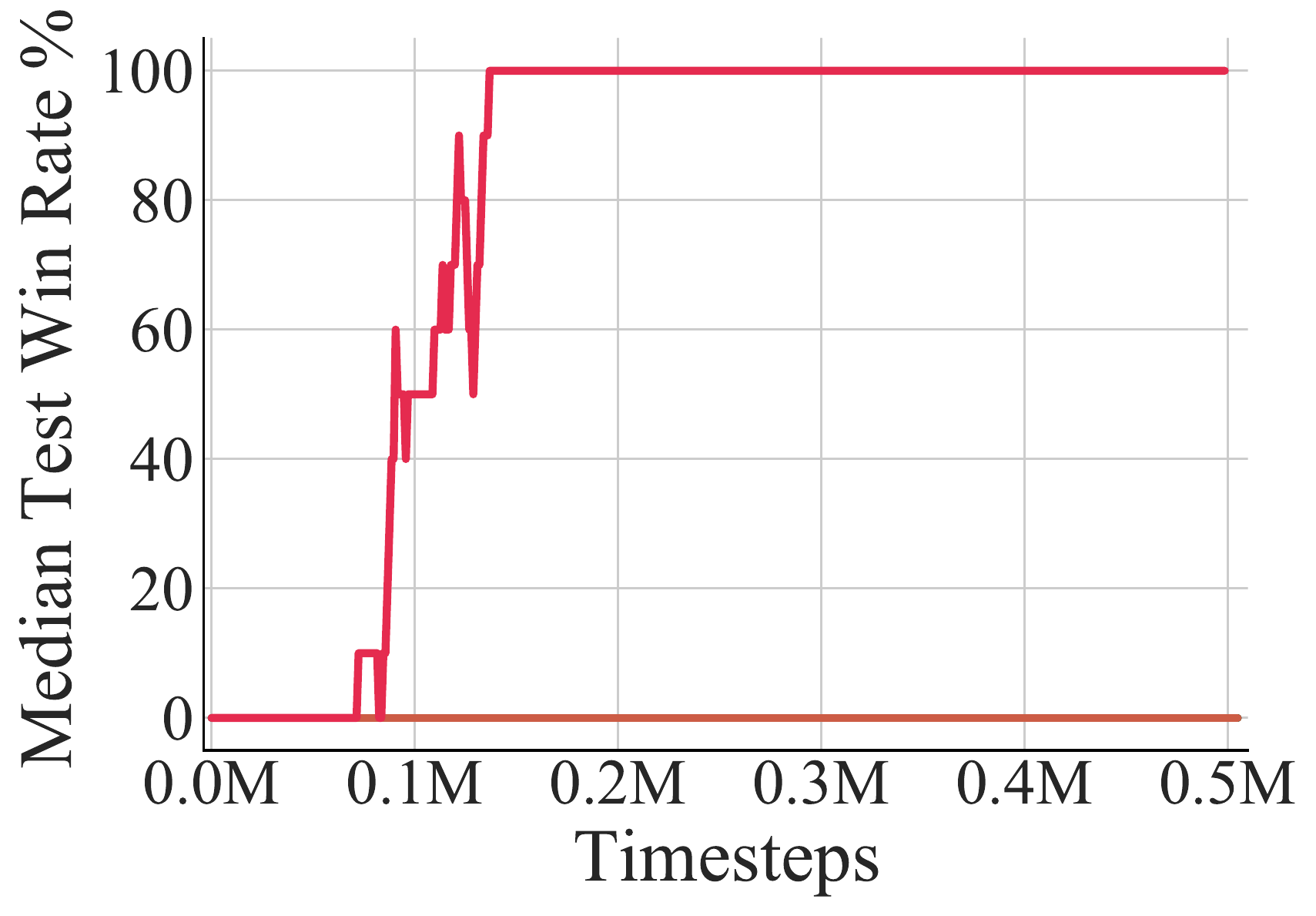}
 \end{minipage}%
}%
\subfigure[\(p=2,\xi=5\)]{
 \begin{minipage}[t]{0.32\linewidth}
 \centering
\includegraphics[width=1.8in]{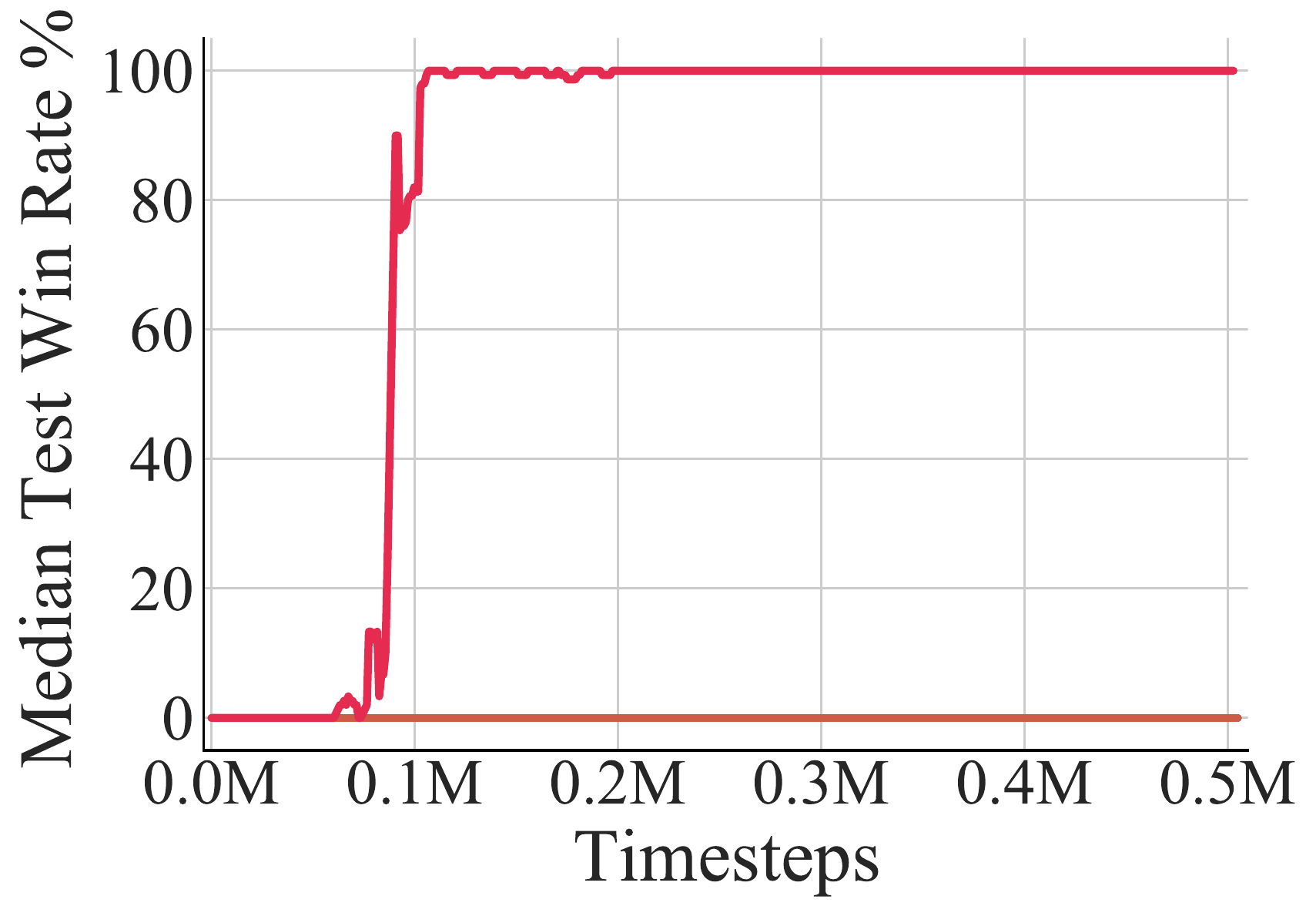}
 \end{minipage}%
}%
 \subfigure[\(p=2,\xi=10\)]{
 \begin{minipage}[t]{0.32\linewidth}
 \centering
 \includegraphics[width=1.8in]{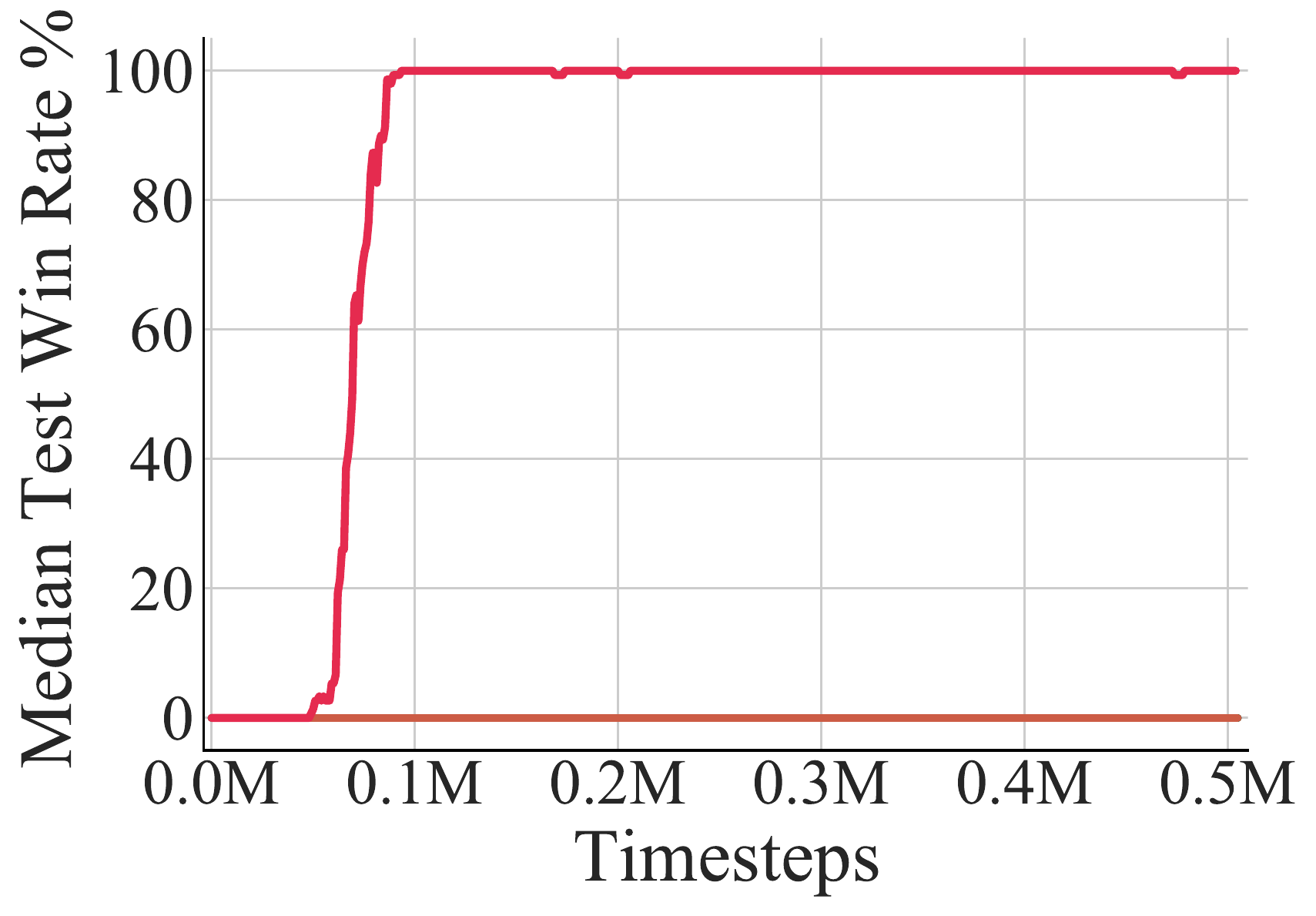}
\end{minipage}
 }%
 
 \subfigure[\(p=2,\xi=20\)]{
 \begin{minipage}[t]{0.32\linewidth}
 \centering
 \includegraphics[width=1.8in]{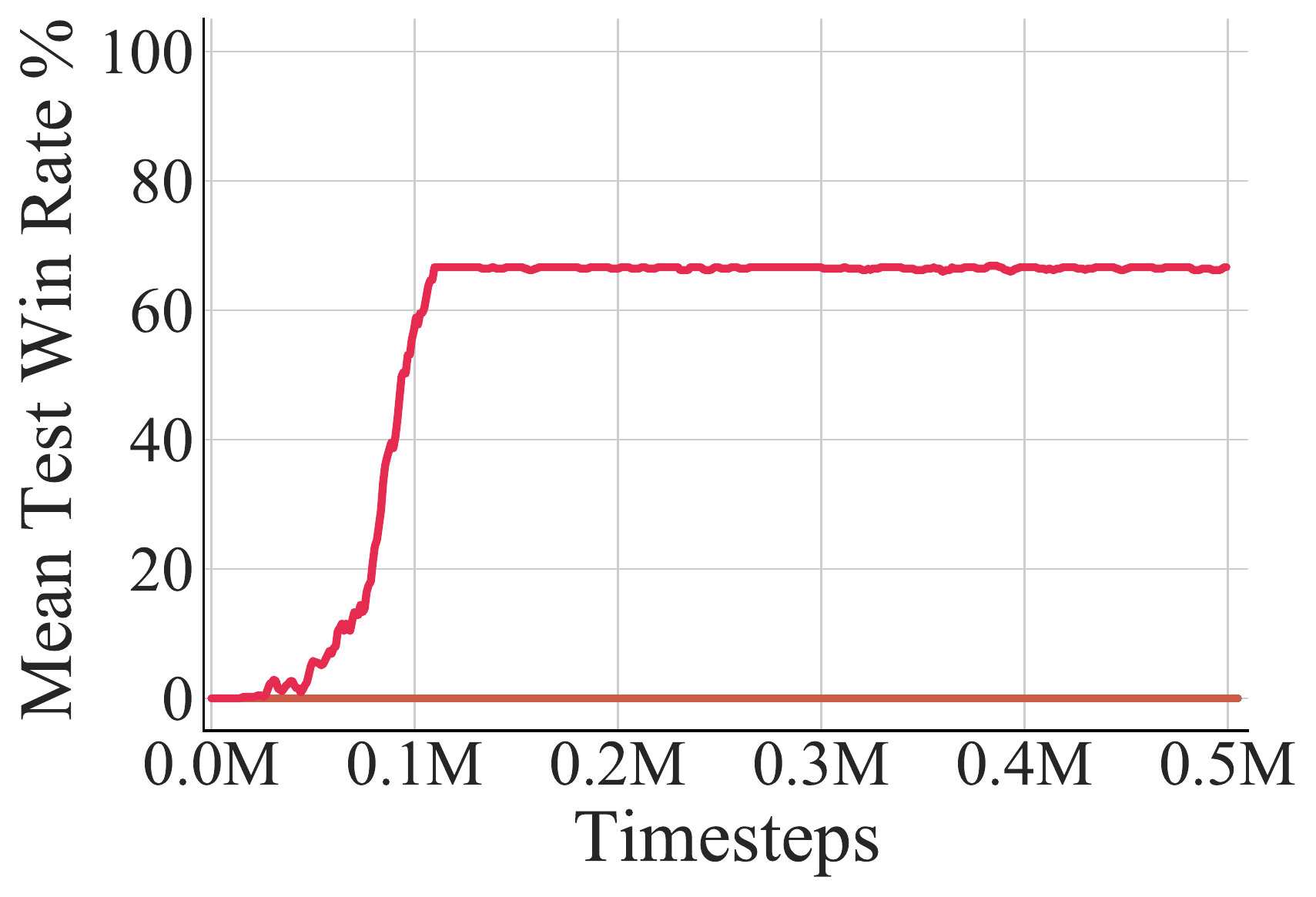}
 \end{minipage}%
 }%
  \subfigure[\(p=2,\xi=30\)]{
 \begin{minipage}[t]{0.32\linewidth}
 \centering
 \includegraphics[width=1.8in]{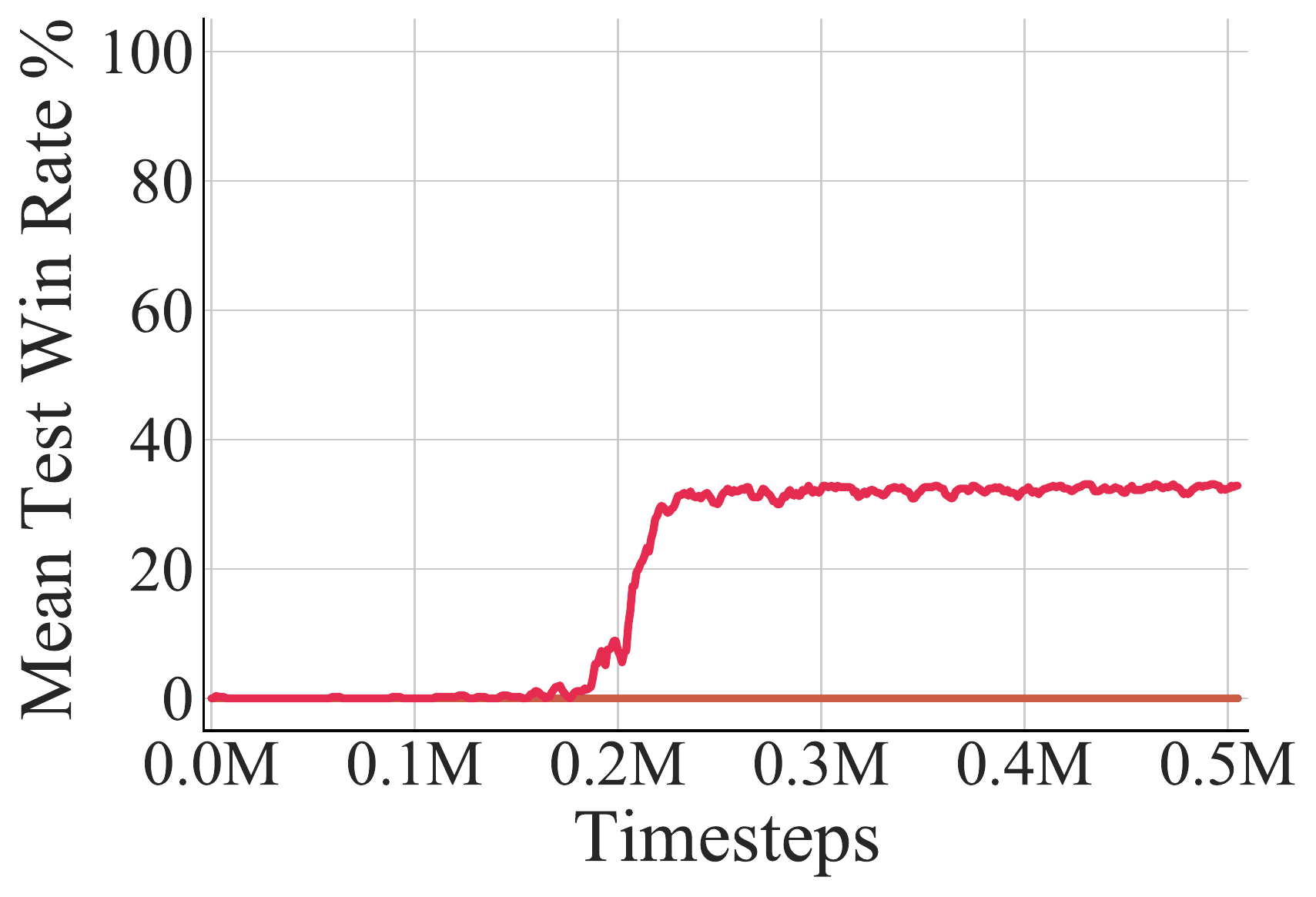}
 \end{minipage}%
 }%
 \subfigure[\(p=2,\xi=50\)]{
 \begin{minipage}[t]{0.32\linewidth}
 \centering
 \includegraphics[width=1.8in]{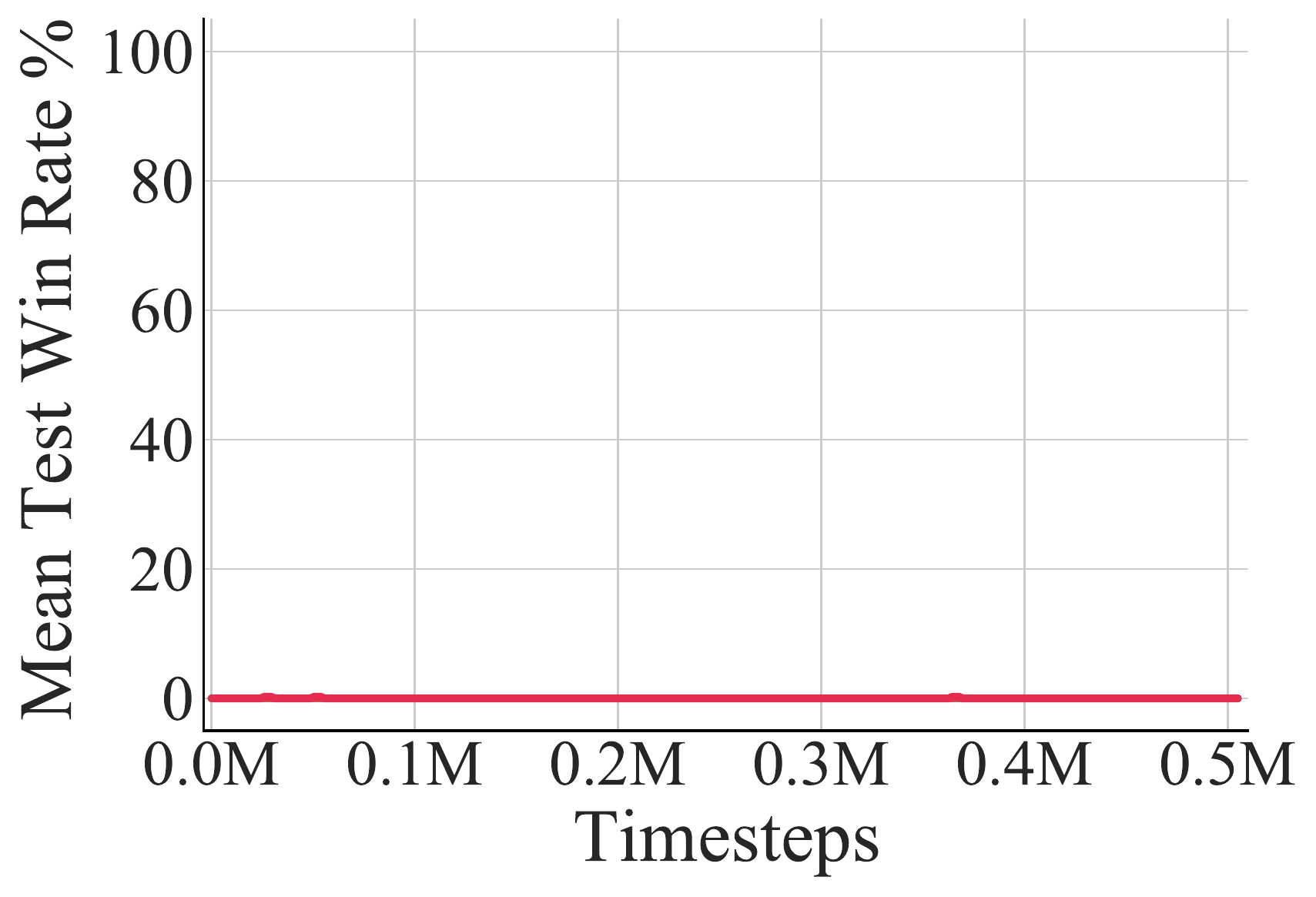}
 \end{minipage}%
 }%
 \centering
 \caption{Results of the coordinated toygame with different scale of noisy reward \label{Fig:toygamemore3}.}
 \end{figure*}

To illustrate the ability of EMC in the stochastic setting, we conduct an ablation study by introducing stochasticity into the gridworld didactic task. In the original gridworld, each agent can move in four directions or stay still at each time step. In the stochastic variant, each agent has a $\xi$\% probability of making a mistake and choose a random action accordingly. Figure~\ref{Fig:toygamemore3}(a-e) shows the performance of EMC and other baselines under different degrees of stochasticity $(\xi=\{0, 5, 10, 20,30, 50\})$. These empirical results show that with proper stochasticity (e.g., $\xi=\{0, 5, 10, 20\}$), EMC can also significantly outperform baselines and solve this hard exploration puzzle. When the environment has a lot of uncertainty (e.g., $\xi=30,50$), it is challenging for EMC and other baselines. Learning in the highly stochastic environment is also an interesting future direction for MARL.

\newpage

\clearpage

\end{document}